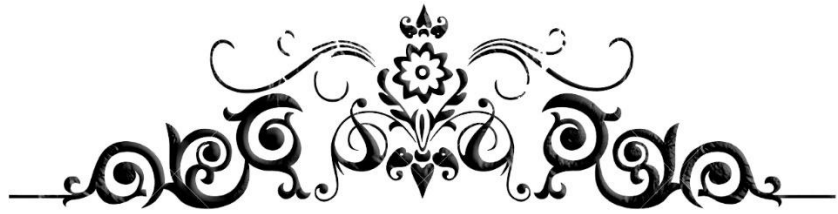

# Introduction to Facial Micro Expressions Analysis Using Color and Depth Images:

## *(A MATLAB Coding Approach)*

<mark>**Second Edition (2023)**</mark>

**By:**

*Seyed Muhammad Hossein Mousavi*

<mark>*July 2016 to June 2023*</mark>

# Introduction to Facial Micro Expressions Analysis Using Color and Depth Images (a MATLAB Coding Approach)

*Seyed Muhammad Hossein Mousavi*

mosavi.a.i.buali@gmail.com **Or** hossein-mousavi@ieee.org

**ORCID ID**: 0000-0001-6906-2152

# Preface:

The book attempts to introduce a gentle introduction to the field of Facial Micro Expressions Recognition (FMER) using Color and Depth images, with the aid of MATLAB programming environment. FMER is a subset of image processing and it is a multidisciplinary topic to analysis. So, it requires familiarity with other topics of Artificial Intelligence (AI) such as machine learning, digital image processing, psychology and more. So, it is a great opportunity to write a book which covers all of these topics for beginner to professional readers in the field of AI and even without having background of AI.

My goal is to provide a standalone introduction in the field of FMER analysis in the form of theorical descriptions for readers with no background in image processing with reproducible MATLAB practical examples. Also, the book describes any basic definitions for FMER analysis and MATLAB library which is used in the text, that helps final reader to apply the experiments in the real-world applications. This book is suitable for students, researchers, and professionals alike, who need to develop practical skills, along with a basic understanding of the field.

It is expected that, after reading this book, the reader feels comfortable with different key stages such as color and depth image processing, color and depth image representation, classification, machine learning, facial micro expressions recognition, feature extraction and dimensionality reduction.

This book is product of several years of researches and experiments and reflects the mindset of the authors for understanding this field as easier as possible. The author encourages the reader to contact him with any comments and suggestions for improvement.

# ACKNOWLEDGMENTS

*This book has improved thanks to the support of a number of colleagues and friends, who have provided generous feedback and constructive comments, during the writing process. S.M.H Mousavi would like to thank his family for their patience and generous support and dedicates this book to all the teachers who have shaped his life.*

---

**GitHub Repository:**

*https://github.com/SeyedMuhammadHosseinMousavi/Introduction-to-Facial-Micro-Expressions-Analysis-Using-Color-and-Depth-Images-a-Matlab-Coding-Appr*

---

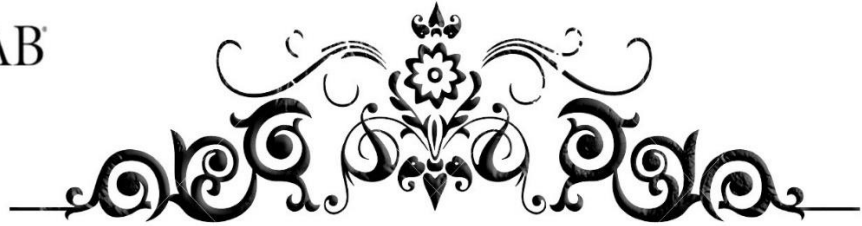

# Introduction to Facial Micro Expressions Analysis Using Color and Depth Images  *(A MATLAB Coding Approach)*

*Seyed Muhammad Hossein Mousavi*

mosavi.a.i.buali@gmail.com Or hossein-mousavi@ieee.org

ORCID ID: 0000-0001-6906-2152

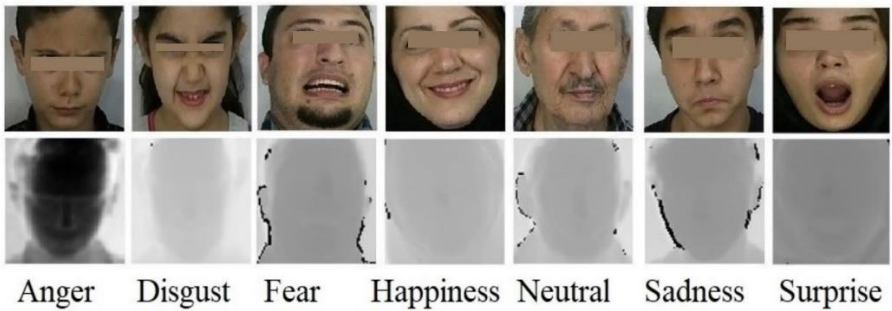

**RGB**

**Depth**

Anger    Disgust    Fear    Happiness    Neutral    Sadness    Surprise

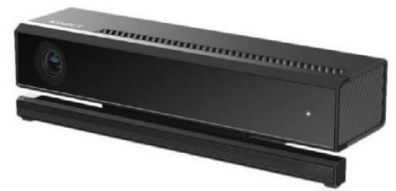

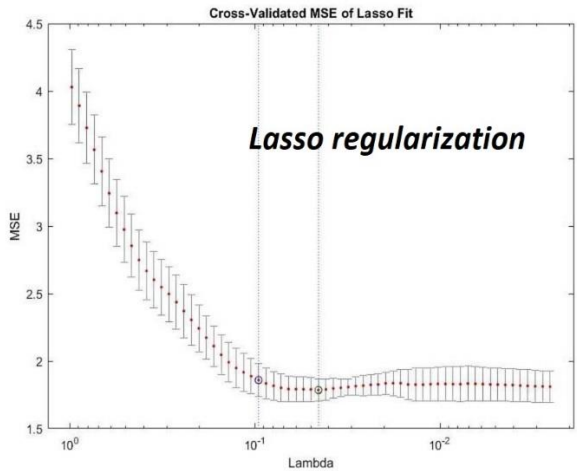

*Lasso regularization*

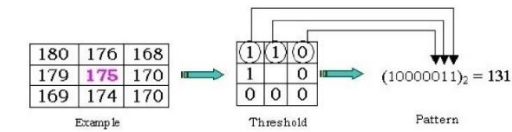

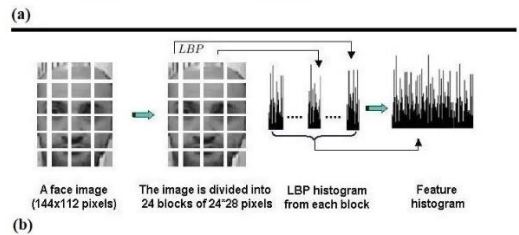

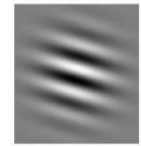

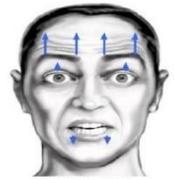

SURPRISE:
AU1+2+5+25+26

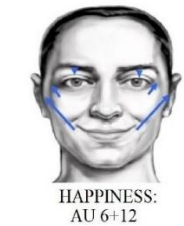

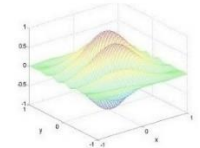

HAPPINESS:
AU 6+12

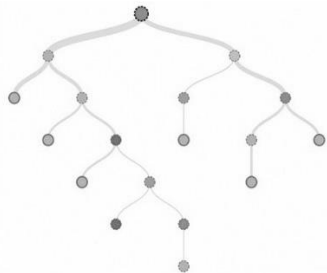

**Decision Tree**

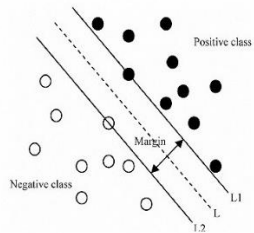

**SVM**

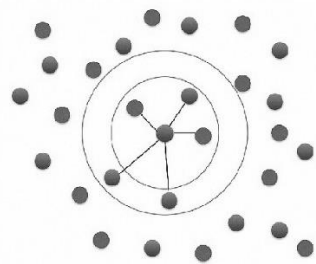

**K-NN**

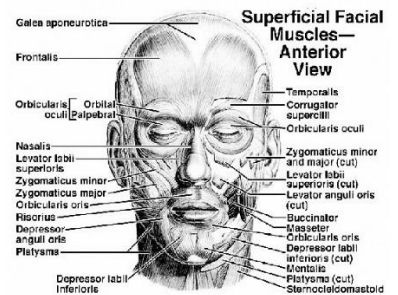

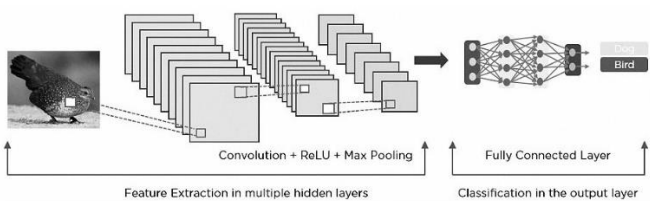

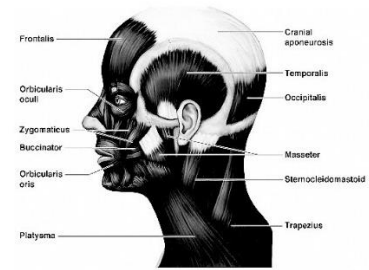

# *LIST OF CONTENTS*

# LIST OF CONTENTS

# *LIST OF FIGURES*

# *LIST OF FIGURES*

# *Chapter    1*
# *Introduction*

# *Chapter 1*

# *Introduction*

## Contents

During recent years, we have witnessed the increasing usage of infrared or night vison sensors [1] alongside with traditional color sensors which aid color sensors to have vision both in day and night times. But infrared sensors could be used in different application like 3-Dimentional (3-D) [2] modeling and calculating the distance between object and sensor. Despite of providing night vision application, Infrared sensors provide Depth data or 2.5-Dimentional (2.5-D) [3] image which could be converted to 3-D model. This 2.5-D depth image provides more details of the object versus traditional color image. For example, it is possible to analysis tiny face wrinkles in both day and night time which color sensors are not capable of.

Infrared sensors have application in surveillance and security [4], medical purposes [5], entertainment [6], military [7], gaming industry [8], agriculture [9], psychology [10] and more.

This book focused on face analysis [11] and especially facial expressions and facial micro expressions recognition [12] purposes using traditional color and depth images. Each chapter explained in details and step by step. The book tries to cover all aspects of image processing from data acquisition to final classification and each step covers with specific Matlab code and examples alongside with final exercises. Also, book covers all types of images and video frames as they are treatable with the same approach.

The book, uses benchmark images and database in validation and experiments for different applications of face analysis such as face detection and face extraction [13], face recognition [14], age estimation [13], facial expression recognition and facial micro expression recognition [12]. Each of these applications explained in their related chapters in details. Also, it is considered that the readers of the current book are in beginner level in the field of face analysis. So, it is decided to cover all aspects of it such as image processing [15, 16, 33], machine learning [17, 18, 19, 20], Facial Action Coding Systems (FACS) [21] and more. All chapters cover reproductive Matlab codes and ended in few examples for final test. The book does not deal with topics in mathematical an equation approach and it is mostly about codes, it's definition and related results.

The first chapter of this book, covers basic color and depth image concepts, applications, libraries, outline of chapters, a note on exercises and what this book is all about.

## 1.1    The MATLAB Libraries

Further to theorical background and explanations of MATLAB codes, each function is provided as separate m-file from different employed libraires in a separated folder. All these functions are explained in details in the Appendix section of this book. MATLAB R2020b 64-bit software is employed in all experiments and for generating all outputs. Figure 1.1 shows two color and depth images in different color spaces. Also, Figure 1.2 presents different channels of a color image in YIQ and YCbCr color spaces [22].

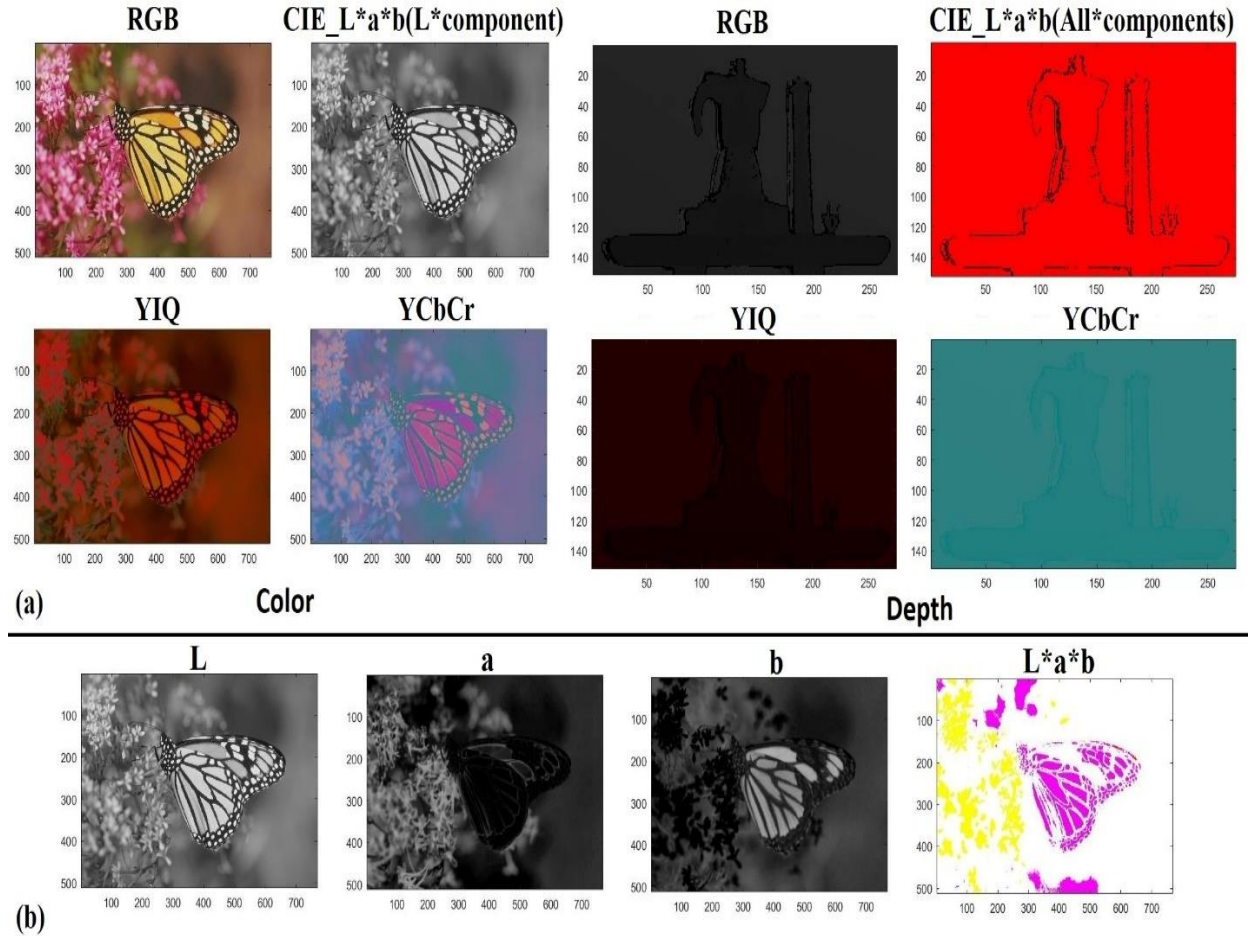

Figure 1.1 (a). Monarch color test image in RGB, CIE – Lab (L component for color data and all components for depth data), YIQ and YCbCr color spaces, and manually recorded depth test image using Kinect V.2 sensor in mentioned color spaces (top)- (b). Monarch test image in different L, a, b and all Lab components

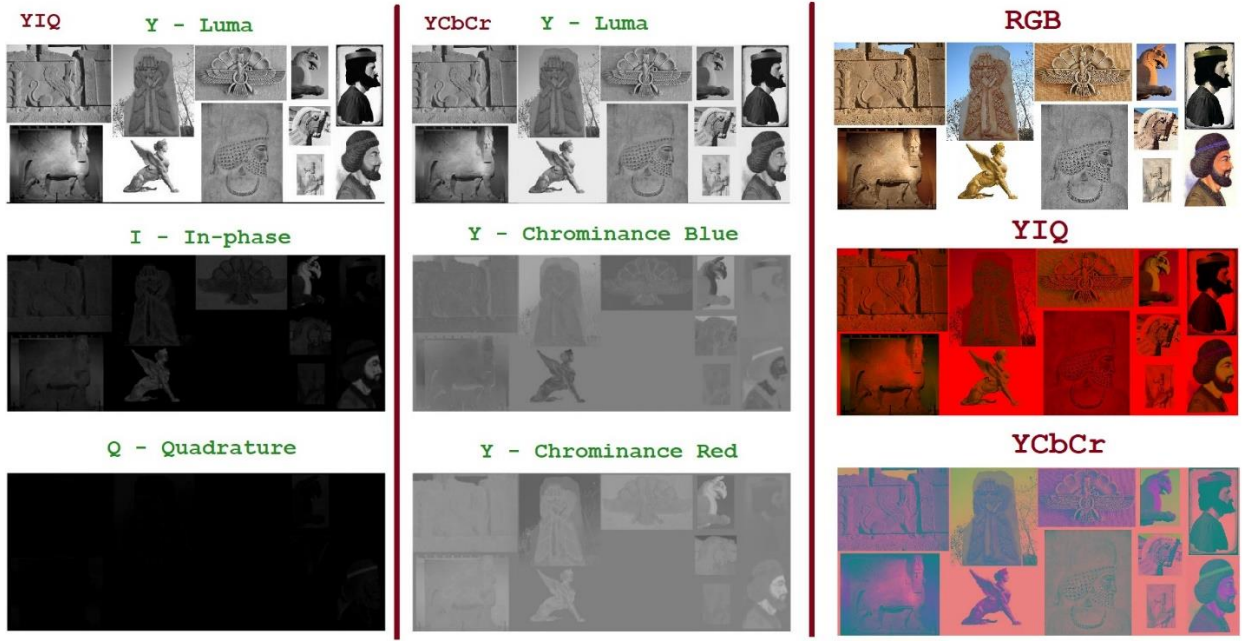

Figure 1.2 YIQ and YCbCr channels

## 1.2    Outline of Chapters

Chapter 2 is all about basic image processing concepts such as color and depth image reading and representations in different domains, image adjustments, pre-processing and depth sensors. Also, Matlab software environment described in this chapter.

Chapter 3 Basic loop, function, covers image histogram, noises, low-pass and high pass-filtering in spatial and frequency domains, working with group of images like a database edge detections and morphological operations.

Chapter 4 covers Face analysis, facial parts and muscle, Facial Action Coding System (FACS), Facial Expressions Recognitions (FER), Facial Micro Expressions Recognition (MFER), weighting facial parts, recording with Kinect, face detection and extraction in color and depth images.

Chapter 5 deals with feature detection and extraction in both spatial and frequency domain for both color and depth images or data. Also, combining the acquired features in to one feature vector for each sample and dimensionality reduction techniques explained in this chapter.

Chapter 6 pays to different classification techniques for labeled or supervised sample of classes to achieve final recognition accuracy in both train and test phases and presentation with different types of plots.

Chapter 7 stands for Neural Network Classification and Deep Learning models for depth samples of real world and benchmark IKFDB database.

Chapter 8 applies all techniques of previews chapters on a FMER database for a full experiment which covers all mentioned techniques in the book.

## 1.3    A Note on Exercises

At the end of each chapter, a set of exercises are provided. The target is to cover the chapter content and extending the interest of the reader in face analysis. Exercises are graded based on their difficulty level in four levels presented in Table 1.1. P stands for Power in Table 1.1.

**Table 1.1 Difficulty level of exercises**

| Level | Description |
|-------|-------------|
| **P 1** | Simple and basic questions which could have short answers or a little bit of coding understanding. (Take minutes to one hour to solve) |
| **P 2** | This medium level requires the understanding of the related chapter and its coding in MATLAB. (Takes hours to solve) |
| **P 3** | Medium to high level of understanding of the related chapter and requires coding in MATLAB. Reader might need to employee external references and libraries. (Takes a day to solve) |
| **P 4** | High level exercise which demands the understanding of the reader to the related chapter Coding and previews ones. Reader might need to employee external references and libraries. Exercises of this level could be used as final project. (Takes more than one day to solve). |

We encourage readers who are new to the field to read all chapters from the beginning of the book and complete all the exercises. More advanced readers can skip certain parts of the book based on their experience.

# *Chapter   2*

## *MATLAB Environment, Digital Images and Sensors and Basic Coding*

# *Chapter 2*

# *MATLAB Environment, Digital Images and Sensors and Basic Coding*

## Contents

This chapter is about fundamentals of coding in Matlab environment which is necessary for remaining chapters. Basics of reading, writing and pre-processing of the input image frames which are main source of calculation and coding. If data is prepared correctly, it is easier to work with it and it is more understandable and professional.

### 2.1 Digital Image

A digital image is a representation of a two-dimensional image as a finite set of digital values, called picture elements or pixels [15]. Pixel values typically represent gray levels, colors, heights, etc. Remember digitization implies that a digital image is an approximation of a real scene. Image processing has application in image enhancement and restoration [23], artistic effects [24], medical visualization [5], industrial inspection [8], law enforcements, human computer interaction [24] and evolutionary algorithms [25, 26, 27], expert systems [28, 88] and more. Figure 2.1 represents a pixel in a digital image.

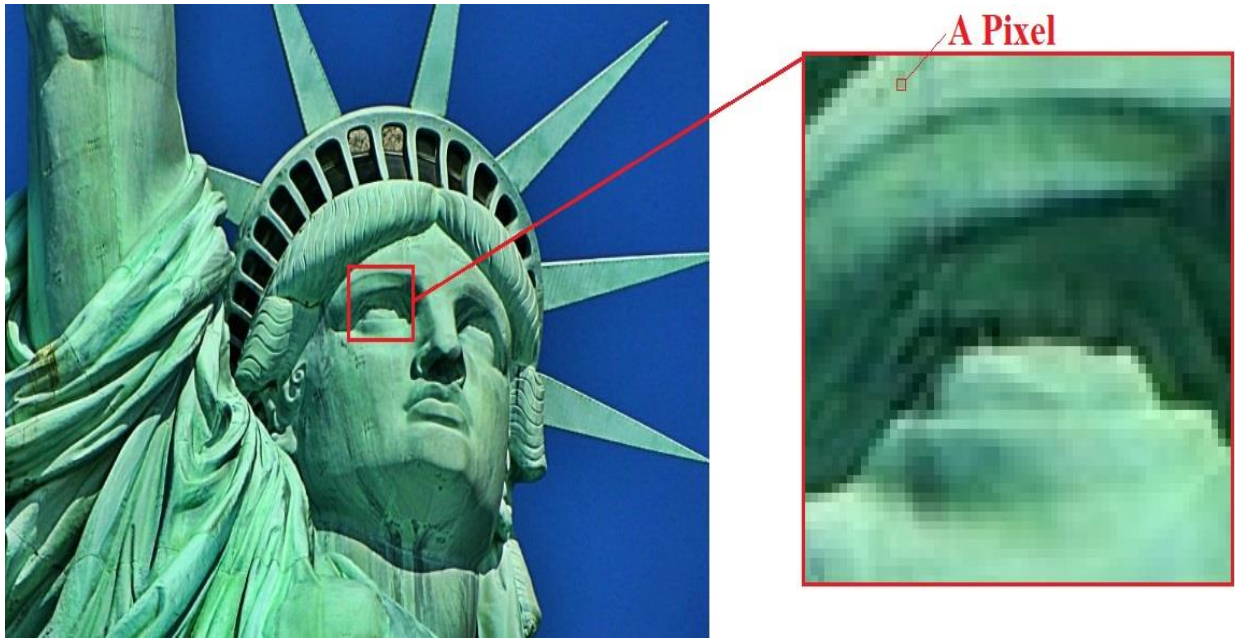

Figure 2.1 A pixel in a digital image

## 2.2 Color and Depth Images and Sensors

Traditional images that mostly consisted of Red-Green-Blue (RGB) channels known to be as color images. But there are other color spaces like CMYK, HSV, YIQ, YCbCr [22] which form the final image with different approaches, but RGB is the most common one and this book uses just the RGB. All color images use more than two channels to represent the image but if just two channels be used, then it is grey image. In the other hand, depth images are shown as a two channeled gray like image which is the output of an infrared sensor. These images could be used as night vision data but they have another application too. There are different infrared-based sensors which Kinect sensor is one of the cheapest and precis sensors among them. Second version of Kinect sensor is released by Microsoft in 2014 which is much better than first version of it. Using Kinect V.2 sensor, it is possible to acquire and record color and depth data simultaneously. Recorded depth image of Kinect sensor is a two channeled and dimensioned image which is not understandable so much by human eyes. It is a gray image which shows the distance between object and sensor by projecting infrared light into the object using emitter and receiving it by its sensor. In this type of image as pixels be blacker it means that object is closer to the sensor and more which means object is farther. Figure 2.2 shows different types of images. Depth images are known as 2.5-Dimensional (2.5-D) images, as it is possible to convert them to 3-Dimentional (3-D) models easily. It is enough to connect returned infrared particles together which it shapes a single model in a the 3-D space. Figure 3 shows Kinect V.1 VS Kinect V.2 specifications. Figure 2.4 represents an example of difference between gray and depth image values. Both are in two dimensions and in each pixel has range of 0-255, but pixels in each image have different illustrations and meanings.

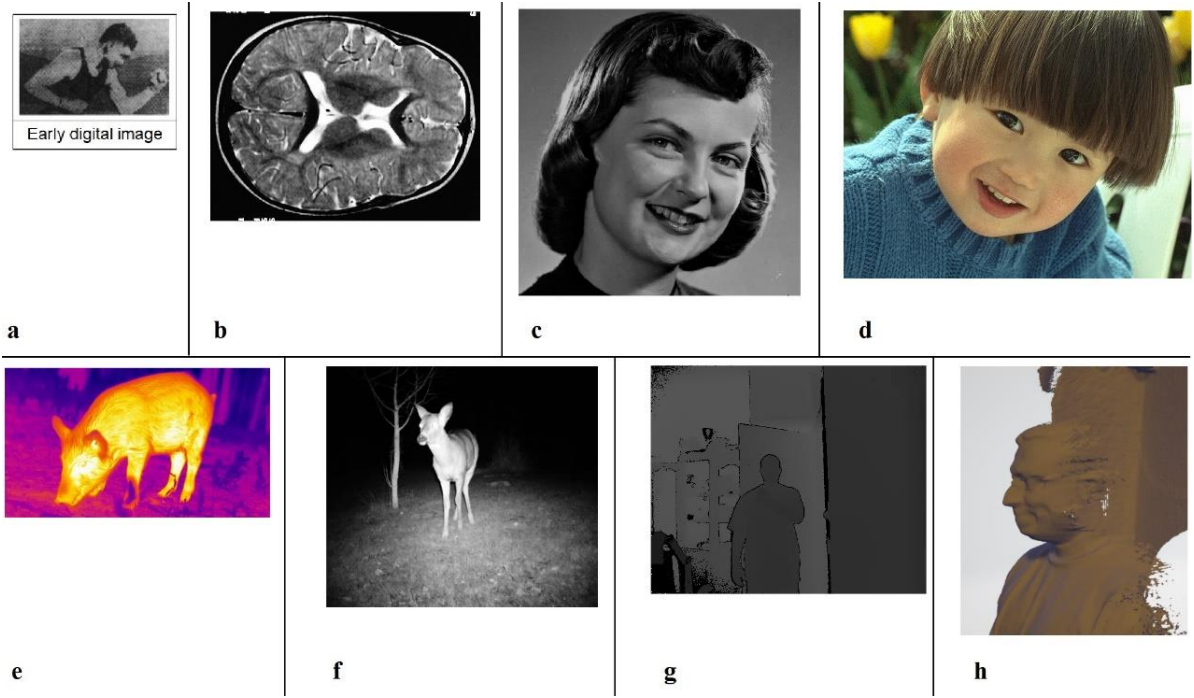

Figure 2.2 a. Early black and white image, b. modern B&W image, c. gray image, d. color RGB image, e. thermal image, f. infrared night vision image, g. infrared 2.5-D depth image, h. 3-D model of 2.5-D depth image

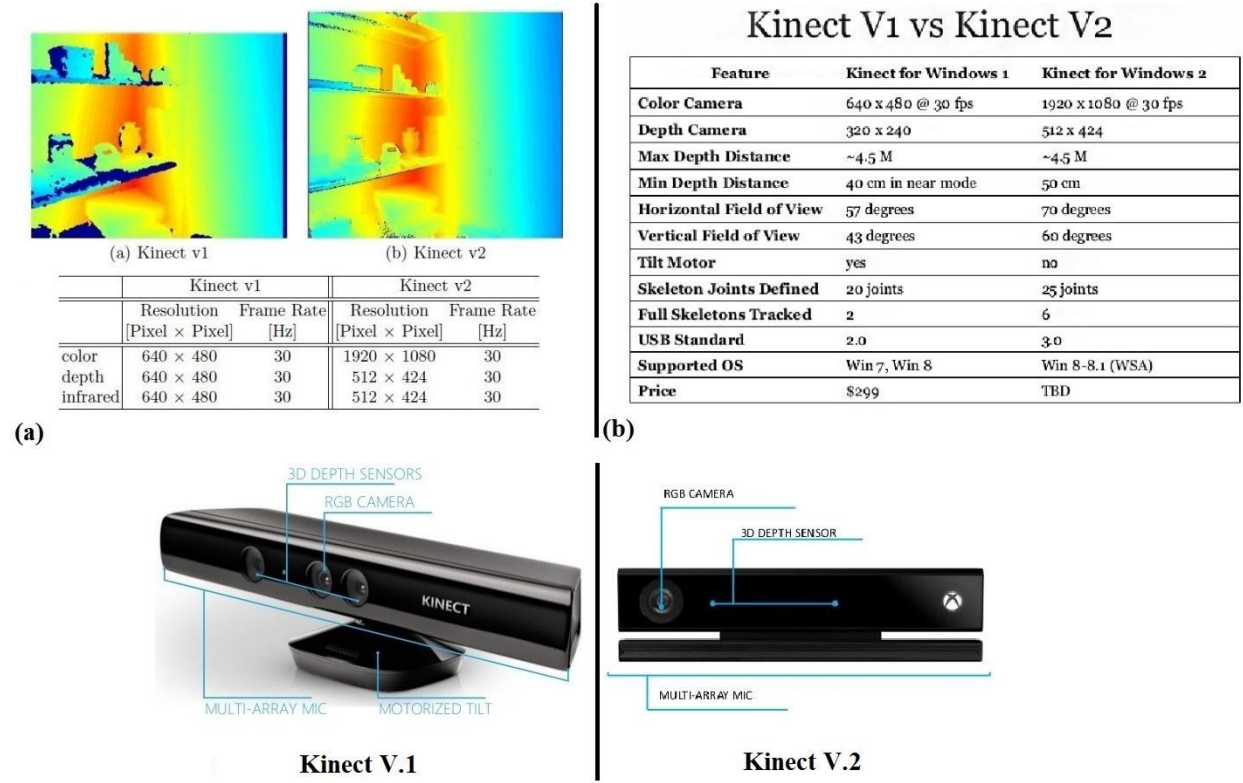

Figure 2.3 Kinect V1 VS Kinect V.2 – a. Recording dimensions, b. Specifications

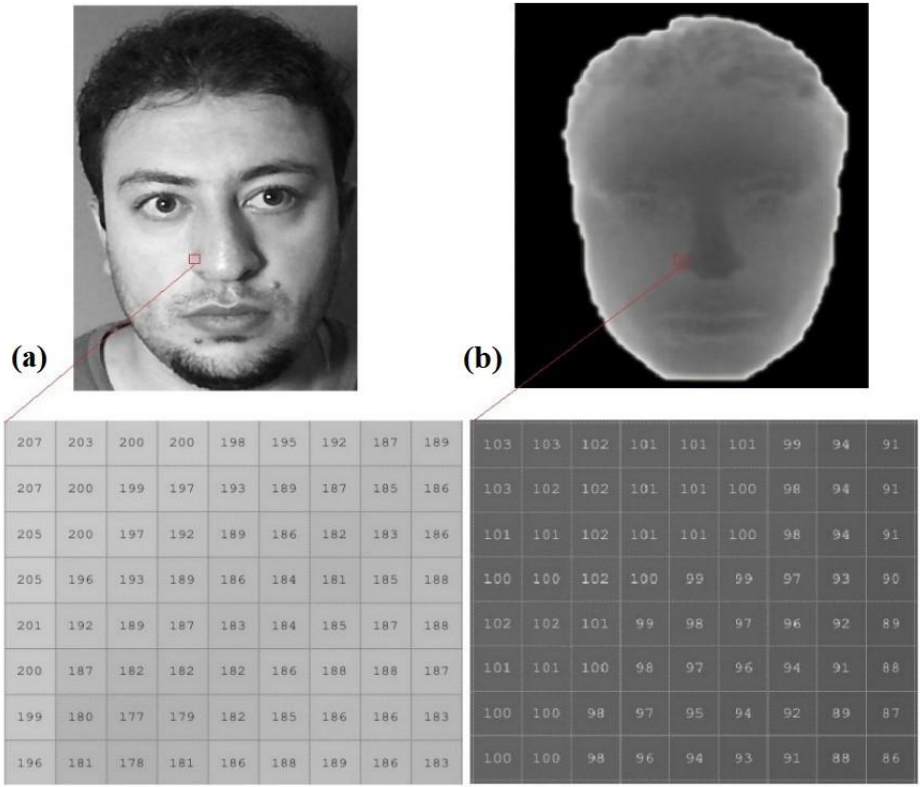

Figure 2.4 a. Gray image values (same region), b. Depth image value (same region)

### 2.2.1 Image Sampling and Quantization

Sampling is the process of converting a function of continuous time or space into a function of discrete time or space. The process is also called analog-to-digital conversion, or simply digitizing. The sampling rate determines the spatial resolution of the digitized image, while the quantization level determines the number of grey levels in the digitized image [29]. A magnitude of the sampled image is expressed as a digital value in image processing. The transition between continuous values of the image function and its digital equivalent is called quantization. Figure 2.5 represents sampling and quantization in one frame.

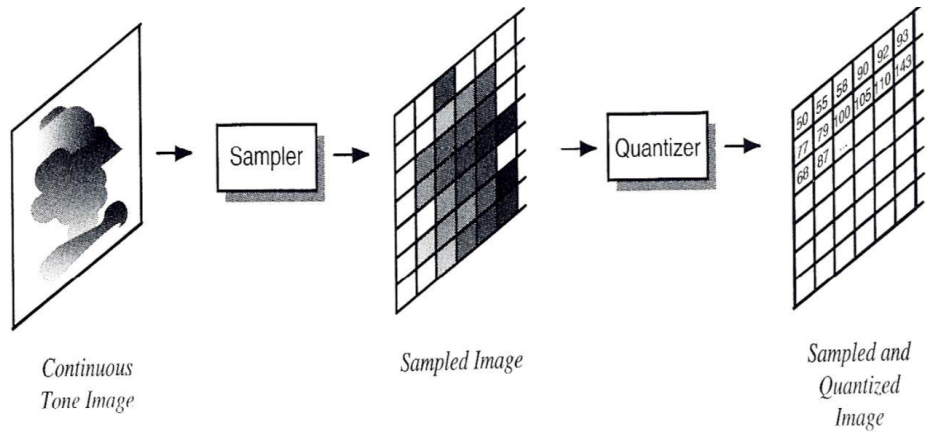

Figure 2.5 Sampling and quantization

## 2.3 MATLAB Software Environment

As it mentioned before, all coding and calculations are done by Matlab R2020b software. Matlab has a lot of libraries and functions for different engineering fields, but here just Artifactual Intelligence (AI) [30] libraries are used. Its environment did not affect any change for years. Matlab is case sensitive, which means different upper- and lower-case alphabets considered for variable's names. It is simple to understand and have unique error and debug system which makes it easier to use for beginners. Most libraries and functions are built in and has a perfect help and documentation inside the software. Also, it has easy to use graphical representations and plots which makes the outputs more understandable. MATLAB [31] allows matrix manipulations, plotting of functions and data, implementation of algorithms, creation of user interfaces, and interfacing with programs written in other languages. Figure 2.6 shows the structure of Matlab environment.

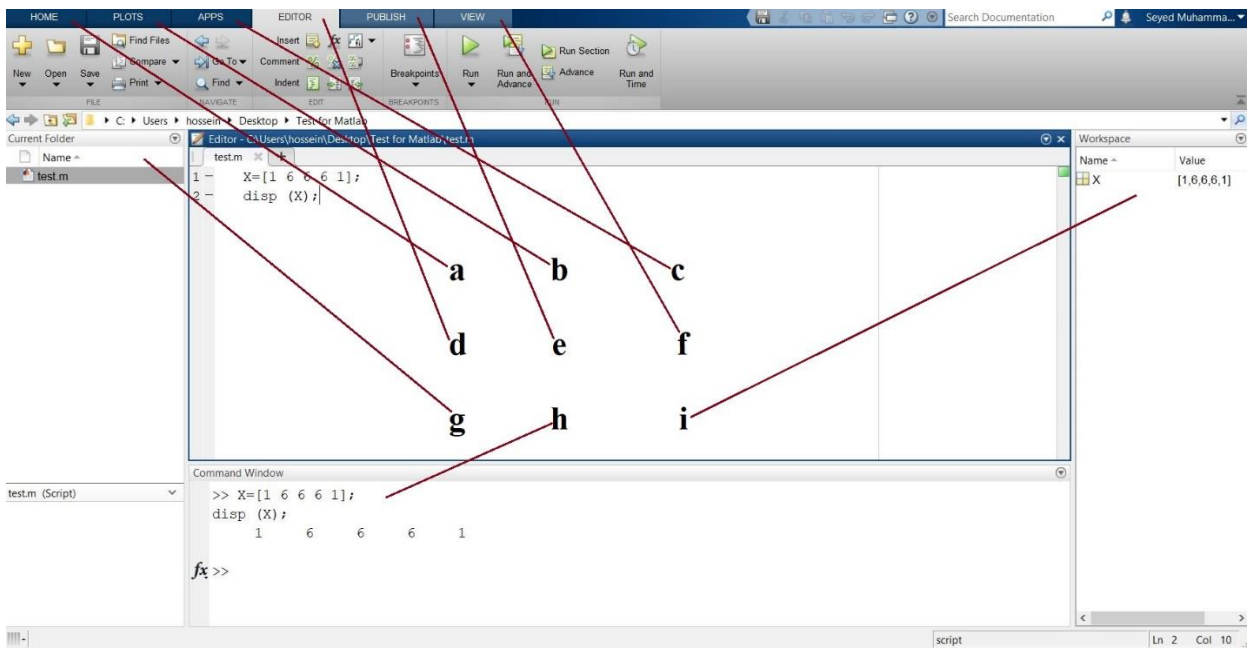

Figure 2.6 Main section which alphabets are in, is editor which you can code and make m files. a. Some basic actions, b. Different types of plots, c. Various applications for different scientific fields, d. Some actions like run, open and save, e. Internet-based actions, f. Changing Matlab environment, g. Working folder destination and files, h. All commands, errors, warnings and some outputs appear in command windows. Also, it is possible to code here, i. Matrixes and variables hold in workspace

## 2.4 Reading, Writing and Showing Image Frames in Matlab

Matlab covers almost all types of image extensions, that you can even drag and drop it into work space. Matlab files are saved with ".m" extension. In order to start coding in Matlab, it is possible to directly code in the command windows which is not rational or start a new coding page with writing "edit" in the command window. The other ways to make a new script is to click on "new Script" under the "Home" tab or simply use shortcuts of "Ctrl + N". Remember, that before start coding, it is essential to determine the destination folder which your m files going to be saved or open. You can do this by clicking on "browse for folder" icon at top of the "current folder" section or "g" in Figure 2.6. For the first code, the text of "This is a coding test" going to be show in command windows. In order to do that we need a variable. Variables could be in different types of numeric or text. Also, numeric number of "98" is stored in variable "Num" and displayed. It has to be mentioned, in order to make comments, "%" is going to be used at the beginning of desire line or in any section of that line. Comments are green. It is possible to make comments by "Ctrl + R" and uncomment by "Ctrl + T". "Clc" and "Clear" commands are suggested to be used and the beginning by any scripts as they clear the command windows and workspace and decreases confusion. In order to run the code, you can push "F5" or click on "run" (green flash icon) in the "Editor tab". If it is first time to run the code you have to save the code in the "current folder". Run the code and see the results, then check command windows and workspace windows and click on variables in workspace windows to see the variable's values both in "Num" and "Txt". This code is saved as "First_Code.m" in the book folder. Each line of code is commented with its related description.

- **Some pages of the book include boxes which contains very important and vital notes. Please make sure to read these boxes for better understanding.**

```
% Displaying variable value
% Code name : – First_code.m
clc;    % Clearing the Command Window
clear;  % Clearing the Workspace
Num = 98;   % Adding Numeric Value to Variable "Num"
Txt = 'This is a coding test';   % Adding Text Value to Variable "Txt"
disp(Num); % Showing Variable Value for "Num"
disp(Txt); % Showing Variable Value for "Txt"
```

Next code belongs to reading an image and showing it. Please be sure that current folder is your coding folder then start coding. The image is "Cat.jpg" and the related Matlab code is "c.2.1.m" in the book folder. Each line has commented for its task. The difference between "imshow" and "subimage" is that "subimage represents X and Y dimensions of the image, but "imshow" does not.

```
% Reading and showing an image
% Code name : c.2.1.m
clc;    % Clearing the Command Window
clear;  % Clearing the Workspace
cat= imread ('cat.jpg');    %Reading Input Image
imshow(cat);    % Showing the Image
title('The Cat')   % Plot Title
xlabel('X Axis')    % X Axis Title
ylabel('Y Axis')    % Y Axis Title
figure; % Creating Another Plot Windows
subimage (cat); % Showing Image with Dimensions
```

> Note:
> - *It is essential to set current folder as the main coding and reading-writing folder. Otherwise, you may find some troubles.*
> - *Each line of code has its comment, which this comment is its complete description.*
> - *All test images and codes are in the books folder.*

Following code, reads a color and a depth image from Iranian Kinect Face Database (IKFDB) [32] and separates color images to three channels of red, green and blue and shows them all using "subplot" function in one figure. Related code is "c.2.2.m". Figure 2.7 represents the code's output. Subject has surprise expression. Facial Action Coding System (FACS) [21, 32] explained completely in chapter 4 of the book.

Subplot presents multiple output at one figure as follow:

Subplot (2,3,1) % it means 2 rows and 3 columns and 1 means first output's position from top left. This figure has 2*3 =6 output places.

Imshow (first image);

Subplot (2,3,2) ;

Imshow (second image);

.

.

.

Subplot (2,3,6) ;

Imshow (sixth image);

```matlab
% Extract the individual red, green, and blue color channels
% Code name : c.2.2.m
clc;     % Clearing the Command Window
clear;  % Clearing the Workspace
RGB=imread('Exp_Sur_Color.jpg'); % Reading color image
Depth= imread('Exp_Sur_Depth.png'); % Reading depth image
colorredChannel = RGB(:, :, 1); % Separating red channel out of color data
colorgreenChannel = RGB(:, :, 2); % Separating green channel out of color data
colorblueChannel = RGB(:, :, 3); % Separating blue channel out of color data
figure('Name','Color and Depth','units','normalized','outerposition',[0 0 1 1]) % Full
screen plot
% Subplot helps to have multiple outputs at the same time
subplot(2,3,1)
subimage(RGB);title('RGB') % Plotting color image
subplot(2,3,2)
subimage(colorredChannel);title('Red') % Plotting color image red channel
subplot(2,3,3)
subimage(colorgreenChannel);title('Green') % Plotting color image green channel
subplot(2,3,4)
subimage(colorblueChannel);title('Blue') % Plotting color image blue channel
subplot(2,3,5)
subimage(Depth);title('Depth') % Plotting depth image
```

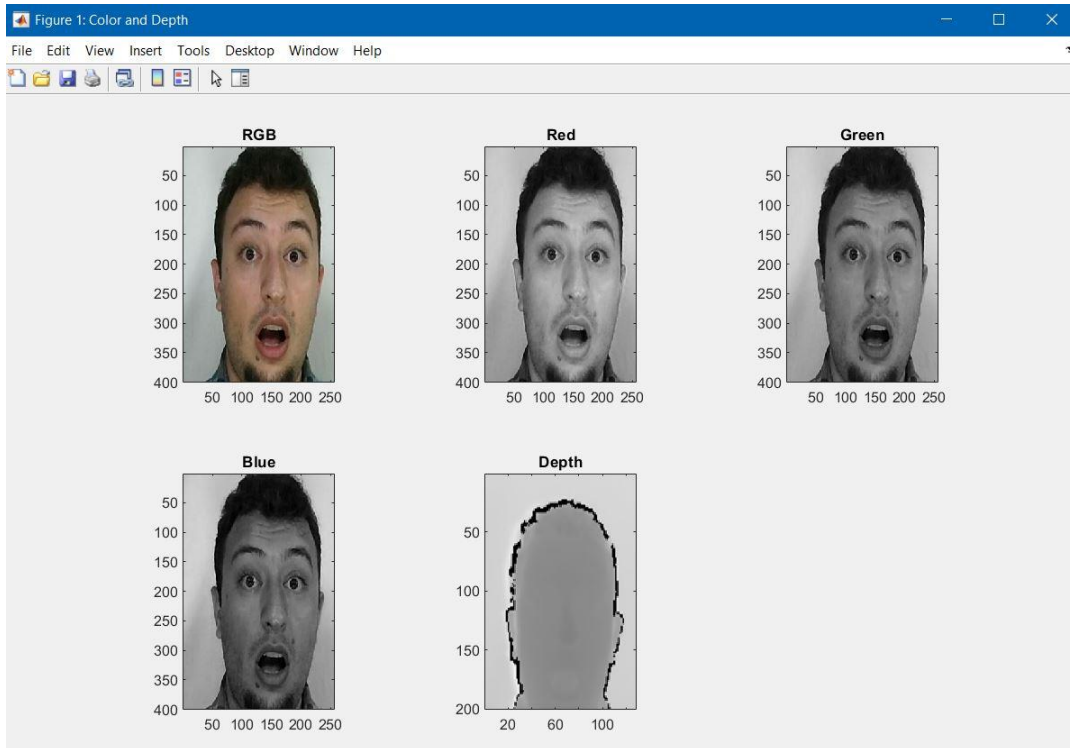

Figure 2.7 Plotting color image in separated channels and related depth image – surprise expression by a subject from IKFDB

Also, in order to write an image on hard drive, "imwrite" function could be used. As below:
```
imwrite(Image,'Write.jpg');
```
It will save the variable "Image" into current folder.

## 2.5    Spatial and Frequency Domains

In spatial domain [29], we deal with images as it is. The value of the pixels of the image change with respect to scene. Whereas in frequency domain [29], we deal with the rate at which the pixel values are changing in spatial domain. We first transform the image to its frequency distribution. Then in the black box system processing performs, and the output of the black box in this case is not an image, but a transformation. After performing inverse transformation, it is converted into an image which is then viewed in spatial domain. A signal can be converted from time or spatial domain into frequency domain using mathematical operators called transforms. Actually, the spatial domain is exactly what we see in the nature by our eyes but, images in frequency domain is not understandable by human eyes and after converting to spatial domain, it is possible to be seen.

There are many kinds of transformation that does this. Here we use Fourier transform [29] as it is the most famous one. In mathematics, a Fourier transform is a mathematical transform that decomposes functions depending on space or time into functions depending on spatial or temporal frequency, such as the expression of a musical chord in terms of the volumes and frequencies of its constituent notes. The Fourier Transform is an important image processing tool which is used to decompose an image into its sine and cosine components. The output of the transformation represents the image in the Fourier or frequency domain, while the input image is the spatial domain equivalent. Following lines of codes are for reading color and depth images from input and showing them in both spatial and frequency domains. Figure 2.8 shows two color and depth images in spatial and frequency domains. It has to be mentioned that for calculating Fourier transform, the image should be in two dimensions. So, color image is converted to gray image using "rgb2gray" function, as it has two dimensions and not data is lost for this type of calculation.

```matlab
% Color and depth images in spatial and frequency domains
% Code name: c.2.3.m
clc;      % Clearing the Command Window
clear;    % Clearing the Workspace
RGB=imread('Exp_Sur_Color.jpg'); % Reading color image
RGB=rgb2gray(RGB); % Converting color image to gray
Depth= imread('Exp_Sur_Depth.png'); % Reading depth image
FreqColor=fft2(RGB); % Two-dimensional Fourier transform of matrix
FreqDepth=fft2(Depth); % Two-dimensional Fourier transform of matrix
subplot(2,2,1)
imshow(log(1+abs(FreqColor)),[]); % Color image in frequency domain
title('Color in frequency domain');
subplot(2,2,2)
imshow(log(1+abs(FreqDepth)),[]); % Depth image in frequency domain
title('Depth in frequency domain');
subplot(2,2,3)
imshow(RGB); % Color image in spatial domain
title('Color in spatial domain');
subplot(2,2,4)
imshow(Depth); % Depth image in spatial domain
title('Depth in spatial domain');
```

### 2.5.1   Electromagnetic Spectrums

Sun projectiles different types of spectrums [29] which one of them is light or the spectrum which is visible by human eye. The electromagnetic spectrum is split up according to the wavelengths of different forms of energy. The colors that we perceive are determined by the nature of the light reflected from an object. Actually, sun spectrums spitted into seven in different frequency level. It starts with Gamma ray to X-ray to U-V to light ray or visible spectrum (human eye frequency) to Infrared ray to Micro wave ray to TV and Radio rays. Visible spectrum or light ray is in the range of 400 nanometer (nm) to 700 nm wavelength or in the fourth level of sun spectrums with 790-430 hertz (Hz) and infrared or night vision is in the range of 700 nm to 1 millimeter (mm) or in the fight level of sun spectrums with 430-30 Hz. In this book just these two spectrums are employed. Figure 2.9 shows different spectrums of sunlight in ranges.

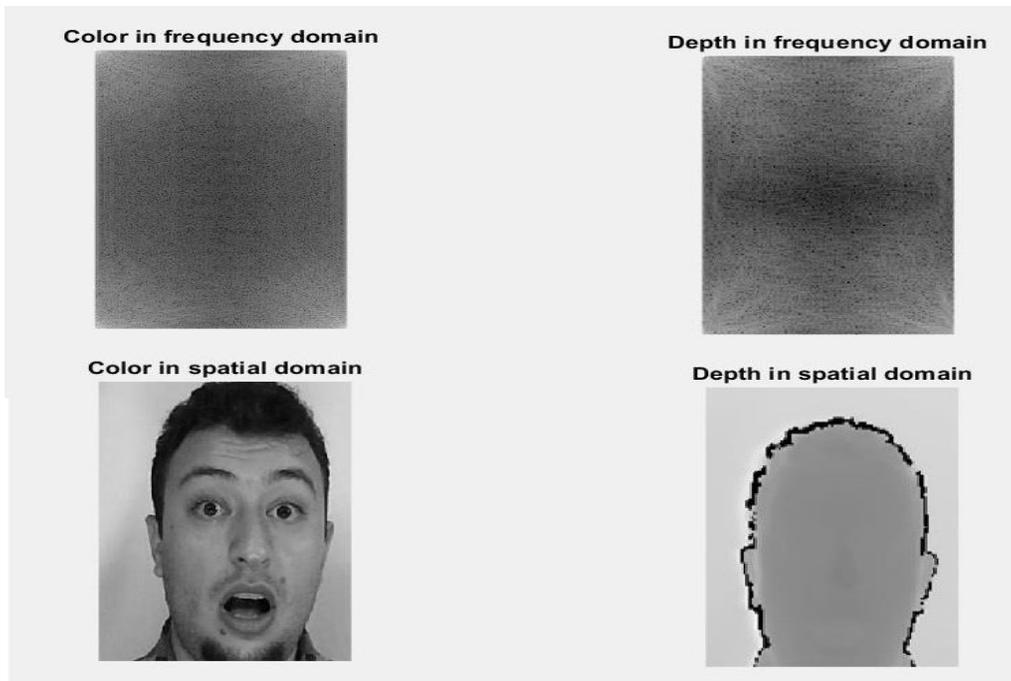

Figure 2.8 Color and depth sample images in spatial and frequency domains

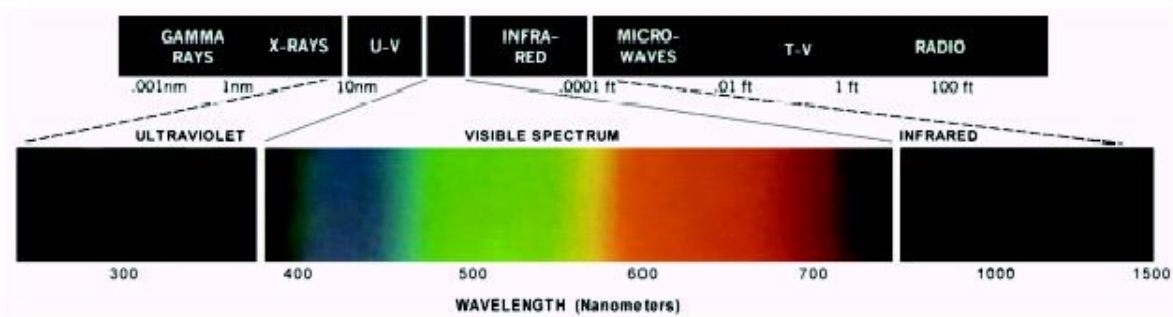

Figure 2.9 Different spectrums of sun light [29]

## 2.6    Pre-Processing

Pre-Processing is an important stage in image processing, as it is possible to get rid of any noise and outliers with it before starting main processing. Here just some of the most important image enhancements are explained and next chapter covers all of the low and high pass filters.

### 2.6.1    Basic Color and Depth Image Enhancements

If the image is polluted with gaussian or impulsive noises [23, 16] or has uncorrected lightning scene or it has unwanted parts, it is vital to get rid of them and restore them before starting the main process or it makes some problems like wrong detection and recognition rate. Most of the color and depth images need to be enhanced and restores by few actions such as gray and black and white conversion, median filtering [23, 16], unsharp masking [23, 16], histogram equalization [23, 16], contrast adjustment, cropping and resizing [32]. Following code does the mentioned basic-preprocessing on a color and a depth image in spatial domain. It has to mentions that median filtering is proper to remove salt and pepper noises [23, 16] and blurring image. Histogram equalization is a method of contrast adjustment using the image's histogram. Through this adjustment, the intensities can be better distributed on the histogram. This allows for areas of lower local contrast to gain a higher contrast. Histogram equalization accomplishes this by effectively spreading out the most frequent intensity values. Unsharp masking is an image sharpening technique, often available in digital image processing software. Its name derives from the fact that the technique uses a blurred, or "unsharp", negative image to create a mask of the original image. Figure 2.10 and 2.11 illustrated the effect of these pre-processing on two color and depth images.

```matlab
% Basic Pre-processing in spatial domain
% Code name : c.2.4.m
clc;     % Clearing the Command Window
clear;   % Clearing the Workspace
RGB=imread('Cpre.jpg'); % Reading color image
RGB=rgb2gray(RGB); % Converting color image to gray
Depth= imread('Dpre.png'); % Reading depth image
RGB = imresize(RGB, [128 128]); % Resizing color image to 128*128 dimensions
Depth = imresize(Depth, [128 128]); % Resizing depth image to 128*128 dimensions
Cbw=im2bw(RGB); % Black and white version of color image
Dbw=im2bw(Depth); % Black and white version of depth image
Cadj=imadjust(RGB); % Contrast adjustment of color image
Dadj=imadjust(Depth); % Contrast adjustment of depth image
Chist = histeq(RGB); % Histogram equalization of color image
Dhist = histeq(Depth); % Histogram equalization of depth image
Cmed = medfilt2(RGB,[2 2]); % Median filter for color image
Dmed = medfilt2(Depth,[2 2]); % Median filter for depth image
Csha = imsharpen(RGB,'Radius',5,'Amount',0.5); % Sharpening the edges of color image
Dsha = imsharpen(Depth,'Radius',5,'Amount',0.5); % Sharpening the edges of depth image
subplot(2,3,1);subimage(RGB);title('Original-Gray');
subplot(2,3,2);subimage(Cbw);title('BW');
subplot(2,3,3);subimage(Cadj);title('Contrast Adjusted');
subplot(2,3,4);subimage(Chist);title('Histogram Equalization');
subplot(2,3,5);subimage(Cmed);title('Median Filter');
subplot(2,3,6);subimage(Csha);title('Sharpening');
figure; % New figure for depth data
subplot(2,3,1);subimage(Depth);title('Original-Gray');
subplot(2,3,2);subimage(Dbw);title('BW');
subplot(2,3,3);subimage(Dadj);title('Contrast Adjusted');
subplot(2,3,4);subimage(Dhist);title('Histogram Equalization');
subplot(2,3,5);subimage(Dmed);title('Median Filter');
subplot(2,3,6);subimage(Dsha);title('Sharpening');
```

*Note:*
- *In Matlab, it is possible to separate each line of code with semicolon. It means you can write two and more commands in a single line.*
- *Color images convert to gray image in order to work easier on them and there is no vital data lose.*
- *Using two "%", you can make a section in Matlab editor*
- *Selecting and highlighting the portion of code in the editor and pushing "f9", runs and compiles that portion only.*

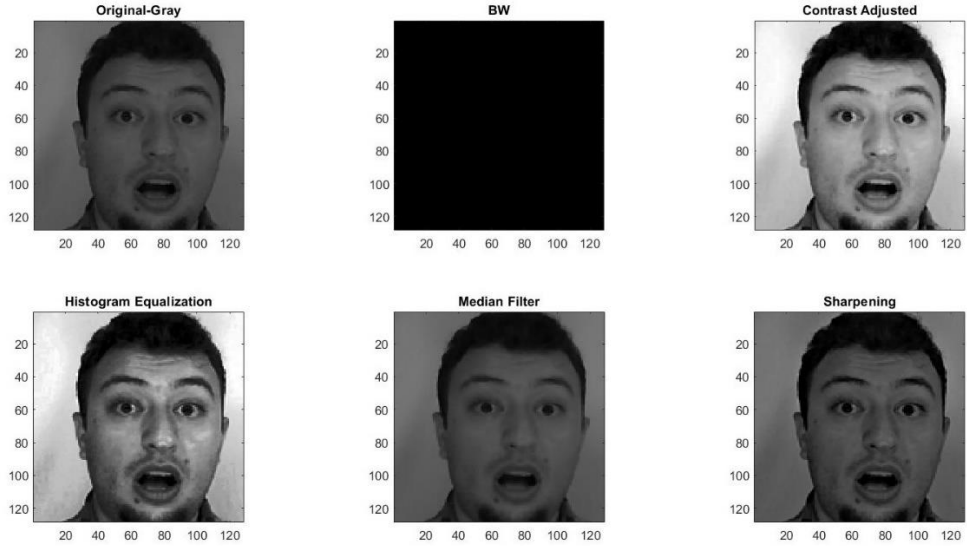

Figure 2.10 effect of basic pre-processing in spatial domain (color data)

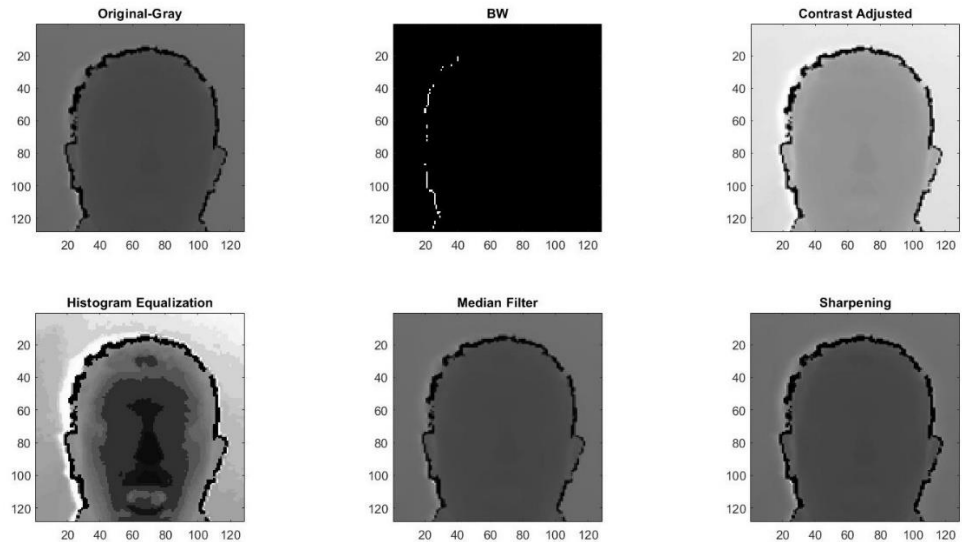

Figure 2.11 effect of basic pre-processing in spatial domain (depth data)

## 2.7 Exercises

1. (P1): a). What is the different between color and depth images and how color and depth sensors works? (in relation of spectrums and working frequency range) - b). What is the difference between color and depth pixel values?

2. (P2): Write a code that separates color image's channels and adjust the contrast of the green channel and show it by its dimensions.

3. (P2): Write a code that takes a depth image as input and pre-process it as follow: - histogram equalization – sharpening – median filter by order. Note: each task must be done on previews task. It means task sharpening should be done on histogram equalization output, not on original image and show them all in one figure.

# *Chapter 3*

# *Mid–Level Coding and More Essential Functions*

# *Chapter 3*

# *Mid–Level Coding and More Essential Functions*

## Contents

This chapter deals with some of the main functions and remaining pre-processing operations alongside with processing a group of images with explained techniques. Main loop structures and how to make a custom function is explained at the beginning of the chapter. Noises changes the histogram of the image, so main types of noises explained and how to remove them, alongside with their related histogram before and after removing noises. Low and high pass filters for spatial and frequency domains and for color and depth images are explained to understand which filter works better for each type of image. There are some types of task which handled better in frequency better and it is suggested rather than using spatial domain filters. Edge detection is one of the most important aspects of image processing which even could be employed in depth images. So, main image processing techniques are explained with examples. Morphological operation helps to get rid of some noises and bugs in the images which some main morphological operations are described in this chapter. For using these techniques in a real-world data, it is needed to deal with batch or group of files not just a single test file. So, image batch processing is explained with number of images in an efficient and easy way. Please read "Notes" boxes in the entire book as they are so helpful to understand the details.

## 3.1    Loops and Conditional Statements

### 3.1.1 Loops

With loop control statements, you can repeatedly execute a block of code. There are two types of loops: "for" and "while".

- "for" statements loop a specific number of times, and keep track of each iteration with an incrementing index variable. You can simply copy and paste the codes and run it into your Matlab editor with "f5", "f9" and other executing ways and see the results in the workspace or command windows.

  For example, pre-allocate a 10-element vector, and calculate five values:

  ```
  x = ones (1,10);
  for n = 2:6
      x(n) = 2 * x (n - 1);
  end
  ```

- "while" statements loop as long as a condition remains true.

  For example, find the first integer n for which factorial(n) is a 100-digit number:

  ```
  n = 1;
  nFactorial = 1;
  while nFactorial < 1e100
      n = n + 1;
      nFactorial = nFactorial * n;
  end
  ```

### 3.1.2 Conditional Statements

There are two main conditional statements of "if" and "switch" in Matlab.

- "if" expression, statements, end evaluates an expression, and executes a group of statements when the expression is true. An expression is true when its result is nonempty and contains only nonzero elements (logical or real numeric). Otherwise, the expression is false. For example, loop through the matrix and assign each element a new value. Assign 2 on the main diagonal, -1 on the adjacent diagonals, and 0 everywhere else. Also, you can combine loops and conditional statements as below:

  ```
  nrows = 4;
  ncols = 6;
  A = ones(nrows,ncols);
  for c = 1:ncols
      for r = 1:nrows
          if r == c
              A(r,c) = 2;
          elseif abs(r-c) == 1
              A(r,c) = -1;
          else
              A(r,c) = 0;
  ```

```
        end
      end
    end
    A
```

- Switch evaluates an expression and chooses to execute one of several groups of statements. Each choice is a case. The switch block tests each case until one of the case expressions is true. Display different text conditionally, depending on a value entered at the command prompt.

```
n = input('Enter a number: ');
switch n
    case -1
        disp('negative one')
    case 0
        disp('zero')
    case 1
        disp('positive one')
    otherwise
        disp('other value')
end
```

---

*Note:*
- *"input" command takes an input from keyboard and acts according of the user code for that input which is stored in the memory.*
- *"ones" command creates a matrix with specific number of "true" or "1" values. Each matrix is variable with more than 1 cell. "zeros" acts inverse of "ones".*
- *"break" command or functions, terminates the execution of loops where it calls.*

---

Following code, takes a depth image from input and replaces a gray value for black spots which acquires in recording process by Kinect sensor. Figure 3.1 shows the effect of the following code.

```
% Removing black spots
% Code name : c.3.1.m
clc;       % Clearing the Command Window
clear;     % Clearing the Workspace
Depth= imread('Dpre.png'); % Reading depth image
DepthTemp=Depth; % Making a clone from Depth image
[d1,d2] = size(DepthTemp); % Extracting numbers of rows and columns of the
image
for i=1:d1 % Loop for row navigation
    for j=1:d2 % Loop for column navigation
        if DepthTemp(i,j)<30 % Conditional statement for black spots threshold
            DepthTemp(i,j)=70; % Replacing value for black spots
        else
            DepthTemp(i,j)=DepthTemp(i,j); % Do not affect any change
        end
    end
end
subplot(1,2,1)
imshow(Depth);title('Original');
```

```
subplot(1,2,2);
imshow(DepthTemp);title('Modified');
```

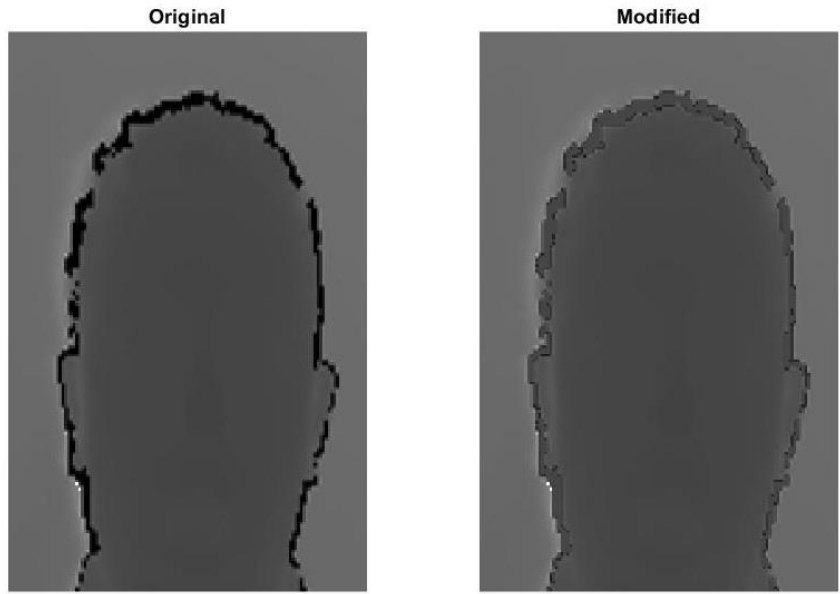

Figure 3.1 Removing black spots from a depth image

## 3.2     Functions

When it is needed to run multiple lines of codes without re coding, functions are employed. It is possible to call multiple lines of code using a single line by calling function's name in the editor or command windows. In professional coding and in dealing with real world data like databases, using functions are vital and almost impossible to make a full project without functions. In Matlab, and after writing the code, it must be saved as a "m file" in the current folder. Actually, all pre-ready Matlab commands such as "ones", "imshow"," subplot" and more are functions and user calls and employs them. For example, define a function in a file named "average.m" that accepts an input vector, calculates the average of the values, and returns a single result.

```
function ave = average(x)
    ave = sum(x(:))/numel(x);
end
```

Call the function from the command line.

```
z = 1:99;
ave = average(z)
```

Define a function in a file named "stat.m" that returns the mean and standard deviation of an input vector.

```
function [m,s] = stat(x)
    n = length(x);
    m = sum(x)/n;
    s = sqrt(sum((x-m).^2/n));
end
```

Call the function from the command line.

values = [12.7, 45.4, 98.9, 26.6, 53.1];

[ave,stdev] = stat(values)

 The following lines of codes, makes a function called "fun" and it has one input of "pic" and two outputs of "pic" and "pica" which first converts color image to gray then sharpening it and then applies median filter to sharpened image and then adjusts the contrast of the filtered image. Finally, plots all changes in a row and returns two outputs of "pic" and "pica" which are gray and final contrast adjusted image into the workspace as variables. Sample image is belonging to IKFDB [32, 36] database having happy expression. Figure 3.2 shows the effect of calling this function.

```matlab
% Simple function definition
% Code name : "fun.m"
function [pic,pica]=fun(pic) % Function "fun" with one input and two outputs
pic=rgb2gray(pic); % Gray image
pics = imsharpen(pic,'Radius',5,'Amount',0.5); % Sharpened image
picw = medfilt2(pics,[5 5]); % Median filter
pica = imadjust(picw); % Contrast adjusted
figure('units','normalized','outerposition',[0 0 1 1])
subplot(1,4,1),imshow(pic);title('Original');
subplot(1,4,2),imshow(pics);title('Sharped');
subplot(1,4,3),imshow(picw);title('Median Filtered');
subplot(1,4,4),imshow(pica);title('Contrast Adjusted');
end
```

 After writing functions and saving it as "m file" it should be called to take effect using following lines of code into the editor or command windows:

```matlab
% Calling the function
% Code name : "calling fun function.m"
clc;     % Clearing the Command Window
clear;   % Clearing the Workspace
RGB=imread('Exp_Hap_Color.jpg'); % Reading color image
[pic,pica]=fun(RGB) % Calling "fun" function (sending one input and receiving two outputs)
```

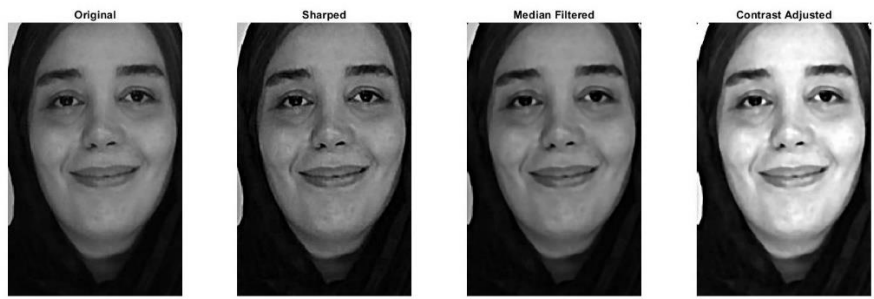

Figure 3.2 Result of calling "fun" function

---

*Note:*
- *If ";" does not been used at the end of code line and after running, the variable's values or process appears in the command windows.*

### 3.3    Image Histogram and Noises

An image histogram [15] is a chart that shows the distribution of intensities in an indexed or grayscale image. The "imhist" function creates a histogram plot by defining n equally spaced bins, each representing a range of data values, and then calculating the number of pixels within each range. You can use the information in a histogram to choose an appropriate enhancement operation. For example, if an image histogram shows that the range of intensity values is small, you can use an intensity adjustment function to spread the values across a wider range.

Image noise is random variation of brightness or color information in images, and is usually an aspect of electronic noise. It can be produced by the image sensor and circuitry of a scanner or digital camera. Image noise can also originate in film grain and in the unavoidable shot noise of an ideal photon detector. There are different types of noises, such as gaussian noise [23], salt and pepper (S&P) or impulse noise [23], shot noise or Poisson noise [23], speckle noise [23] and more. Each of these noises could be restored by some percentages and not 100%. For example, gaussian noise could be fixed by low pass averaging or blurring filter or salt and pepper noise could be fixed by median filtering. Following code represents a depth image having four types of noises and their related image histogram. Also, impulse or S&P polluted image is restored using median filter which its histogram is represented too. Figure 3.3 shows the result of the below code.

```matlab
% Noise and image histogram
% Code name : "c.3.2.m"
clc;      % Clearing the Command Window
clear;    % Clearing the Workspace
Depth= imread('Exp_Hap_Depth.png'); % Reading depth image
Gau = imnoise(Depth,'gaussian'); % Adding gaussian noise
SP = imnoise(Depth,'salt & pepper', 0.1); % Adding impulse noise
Poi = imnoise(Depth,'poisson'); % Adding Poisson noise
Spe = imnoise(Depth,'speckle'); % Adding speckle noise
SPmed = imadjust(medfilt2(SP,[3 3])); % Contrast adjustment for impulse
polluted image
subplot(2,3,1);imshow(Depth); title ('Original');
subplot(2,3,2);imshow(Gau); title ('Gaussian noise');
subplot(2,3,3);imshow(SP); title ('Salt & pepper noise');
subplot(2,3,4);imshow(Poi); title ('Poisson noise');
subplot(2,3,5);imshow(Spe); title ('Speckle noise');
subplot(2,3,6);imshow(SPmed); title ('Removing S&P noise');
figure;
subplot(2,3,1);imhist(Depth); title ('Original');
subplot(2,3,2);imhist(Gau); title ('Gaussian noise');
subplot(2,3,3);imhist(SP,256); title ('Salt & pepper noise');
subplot(2,3,4);imhist(Poi); title ('Poisson noise');
subplot(2,3,5);imhist(Spe); title ('Speckle noise');
subplot(2,3,6);imhist(SPmed,256); title ('Removing S&P noise');
```

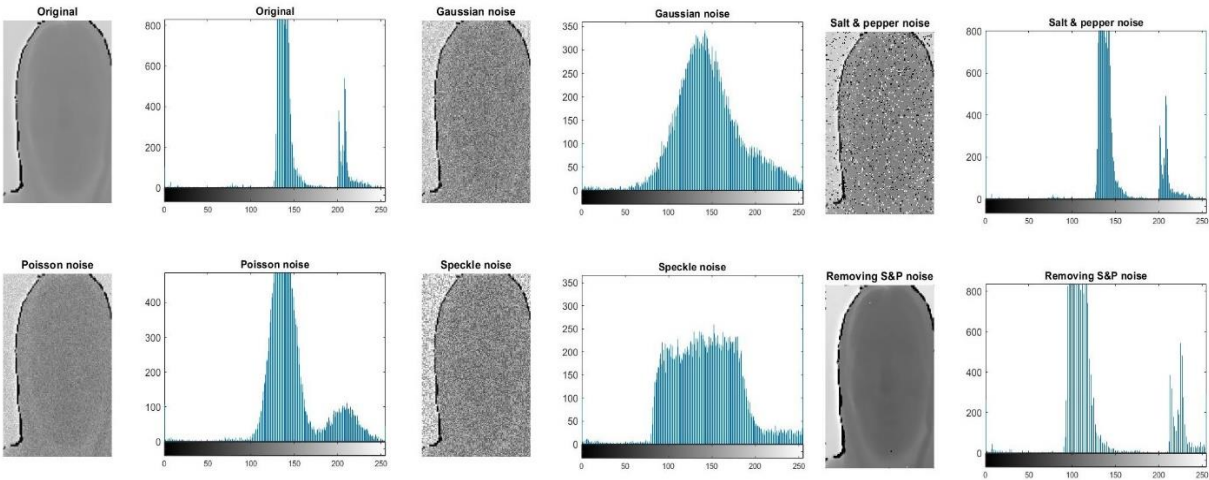

Figure 3.3 Adding noises on depth data and their histograms

## 3.4 Low and High Pass Filtering in Spatial and Frequency Domains

Blurring and sharpening [34, 29] are two most important filters in image processing as they could remove some noises and reveal more edges. These techniques could be done in both spatial and frequency domains and basically are parts of pre-processing stage. Figure 3.4 shows the different filters wave from. Also, Figure 3.5 represents some of the low and high pass filters in the spatial domain.

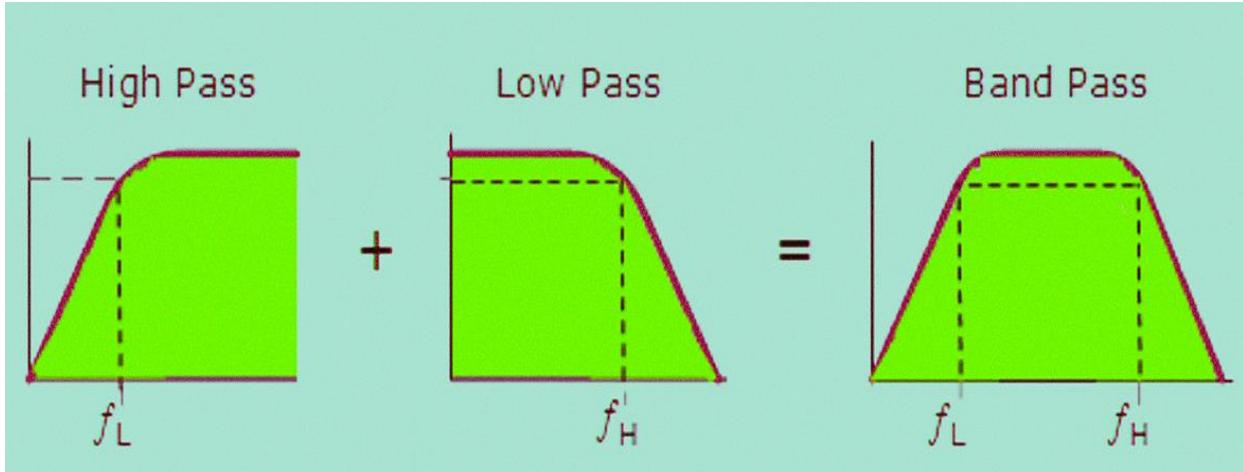

Figure 3.4 Low, high and band pass wave forms

Following code, runs some of the low and high pass filters in the spatial domain such as sharp filter, adaptive filter, gaussian filter, gradient magnitude and direction filter and standard deviation filter. Also, sample image is in sad expression from KDEF dataset [35].

```
% Spatial low and high pass
% Code name : "c.3.3.m"
clc;     % Clearing the Command Window
clear;   % Clearing the Workspace
RGB=imread('KDEF_Sad.jpg'); % Reading color image
MRGB=rgb2gray(RGB); % Converting color image to gray
```

```
MRGB=imadjust(MRGB); %Contrast adjustment
highpass = imsharpen(MRGB,'Radius',2,'Amount',1);% Sharpening image
weinlow = wiener2(MRGB,[5 5]);% Remove Noise By Adaptive Filtering
% Filter the image with a Gaussian filter with standard deviation of 2.
gaulow = imgaussfilt(MRGB,2);% 2-D Gaussian filtering of images
[Gmag, Gdir] = imgradient(MRGB,'prewitt'); %Calculate Gradient Magnitude and
Gradient Direction using "prewitt edge detection operator'
std = stdfilt(MRGB); % Local standard deviation of image
subplot(2,4,1);subimage(RGB);title('Original');
subplot(2,4,2);subimage(MRGB);title('Gray-Adjusted');
subplot(2,4,3);subimage(highpass);title('Sharpened');
subplot(2,4,4);subimage(weinlow);title('Adaptive Filtering');
subplot(2,4,5);subimage(gaulow);title('Gaussian filtering');
subplot(2,4,6);imshow(Gmag,[]);title('Gradient Magnitude');
subplot(2,4,7);imshow(Gdir,[]);title('Gradient Direction');
subplot(2,4,8);imshow(std,[]);title('Local standard deviation');
```

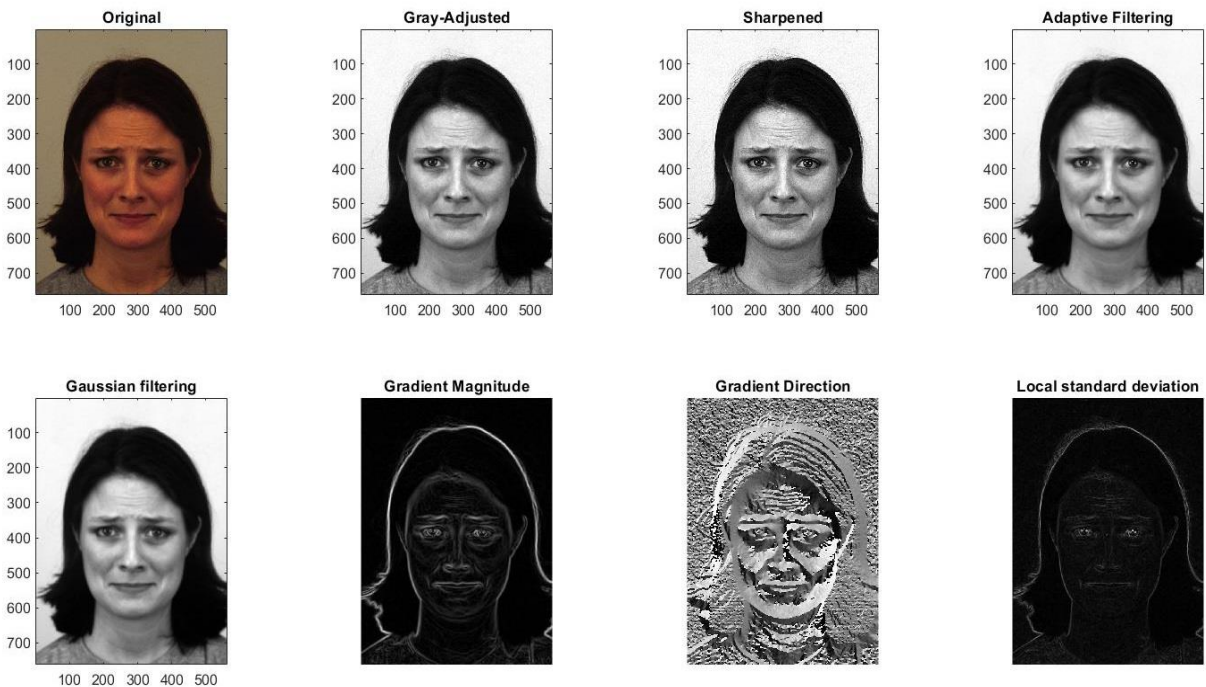

Figure 3.5 Some of the low and high pass filters in the spatial domain

In order to present low and high pass filters in frequency domain, just Butterworth [37] filter is employed which with changing parameters it could be in both passes. Following contains a function called "butterworthbpf" and it's calling code of "c.3.4.m'. Image sends to frequency domain using Fourier transform [29] and takes effect of the filter and returns to spatial domain. Figure 3.6 shows the effect of the following lines of codes.

```
% Frequency low and high pass
% Code name : "butterworthbpf.m"
function filtered_image = butterworthbpf(I,d0,d1,n)
% Butterworth Bandpass Filter
%    I = The input grey scale image
%    d0 = Lower cut off frequency
%    d1 = Higher cut off frequency
```

```matlab
%   n = order of the filter
% The function makes use of the simple principle that a bandpass filter
% can be obtained by multiplying a lowpass filter with a high pass filter
% where the lowpass filter has a higher cut off frequency than the high pass
filter.
f = double(I);
[nx ny] = size(f);
f = uint8(f);
fftI = fft2(f,2*nx-1,2*ny-1);
fftI = fftshift(fftI);

subplot(1,4,1)
imshow(f,[]);
title('Original Image')
subplot(1,4,2)
imshow(log(1+abs(fftI)),[]);
title('Image in Fourier Domain')
% Initialize filter.
filter1 = ones(2*nx-1,2*ny-1);
filter2 = ones(2*nx-1,2*ny-1);
filter3 = ones(2*nx-1,2*ny-1);
for i = 1:2*nx-1
    for j =1:2*ny-1
        dist = ((i-(nx+1))^2 + (j-(ny+1))^2)^.5;
        % Create Butterworth filter.
        filter1(i,j) = 1/(1 + (dist/d1)^(2*n));
        filter2(i,j) = 1/(1 + (dist/d0)^(2*n));
        filter3(i,j)= 1.0 - filter2(i,j);
        filter3(i,j) = filter1(i,j).*filter3(i,j);
    end
end
% Update image with passed frequencies.
filtered_image = fftI + filter3.*fftI;
subplot(1,4,3)
imshow(log(1+abs(filter3)),[]);
title('Filter Image')
filtered_image = ifftshift(filtered_image);
filtered_image = ifft2(filtered_image,2*nx-1,2*ny-1);
filtered_image = real(filtered_image(1:nx,1:ny));
filtered_image = uint8(filtered_image);
subplot(1,4,4)
imshow(filtered_image,[])
title('Filtered Image')
```

Calling with:
```matlab
% Calling "butterworthbpf.m" function
% Code name : "c.3.4.m"
%   d0 = Lower cut off frequency
%   d1 = Higher cut off frequency
%   n = order of the filter
ima = imread('KDEF_Sad.jpg');
ima = rgb2gray(ima);
% Low pass
filtered_image = butterworthbpf(ima,2,10,2);
title ('Frequency domain low-pass');
figure;
```

```
% High pass
filtered_image = butterworthbpf(ima,3,300,5);
title ('Frequency domain low-pass');
```

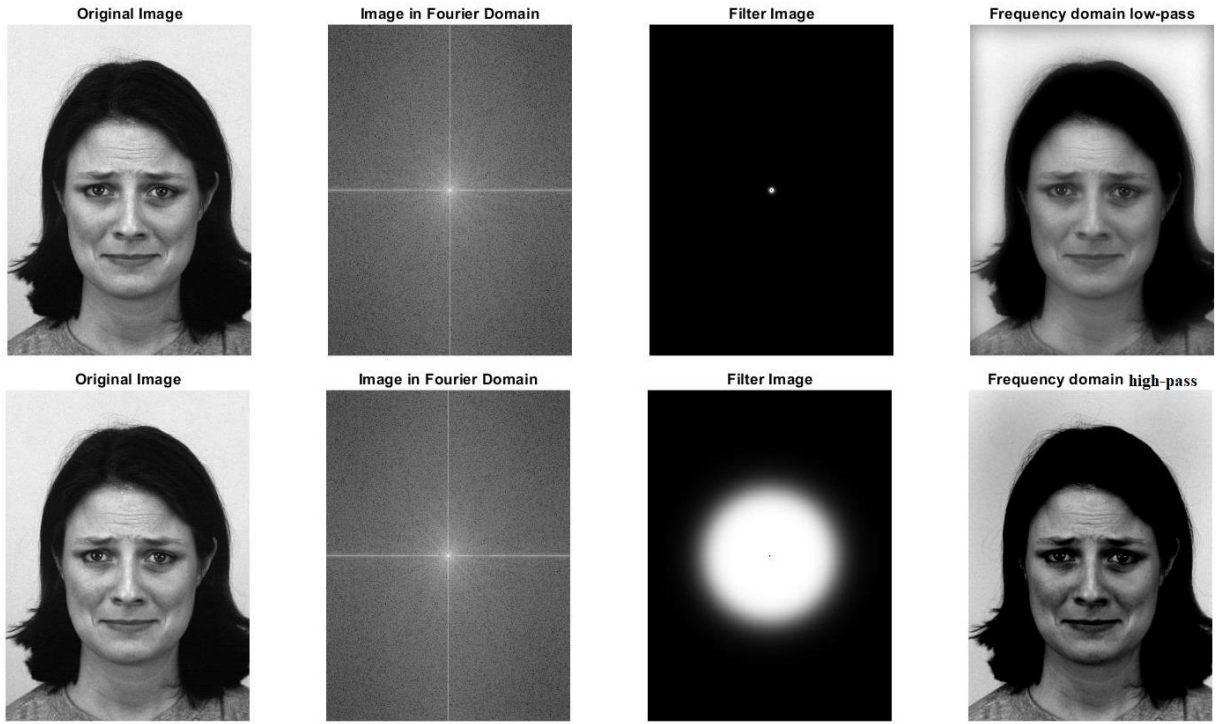

Figure 3.6 Butterworth low and high pass filters

### 3.5 Edge Detection

Feature extraction is one the most factors in image processing and machine learning. Features and especially facial features are extracted based on edges in the image. Edge detection techniques [23, 33] aid feature extraction methods to getting rid of outliers and unnecessary data which might be considered as features wrongly. There are different types of edge detection methods which some of the are presented here for color and depth images. Basically, any sudden change in gray value of pixels with their neighbors could be considered as an edge. Totally these pixels make edges. Following piece of code take a color and a depth image from IKFDB database and applies seven spatial domain edge detection techniques of Sobel [23, 33], Prewitt [23, 33], Roberts [23, 33], Logarithm [23, 33], Zero crossing rate [23, 33], Canny and Approximated canny [38]. Figures 3.7 and 3.8 represent different edge detection method's results for two color and depth images from IKFDB database.

```
% Edge Detection
% Code name : "c.3.5.m"
clc;    % Clearing the Command Window
clear;  % Clearing the Workspace
RGB=imread('Exp_Nut_Color.jpg'); % Reading color image
RGB=imadjust(rgb2gray(RGB)); % Converting color image to gray
Depth= imread('Exp_Nut_Depth.png'); % Reading depth image
RGB = imresize(RGB, [256 256]); % Resizing color image to 256*256 dimensions
Depth = imresize(Depth, [256 256]); % Resizing depth image to 256*256
dimensions
```

```
Sob = edge(RGB,'Sobel');SobD = edge(Depth,'Sobel');
Pre = edge(RGB,'Prewitt');PreD = edge(Depth,'Prewitt');
Rob = edge(RGB,'Roberts');RobD = edge(Depth,'Roberts');
Log = edge(RGB,'log');LogD = edge(Depth,'log');
Zer = edge(RGB,'zerocross');ZerD = edge(Depth,'zerocross');
Can = edge(RGB,'Canny');CanD = edge(Depth,'Canny');
App = edge(RGB,'approxcanny');AppD = edge(Depth,'approxcanny');
subplot(2,4,1);subimage(RGB);title('Original Color');
subplot(2,4,2);subimage(Sob);title('Sobel');
subplot(2,4,3);subimage(Pre);title('Prewitt');
subplot(2,4,4);subimage(Rob);title('Roberts');
subplot(2,4,5);subimage(Log);title('log');
subplot(2,4,6);subimage(Zer);title('zerocross');
subplot(2,4,7);subimage(Can);title('Canny');
subplot(2,4,8);subimage(App);title('approxcanny');
figure;
subplot(2,4,1);subimage(Depth);title('Original Depth');
subplot(2,4,2);subimage(SobD);title('Sobel');
subplot(2,4,3);subimage(PreD);title('Prewitt');
subplot(2,4,4);subimage(RobD);title('Roberts');
subplot(2,4,5);subimage(LogD);title('log');
subplot(2,4,6);subimage(ZerD);title('zerocross');
subplot(2,4,7);subimage(CanD);title('Canny');
subplot(2,4,8);subimage(AppD);title('approxcanny');
```

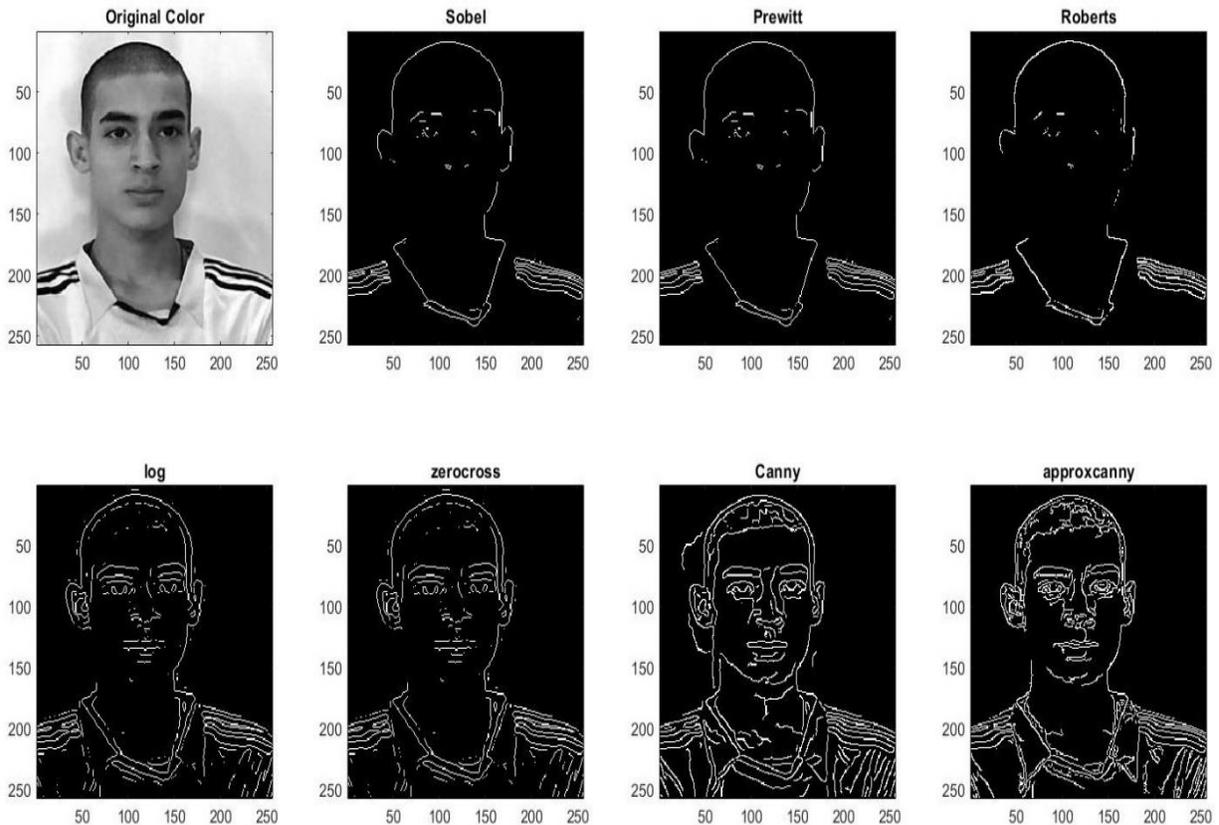

Figure 3.7 Different edge detection methods (color)

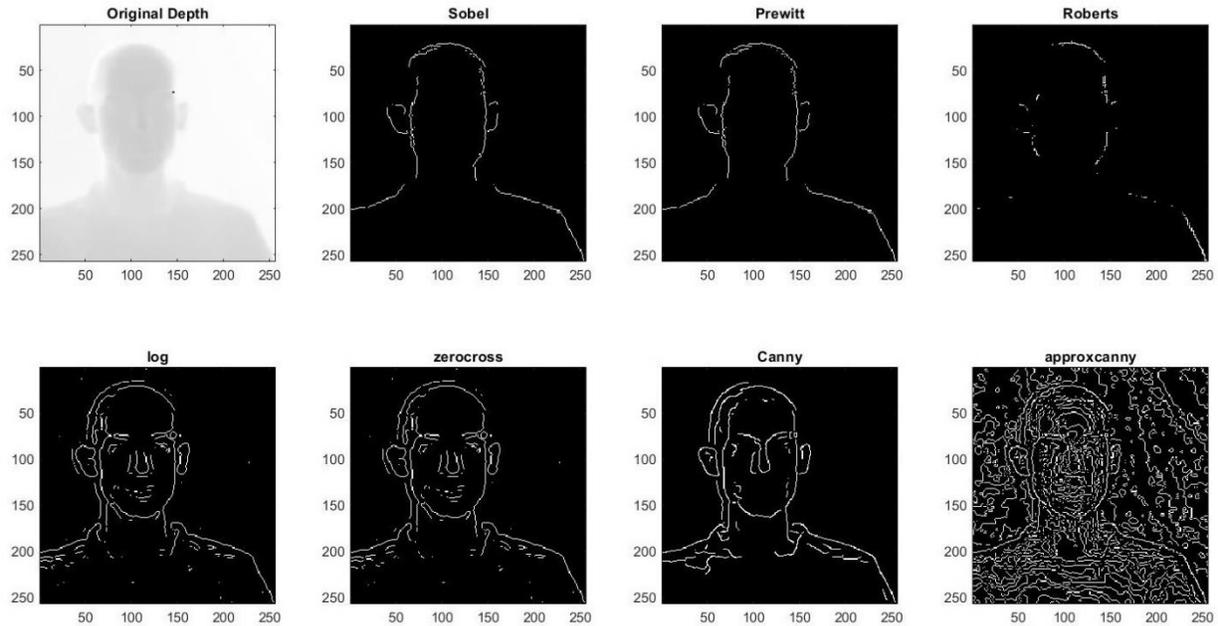

Figure 3.8 Different edge detection methods (depth)

## 3.6    Morphological Operations

Morphology [39] is a broad set of image processing operations that process images based on shapes. In a morphological operation, each pixel in the image is adjusted based on the value of other pixels in its neighborhood. By choosing the size and shape of the neighborhood, you can construct a morphological operation that is sensitive to specific shapes in the input image. Main seven of them are introduces here which are image dilation, erosion, filling holes, closing, opening, top hat filtering and bottom hat filtering [39]. They are kind of post-processing tasks and work better on binary or black and white images, but good enough in gray images too. In essence, dilation expands an image and erosion shrinks it. Opening, smooths the contour of an image, breaks isthmuses, eliminates protrusions. Closing, smooths sections of contours, but it generally fuses breaks, holes, gaps, etc. Following code, performs these operations on two color and depth images. Figures 3.9 and 3.10 show the effect of the code on color and depth images from IKFDB database respectively. Also, there are other morphological operations like hit or miss [39] and more which are not coded here.

```
% Morphological Operations
% Code name : "c.3.6.m"
clc;      % Clearing the Command Window
clear;    % Clearing the Workspace
RGB=imread('Exp_Nut_Color.jpg'); % Reading color image
RGB=imadjust(rgb2gray(RGB)); % Converting color image to gray
Depth= imread('Exp_Nut_Depth.png'); % Reading depth image
RGB = imresize(RGB, [256 256]); % Resizing color image to 256*256 dimensions
Depth = imresize(Depth, [256 256]); % Resizing depth image to 256*256 dimensions
se1 = strel('disk',10);% Morphological structuring element
se2 = strel('disk',100);% Morphological structuring element
se = strel('disk',3);% Morphological structuring element
dilated = imdilate(RGB,se);ddilated = imdilate(Depth,se); % Dilate operator
```

```matlab
eroded = imerode(RGB,se);deroded = imerode(Depth,se); % Erode operator
fill = imfill(RGB);dfill = imfill(Depth); % Fill holes
closing = imclose(RGB,se);dclosing = imclose(Depth,se); % Closing operator
opening = imopen(RGB,se);dopening = imopen(Depth,se); % Opening operator
tophat = imtophat(RGB,se1);dtophat = imtophat(Depth,se2); % Top-hat filter
bothat = imbothat(RGB,se1);dbothat = imbothat(Depth,se2); % Bottom-hat filter
subplot(2,4,1);subimage(RGB);title('Original Color');
subplot(2,4,2);subimage(dilated);title('Dilation');
subplot(2,4,3);subimage(eroded);title('Erosion');
subplot(2,4,4);subimage(fill);title('Filling');
subplot(2,4,5);subimage(closing);title('Closing');
subplot(2,4,6);subimage(opening);title('Opening');
subplot(2,4,7);subimage(tophat);title('Top hat');
subplot(2,4,8);subimage(bothat);title('Bottom hat');
figure; % Depth image plots
subplot(2,4,1);subimage(Depth);title('Original Depth');
subplot(2,4,2);subimage(ddilated);title('Dilation');
subplot(2,4,3);subimage(deroded);title('Erosion');
subplot(2,4,4);subimage(dfill);title('Filling');
subplot(2,4,5);subimage(dclosing);title('Closing');
subplot(2,4,6);subimage(dopening);title('Opening');
subplot(2,4,7);subimage(dtophat);title('Top hat');
subplot(2,4,8);subimage(dbothat);title('Bottom hat');
```

> *Note:*
> - *It is possible to call functions inside of each other like converting color to gray and adjusting the contrast of it in one line of command and setting running priority by parenthesis.*
>   *RGB=imadjust(rgb2gray(RGB));*

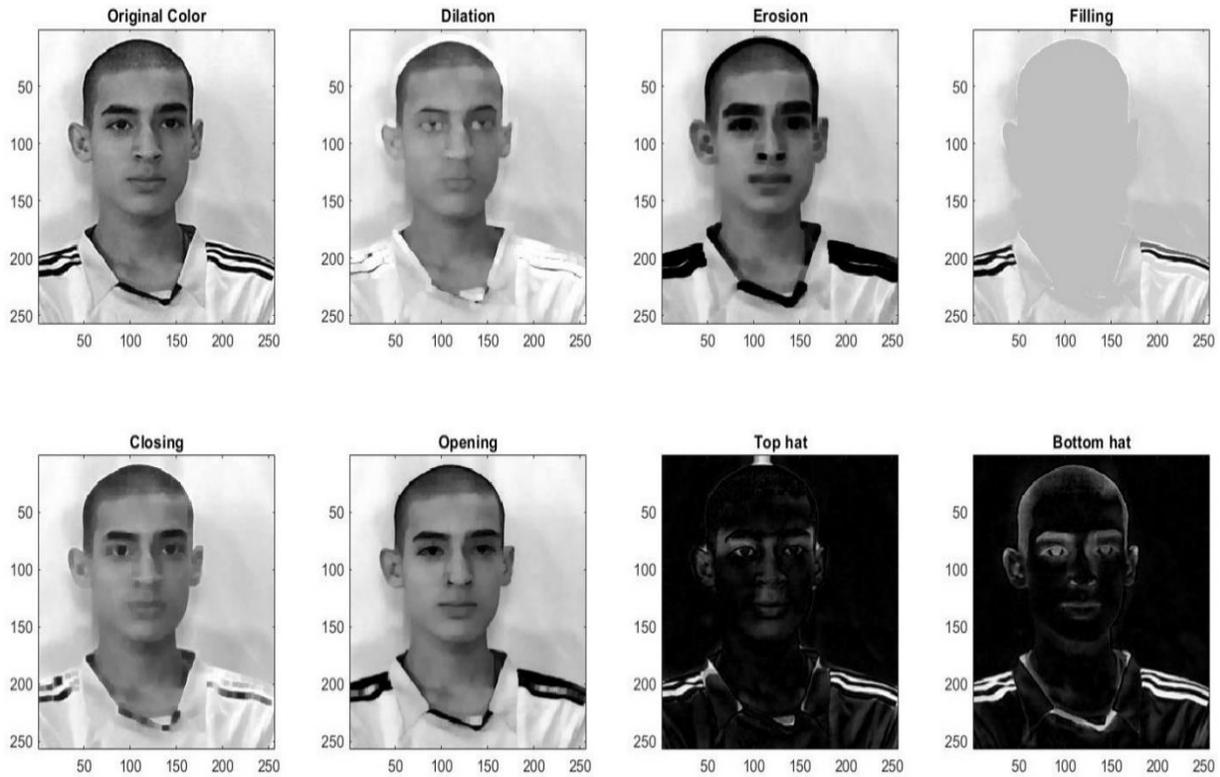

Figure 3.9 Morphological operation effect (color)

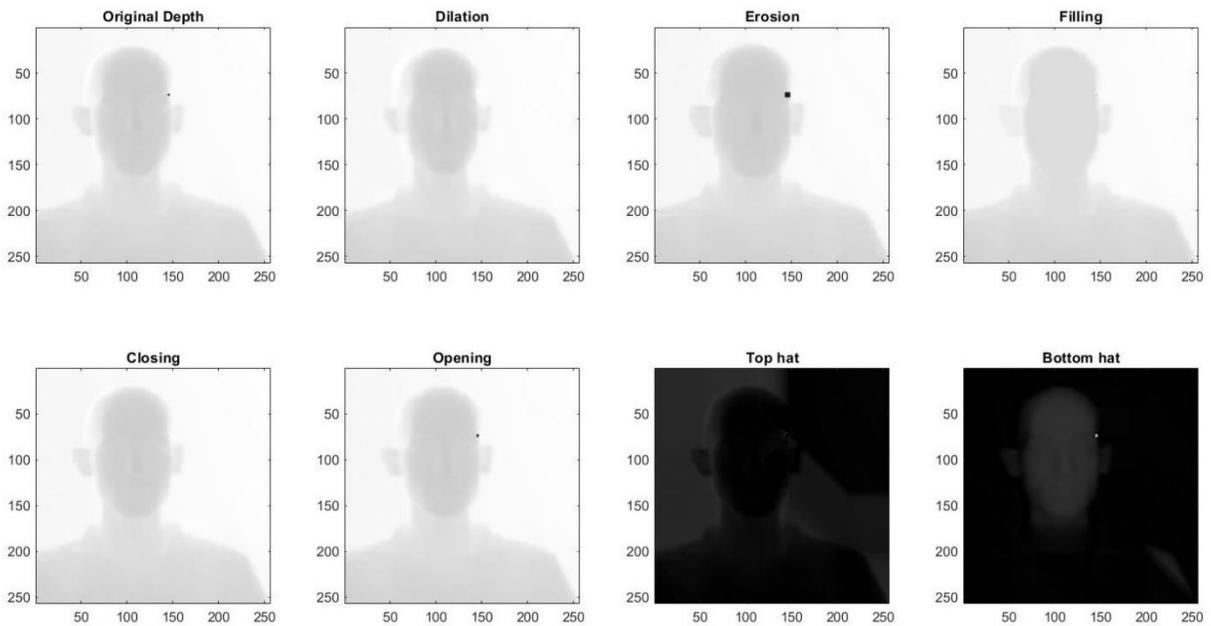

Figure 3.10 Morphological operation effect (depth)

## 3.7    Image Quality Assessment (IAQ) Metrics

With increasing technological power of software and hardware in recent devices, it is needed to make more precise algorithms to deal with them. Digital image is not an exception in this area. Assessing the quality of digital image is important in three main fields. First, it can be used to monitor image quality for quality control systems. For instance, a video acquisition system could get help from Image Quality Assessments (IQA) metrics [23, 16, 29, 22] and adjust itself with the best receiving image possible. Second, IQA metric could be used in benchmark systems, which are used in image processing applications. Third, it can be combined into an image processing system to optimize the algorithms and the parameter settings.

A lot of traditional IQA metrics such as Mean squared error (MSE) [23, 16, 29, 22], Peak signal-to-noise ratio (PSNR) [23, 16, 29, 22] and Structural similarity (SSIM) [23, 16, 29, 22] are completely use visual information inside an image to measure its quality. This information is too much while the Human Visual System (HVS) [6, 7] gets an image mainly based on its low level such as edges, zero-crossings, corners and lines [40, 41]. So, realizable image degradations induce corresponding changes in low-level image features. Digital images are subject to a wide variety of distortions during transmission, acquisition, processing, compression, storage and reproduction and this may result in degradation of visual quality. Image quality can refer to the level of accuracy that all imaging systems and sensors capture, process, store, compress, transmit and display [42]. As it mentioned, HVS is a fine tool for IQA, but it is not still a device to use. Automatic IQA metrics are made for these purposes. IQA metrics fall into two main categories. Subjective and objective IQA metrics. Assessing the quality of an image based on HVS is called subjective IQA. Makin an automatic IQA system which works as HVS is called objective IQA system. Many researches have been done during years to make a universal IQA system, but it was not so successful. These types of systems fall into three main categories. Full-Reference (FR) [43], Non- Reference (NR) [44] and Reduced-Reference (RR) [45]. Here just full and no reference objective metrics are mentioned as they are the most applicable. This method is the most, accurate and famous one, which needs both reference and distorted images. It is called blind IQA. In this approach reference image is not exist. For example, in digital photography, a non-reference assessment algorithm is used in order to inform the user which a low-quality or high-quality image has been received. HVS could determine the quality in this condition, but it is so difficult to make an automatic system with same characteristics as human eyes. PSNR shows the level of losses or signals unity. The Peak Signal to Noise Ratio (PSNR) block computes the peak signal-to-noise ratio, in decibels, between two images. This ratio is often used as a quality measurement between the original and a compressed image. The Mean Square Error (MSE) and the Peak Signal to Noise Ratio (PSNR) are the two-error metrics used to compare image compression quality. The MSE represents the cumulative squared error between the compressed and the original image, whereas PSNR represents a measure of the peak error. Structural similarity can be obtained by comparing pixel intensities' local patterns that have been normalized for luminance and contrast. Their calculation depends on the separate calculation of luminance, contrast, and structure. No referenced IQA here are Blind/Reference less Image Spatial Quality Evaluator (BRISQUE) [46], Naturalness Image Quality Evaluator (NIQE) [47] and Perception based Image Quality Evaluator (PIQE) [48]. Following code add gaussian noise to original image and calculates PSNR, MSE and SSIM from full reference objective IQA category and BRISQUE, NIQE and PIQE from no referenced objective IQA category and returns the result.

```
% IQA metrics
% Code name : "c.3.7.m"
clc;      % Clearing the Command Window
clear;  % Clearing the Workspace
Depth= imread('Exp_Nut_Depth.png'); % Reading depth image
Depth = imresize(Depth, [256 256]); % Resizing depth image to 256*256
dimensions
noisy = imnoise(Depth,'gaussian', 0.02); % Adding gaussian noise
%%Full refrence
% Peak Signal to Noise Ratio (PSNR)
[peaksnr, snr] = psnr(noisy, Depth);
fprintf('\n The Peak-SNR value is %0.4f', peaksnr);
% Mean Square Error (MSE)
```

```matlab
mse = immse(noisy, Depth);
fprintf('\n The mean-squared error is %0.4f\n', mse);
% Structural similarity (SSIM) index
[ssimval,ssimmap] = ssim(noisy, Depth);
fprintf('\n The SSIM error is %0.4f\n', ssimval);
%% No refrence
% Blind/Referenceless Image Spatial Quality Evaluator (BRISQUE)
brisqueI = brisque(Depth);
fprintf('BRISQUE score for original image is %0.4f.\n',brisqueI)
brisqueInoise = brisque(noisy);
fprintf('BRISQUE score for noisy image is %0.4f.\n',brisqueInoise)
% Naturalness Image Quality Evaluator (NIQE)
niqeI = niqe(Depth);
fprintf('NIQE score for original image is %0.4f.\n',niqeI)
niqeInoise = niqe(noisy);
fprintf('NIQE score for noisy image is %0.4f.\n',niqeInoise)
% Perception based Image Quality Evaluator (PIQE)
piqeI = piqe(Depth);
fprintf('PIQE score for original image is %0.4f.\n',piqeI)
piqeInoise = piqe(noisy);
fprintf(' PIQE score for noisy image is %0.4f.\n',piqeInoise)
```

The result is:

**The PSNR value is 22.3076**

**The MSE is 382.2285**

**The SSIM is 0.1857**

**BRISQUE score for original image is 47.8808.**

**BRISQUE score for noisy image is 44.7610.**

**NIQE score for original image is 9.1109.**

**NIQE score for noisy image is 40.9300.**

**PIQE score for original image is 100.0000.**

**PIQE score for noisy image is 75.0568.**

### 3.7    Group or Batch Processing

As it mentioned earlier for functions, when working on a real project, functions play important roles just like batch processing. In dealing with real world data, it is necessary to use batch or group processing. That is due to dealing with high number of data which here is image type. For example, and in a project for face FER task, you might work with a dataset with 1000 and more images, and it is not rational to make variable for each one of them and code them with 1000 lines of codes. In this situation, a loop with 3 or 4 lines of code makes the task as easy as possible. Fortunately, Matlab software makes batch processing possible with few lines of codes. A cell array is a data type with indexed data containers called cells, where each cell can contain any type of data. Cell arrays commonly contain either lists of text, combinations of text and numbers, or numeric arrays of different sizes. Following lines of codes takes JAFFE facial expression dataset's [49] images as input from a specific path of hard drive and save it into "images" cell array. Each cell contains an image which makes it easier to achieve any image in any time just by calling its cells. So, it is possible to make any change just by a simple loop over the "images" cell. For example, converting color images into gray is possible to be done just by writing 4 lines of codes. After converting it to gray, histogram equalization, contrast adjustment and resizing, it is time to show them all together with "montage" command or function. Figure 3.11 shows the dataset after batch processing. Also, Figure 3.12 shows some of the variable after processing of the code.

```matlab
% Batch processing
% Code name : "c.3.8.m"
% Read the input images
path='c:\dataset';% Data Directory Path of Hard Drive
fileinfo = dir(fullfile(path,'*.jpg')); % Files properties
filesnumber=size(fileinfo); % Number of files
for i = 1 : filesnumber(1,1) % Reading loop as number of files
images{i} = imread(fullfile(path,fileinfo(i).name)); %Creating cell array of
"images"
    disp(['Loading image No :   ' num2str(i) ]); % Reading iterations in the
loop as message
end;
%RGB to gray convertion
for i = 1 : filesnumber(1,1)
gray{i}=rgb2gray(images{i});
    disp(['To Gray :   ' num2str(i) ]);
end;
%Histogram equalization
for i = 1 : filesnumber(1,1)
histequation{i}=histeq(gray{i});
    disp(['Hist EQ :   ' num2str(i) ]);
end;
%Contrast adjustment
for i = 1 : filesnumber(1,1)
adjusted{i}=imadjust(histequation{i});
    disp(['Image Adjust :   ' num2str(i) ]);
end;
% Resize the final image size
for i = 1 : filesnumber(1,1)
resized{i}=imresize(adjusted{i}, [128 128]);
    disp(['Image Resized :   ' num2str(i) ]);
end;
%Display multiple image frames as rectangular montage
montage(resized, 'Size', [10 20]);
```

---

*Note:*
> *"Montage" function or command, could present multiple images at the same time in one figure without any empty spaces between them. It is helpful when we have series of related images.*

---

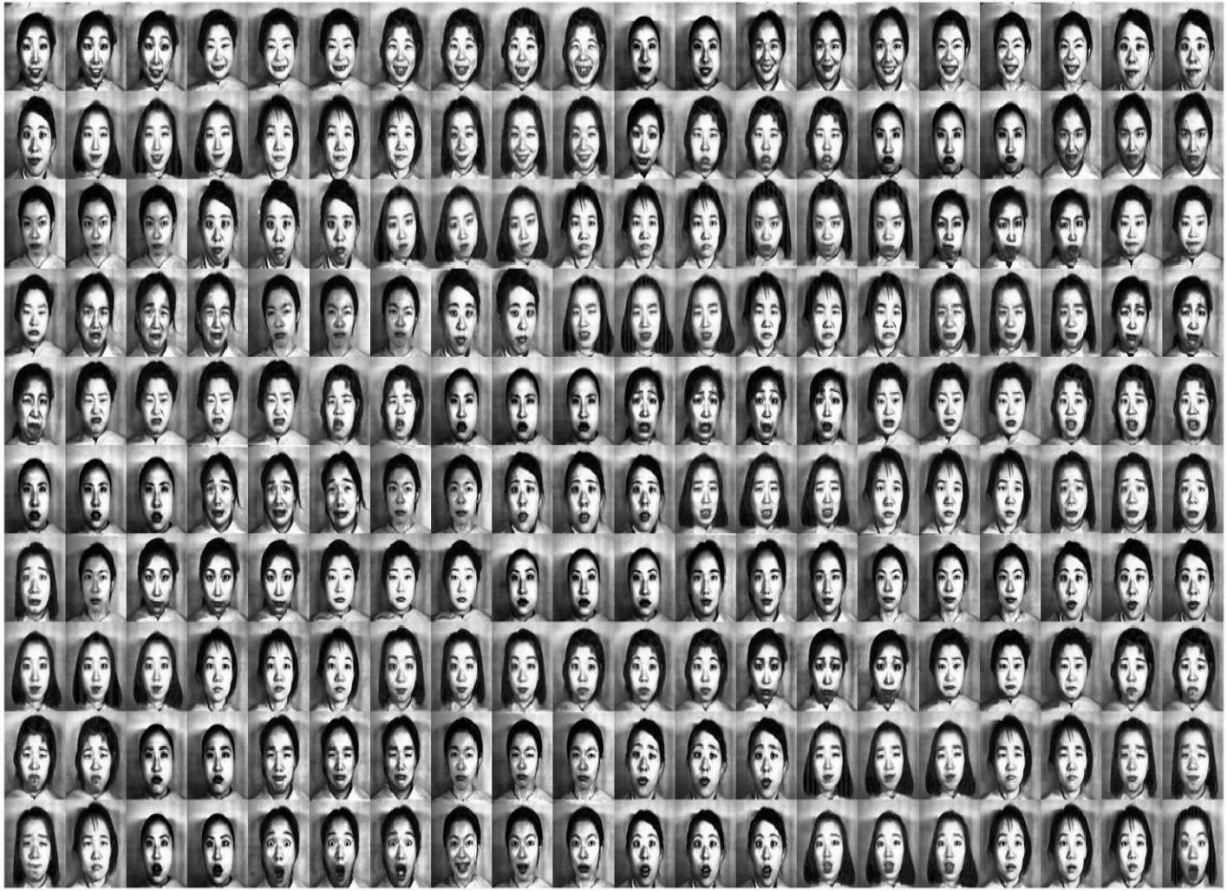

Figure 3.11 Jaffe dataset [49] after batch processing (plotted with "montage" command or function)

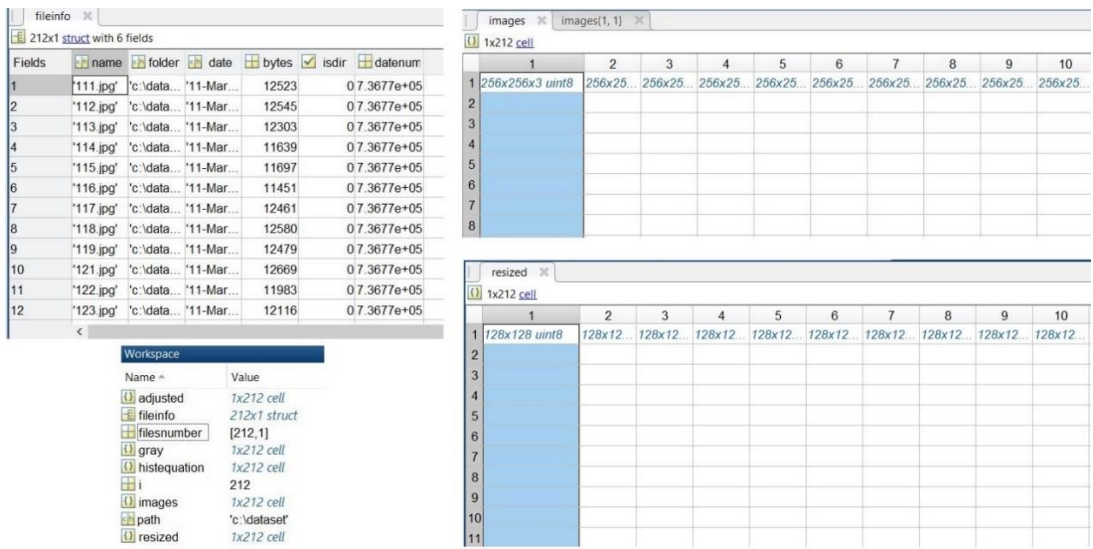

Figure 3.12 "images" and "resized" cells, "fileinfo" struct and the variables in workspace after running code "c.3.8.m"

**3.8 Exercises**

1. (P1): What is the different between "for" and "while" loops (with an example)?

2. (P2): Convert the following code from "if" to "switch" and from for to "while"?

```
nrows = 4;
ncols = 6;
A = ones(nrows,ncols);
for c = 1:ncols
    for r = 1:nrows
        if r == c
            A(r,c) = 2;
        elseif abs(r-c) == 1
            A(r,c) = -1;
        else
            A(r,c) = 0;
        end
    end
end
```

2. (P2): Write a function called "preproc" that has one input of "image" and two outputs of "histo" and "noisy". First read the image, converted to gray and then send it to function for process. Function should add Gaussian noise to the image and calculate its histogram. Place their values into "noisy" and "histo" and return them into the workspace. Final images should be plotted with their dimensions.

3. (P3): Read "happydepth.jpg" image from test folder of the book and transfer it to the frequency domain and smoothed with Gaussian Low Pass Filter (GLPF) having cut of frequency of 15. Use inverse Fourier transform and after having it in spatial domain, sharpened the result image from frequency image using unsharp masking filter in spatial domain and montage both images for final representation. (needs external reference and research by the reader).

4. (P4): Read 10 first images from "JAFFE" dataset from book test folder and detect the edges using "Fuzzy edge detection" and use "hit or miss" morphological operation to make the edges narrower. Add speckle noises to all images and calculate "SSIM" IQA metric on both original and noisy images and print them out in the command windows. (needs external reference and research by the reader.

# *Chapter    4*

## *Face Analysis,  FER and FMER*

# *Chapter  4*

## *Face Analysis, FER and FMER*

## Contents

This chapter is slightly smaller chapter with less code. This chapter pays to face analysis totally. Application of face analysis, explaining facial parts and muscles, Facial Action Coding System (FACS) [50], Facial Expressions Recognition (FER) [51] and Facial Micro Expressions Recognition (FMER) [51], weighting facial parts for having better recognition accuracy, recording with Kinect V.2 sensor for color and depth data and finally famous face detection and extraction techniques and methods for color and depth images.

### 4.1   Face Analysis

Any experiment on human face to achieve data and understanding from that person is called face or facial analysis. Tasks such as, age estimation, face recognition, face detection, FER, FMER and more are part of facial analysis definition. Facial expressions are the facial changes in response to a person's internal emotional states, intentions, or social communications. Facial expression analysis has been an active research topic for behavioral scientists since the work of Darwin in 1872 [52]. After that, much progress has been made to build computer systems to help us understand and use this natural form of human communication. Facial expression analysis refers to computer systems that attempt to automatically analyze and recognize facial motions and facial feature changes from visual information. Sometimes the facial expression analysis has been confused with emotion analysis in the computer vision domain. For emotion analysis, higher level knowledge is required. For example, although facial expressions can convey emotion, they can also express intention, cognitive processes, physical effort, or other intra- or interpersonal meanings. Computer facial expression analysis systems need to context, body gesture, voice, individual differences, and cultural factors as well as by facial analyze the facial actions regardless of context, culture, gender, and so on.

**4.2   Facial Parts and Muscles**

In FER and FMER, parts such as, eyes, noise, mouth, forehead, eyebrows and cheeks have the highest impact in the recognition accuracy. For example, just by having eyes, mouth and forehead muscles, it is possible to have good FER and FMER recognition accuracy. Figure 4.1 shows human body anatomy along side with each muscle names. Also, figure 4.2 presents head muscles names as they are making expressions on the face. Facial features are using to determine race, gender, mood, age and etc. Some of these features are permanent like bone structure, skin texture color and some of them are temporary like cosmetics, glass, bear and facial muscle exercises. In total facial muscles are the main factor for facial expressions appearance [53].

- Occipitofrontalis

The large muscle of the forehead. Some experts omit this muscle when reconstructing the face as it is thin and they feel that it does not contribute significantly to the overall contours of the face.

- Temporalis

A thick-fan shaped muscle that closes the mouth and assists the jaw to move side-to-side to grind up food.

- Buccinator

Sometimes known as the 'trumpeter muscle', the Buccinator's role is to puff out the cheeks and prevent food from passing to the outer surface of the teeth during chewing.

- Masseter

This runs from the cheekbone to the lower jaw and brings the teeth back together to grind up food. The Masseter is the strongest muscle in the human body.

- Mentalis

Sometimes called the 'pouting muscle', contraction of the Mentalis raises and thrusts out the lower lip to make us pout.

- Depressor Labii Inferioris

This muscle pulls down the bottom lip allowing us to sulk.

- Orbicularis Oris

The circular muscle around our mouth is Orbicularis Oris and this muscle brings our lips together.

- Levator Anguli Oris

The happy muscle, making the corners of our mouth turn upwards into a smile.

- Levator Labii Superioris

The muscle that deepens the furrows either side of our nose and top lip when we feel sad.

- Depressor Anguli Oris

This muscle lowers the corners of our mouth into a frown.

- Levator Labii Superioris Alaeque Nasi

This muscle dilates the nostrils and raises the upper lip. It's often referred to as the 'Elvis muscle' in homage to Elvis Presley.

- Zygomatic Major and Minor

These are strap muscles that help to form the shape of the cheeks. Both muscles are involved in elevating the upper lip to generate a smile, with the minor muscle allowing us to curl our top lip which usually demonstrates smugness.

- The Orbicularis Oculi

A distinctive muscle that is constructed of two parts, the palpebral and orbital. The palpebral area lies at the center of this sphincter muscle and forms the eyelids with the orbital region encasing it concentrically. The Orbicularis Oculi muscle opens and closes the eyelids thus allowing us to blink, wink or squint in bright sunlight.

- Risorius

A very thin and delicate muscle that pulls the lips horizontally creating a large, albeit insincere smile.

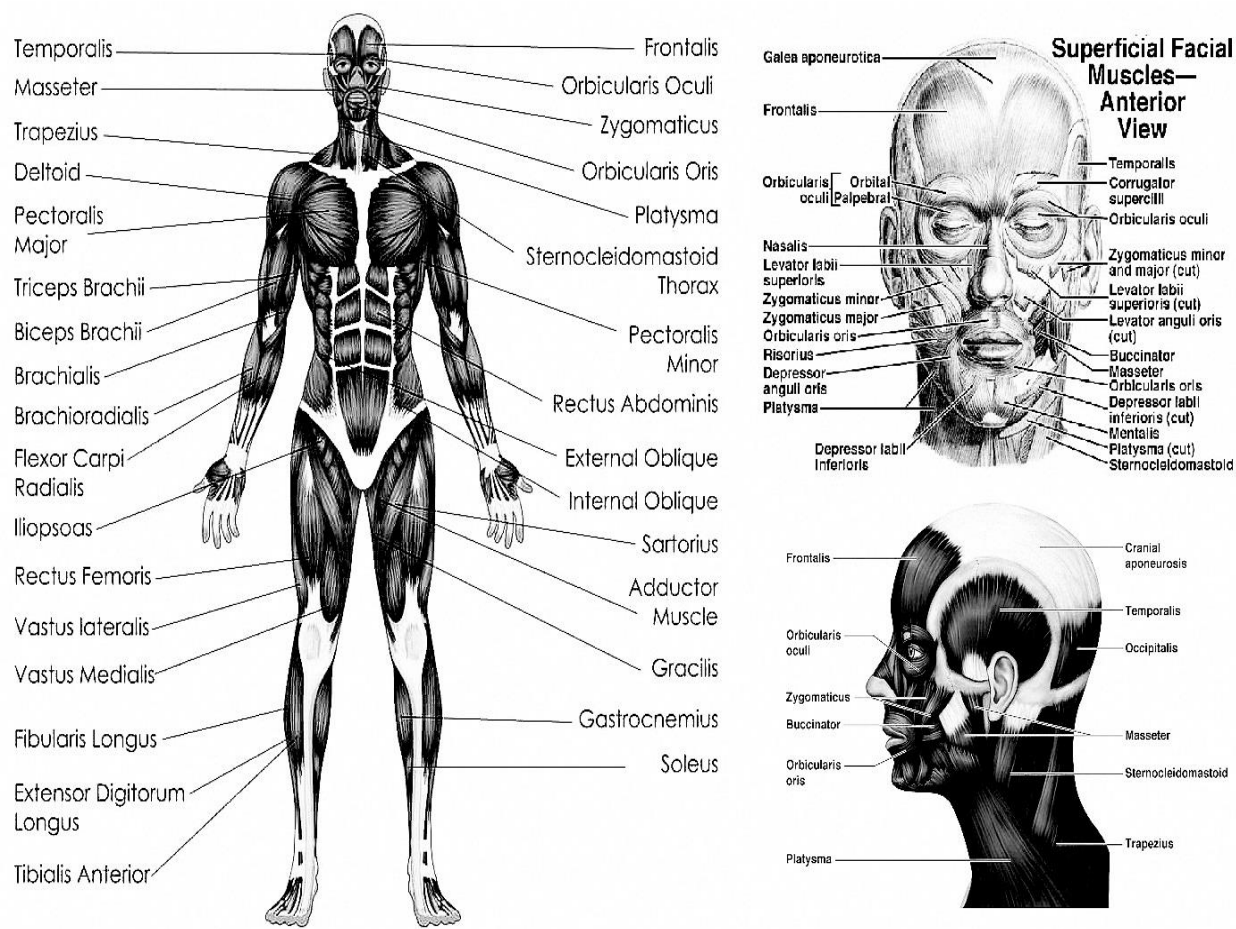

Figure 4.1 Human body anatomy (body and head muscles names) [53]

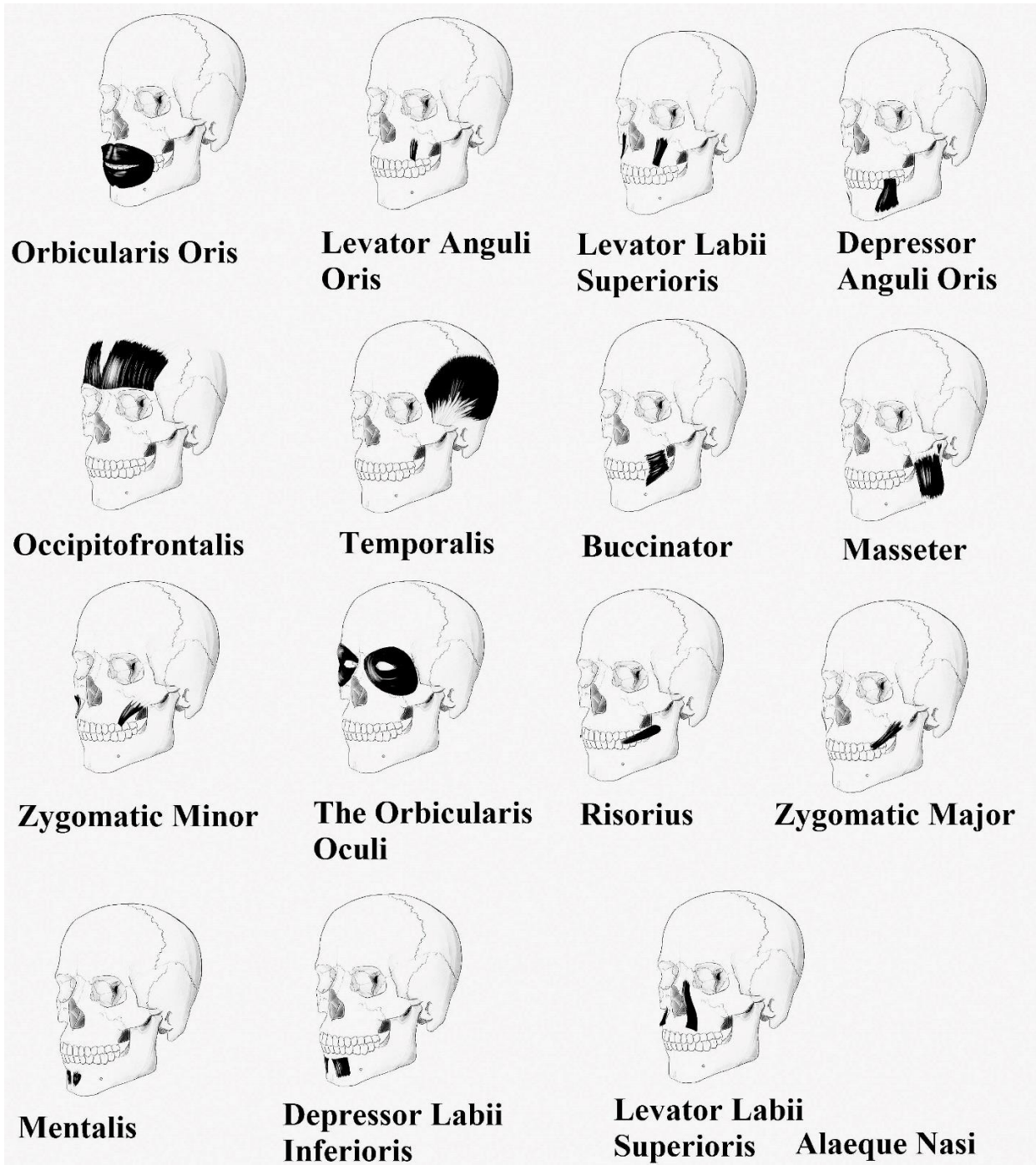

Figure 4.2 human head muscles names [53]

**4.3 Facial Action Coding System (FACS)**

Facial Action Coding System (FACS) [50, 21] is the best way for coding human facial muscles movement known. Each action consisted of a muscle movement which with combining them, making every expression is possible. FACS is made in 1978 and included 44 action units and during time increased till 51 action units in 2002 by scientists. Action units 1 to 7 are related to upper parts of the face and others to lower parts. First 20 action units along with their specifications or muscle descriptions are described in the Table 4.1.

By considering action units and combining them, it is possible to get all expressions possible on human face. Below just some action unit's combination and their related expressions are mentioned. Figure 4.3 presents some expressions based on FACS.

- 26+25+5+2+1= Surprise
- 6+12 = Joy
- 17+15+1= Sadness
- 25+10+9+7+4 = Anger

Table 4.1 First 20 actions unit in FACS

| AU number | FACS name | Muscular basis |
|---|---|---|
| **0** | Neutral face | - |
| **1** | Inner brow raiser | frontalis (pars medialis) |
| **2** | Outer brow raiser | frontalis (pars lateralis) |
| **4** | Brow lowered | depressor glabellae, depressor supercilii, corrugator supercilii |
| **5** | Upper lid raiser | levator palpebrae superioris, superior tarsal muscle |
| **6** | Cheek raiser | orbicularis oculi (pars orbitalis) |
| **7** | Lid tightener | orbicularis oculi (pars palpebralis) |
| **8** | Lips toward each other | orbicularis oris |
| **9** | Nose wrinkle | levator labii superioris alaeque nasi |
| **10** | Upper lip raiser | levator labii superioris, caput infraorbitalis |
| **11** | Nasolabial deepener | zygomaticus minor |
| **12** | Lip corner puller | zygomaticus major |
| **13** | Sharp lip puller | levator anguli oris (also known as caninus) |
| **14** | Dimple | buccinator |
| **15** | Lip corner depressor | depressor anguli oris (also known as triangularis) |
| **16** | Lower lip depressor | depressor labii inferioris |
| **17** | Chin raiser | mentalis |
| **18** | Lip pucker | incisivii labii superioris and incisivii labii inferioris |
| **19** | Tongue show | - |
| **20** | Lip stretcher | risorius w/ platysma |

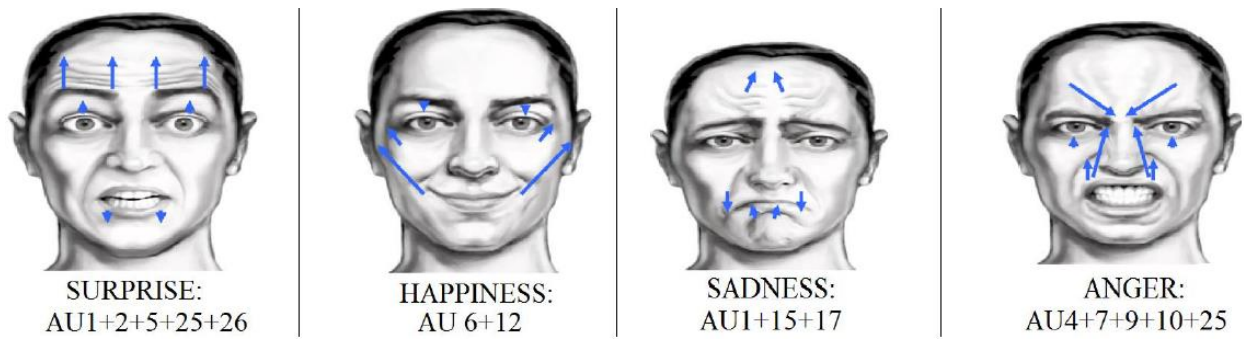

SURPRISE:  HAPPINESS:  SADNESS:  ANGER:
AU1+2+5+25+26  AU 6+12  AU1+15+17  AU4+7+9+10+25

Figure 4.3 Four famous expressions based FACS

**4.4   Facial Expressions and Micro Expressions Recognition (FER) and (FMER)**

In order to explain the Facial Expressions Recognition (FER) [51, 12] and Facial Micro Expressions Recognition (FMER) [51, 54], it is needed to explain face detection and recognition first. If a system could distinguish the face objects out any other objects in a digital image, then this system is called face detection system. Now if a system could distinguish a specific identity by face in a bunch other face images, then the system is called face recognition system. Each human face, despite of gender, age and race could express seven main expressions in general. Expression recognition states which type of emotion subject is in, out of seven main emotions. It is mentionable that other emotions or expressions are combinations of these seven emotions, or it can be said combination of facial muscle which are employed to express seven main emotions. These seven main emotions or face expressions are joy or happiness, sadness, anger, surprise, disgust, fear and neutral. Facial micro expressions are completely similar to the facial expressions, except the time of accruing. Facial micro expression happens in 1/3, 1/5 or 1/25 second [51, 54]. So, for recording these very fast actions, employment of sensors with 30-90 frame per second is necessary. Figure 4.4 shows all seven main facial expressions along with their 3-D model. Figure 4.5 shows some of the micro expressions' samples from CASME dataset [51] with its relative action units.

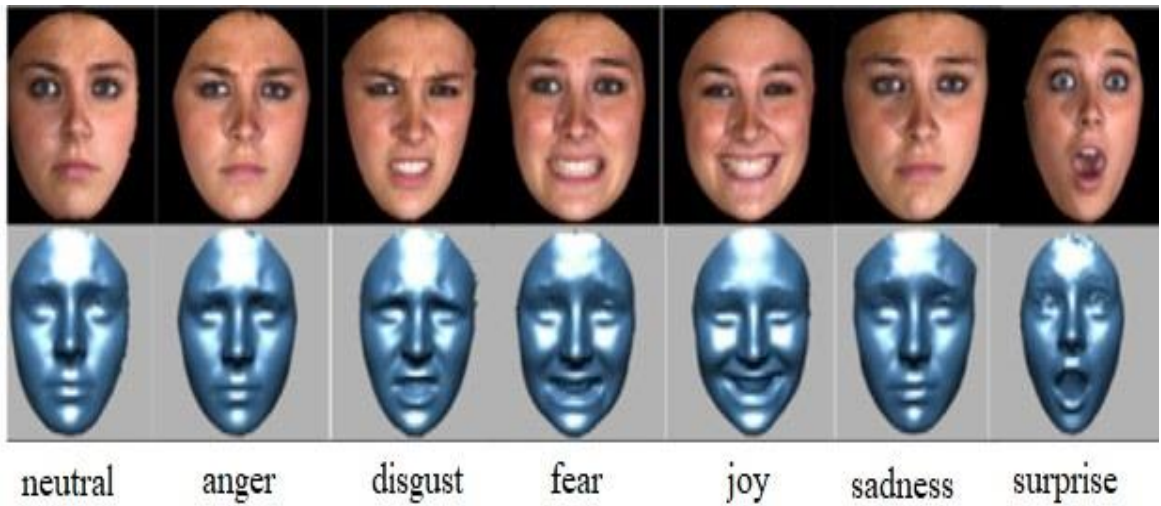

Figure 4.4 Seven main facial expressions

| Micro Expression Happy | Micro Expression Disgust | Micro Expression Anger | Micro Expression Fear | Micro Expression Neutral | Micro Expression Sadness | Micro Expression Surprise |
|---|---|---|---|---|---|---|
| Action Unit(s): | Action Unit(s): | Action Unit(s): | Action Unit(s): | Action Unit(s): | Action Unit(s): | Action Unit(s): |
| 12 | 15+9+4 | 24+4 | 7+20 | 0 | 15+4+1 | 26+2+1 |

Figure 4.5 Micro expression samples from CASME dataset [51]

> *Note:*
> *This book just mentioned most important FACS and action units. There are more FACS and action units which makes possible to make more than seven facial expressions.*

## 4.5 Weighting Facial Parts

It is better to weight each face element to have better result in final recognition accuracy. Eyes and mouth due to have higher effect, should have more weight than other parts. Facial parts weighting means some facial elements should have more impact in feature extraction and classification tasks. Figure 4.6 shows some facial elements or parts and facial elements weighting. Also, Figure 4.7 shows mouth, eye and nose facial parts from different datasets.

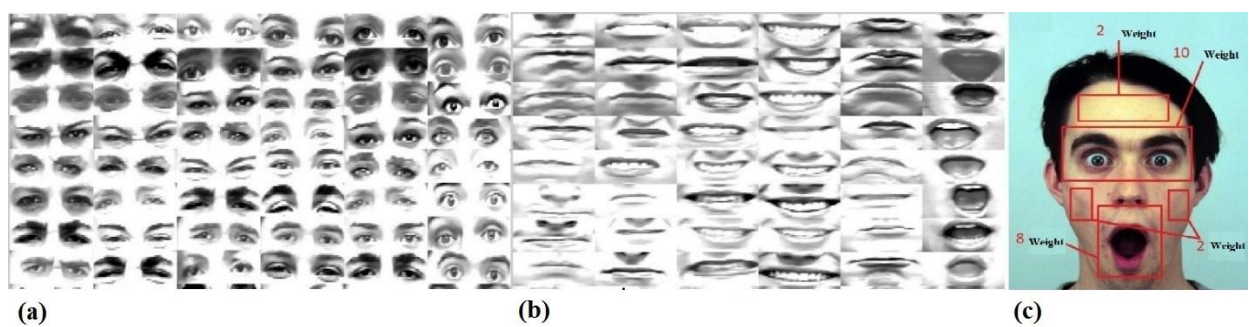

(a)                                    (b)                                    (c)

Figure 4.6 Face part eye (a), face part mouth (b) and facial parts weighting (c)

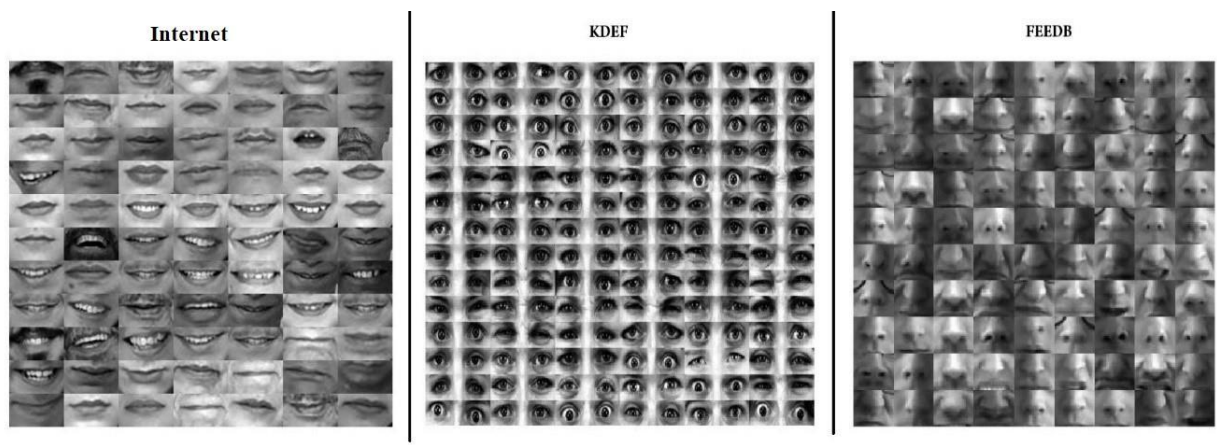

Figure 4.7 Some samples of Internet based, KDEF [35] and FEEDB [55] databases

## 4.6 Recording Color and Depth Data with Kinect

In order to record with Kinect, it is possible to record both color and depth images frame by frame, simultaneously. To run Kinect sensor into Matlab, it is needed to install some toolboxes and software's on Matlab software. First you have to install KinectSDK, then Image Acquisition Toolbox Support Package for Kinect for Windows Sensor. Also, installing Kinect developer toolkit and MicrosoftSpeechPlatformSDK are recommended. Totally you should provide following things:

- Requirements for the Kinect V2 Support

The following requirements apply to the Kinect V2 support in Image Acquisition Toolbox.

The Image Acquisition Toolbox Support Package for Kinect for Windows Sensor must be installed.
The Kinect adaptor is supported on 64-bit Windows.
The Kinect V2 support requires Windows 8.0.
You must use a USB 3.0 host controller to connect your Kinect V2 device to your computer.
To use Kinect for Windows V2 support, you must have version 2.0 of the Kinect for Windows Runtime installed on your system. If you do not already have it installed, it installs with the Kinect support package.
You can only use one Kinect V2 device at a time with the Image Acquisition Toolbox. You can use a Kinect V2 and a Kinect V1 at the same time, but only one Kinect V2 device at a time. This is a hardware limitation.

- Features of the Kinect V2 Support

Kinect for Windows V2 offers these new features:
Higher resolution capabilities, including three times more depth fidelity and a cleaner depth map
Wider depth and color field of view
A 25-point skeleton for each of up to six people (Kinect V1 has 20 joints)
Tracks up to six people simultaneously (Kinect V1 tracks only 2)
Open-hand and closed-hand gesture recognition
Biocorrect body joints (particularly hip, shoulder, and spine)
Higher confidence for joints and more points for hands

Following lines of code, runs both color and depth sensors for 100 frames or 3 seconds and save them into "kinect rgb and depth" folder. Obviously, you can increase number of frames and change the saving destination folder.

```matlab
% Recording with Kinect
% Code name : "c.4.1.m"
clear;
% Add Utility Function to the MATLAB Path
utilpath = fullfile(matlabroot, 'toolbox', 'imaq', 'imaqdemos','html',
'KinectForWindows');
addpath(utilpath);
% separate VIDEOINPUT object needs to be created for each of the color and
depth(IR) devices
% The Kinect for Windows Sensor shows up as two separate devices in
IMAQHWINFO
hwInfo = imaqhwinfo('kinect')
hwInfo.DeviceInfo(1)
hwInfo.DeviceInfo(2)
% Create the VIDEOINPUT objects for the two streams
colorVid = videoinput('kinect',1)
depthVid = videoinput('kinect',2)
% Set the triggering mode to 'manual'
triggerconfig([colorVid depthVid],'manual');
%  In this example 100 frames are acquired to give the Kinect for Windows
sensor sufficient time to
%  start tracking a skeleton.
numberofframe=100;
colorVid.FramesPerTrigger = numberofframe;
depthVid.FramesPerTrigger = numberofframe;
% Start the color and depth device. This begins acquisition, but does not
% start logging of acquired data.
start([colorVid depthVid]);
% Trigger the devices to start logging of data.
```

```
trigger([colorVid depthVid]);
% Retrieve the acquired data
[colorFrameData,colorTimeData,colorMetaData] = getdata(colorVid);
[depthFrameData,depthTimeData,depthMetaData] = getdata(depthVid);
% Stop the devices
stop([colorVid depthVid]);

%% converting 4-d matrix to 3-d rgb images
rgb4=size(colorFrameData)
for i=1:rgb4(1,4)
    rgb{i}=colorFrameData(:,:,:,i)
end;
%%%%%%%%%%%%%%%%%%%%%%%%%%%%
% converting 4-d matrix to 3-d depth images
depth4=size(depthFrameData)
for i=1:depth4(1,4)
    depth{i}=depthFrameData(:,:,:,i)
end;
%% Saving image to drive c(RGB)
%first delete previews files from specific folder
delete('c:\kinect rgb and depth\rgb\*.jpg');
%then saving new files
for i = 1 : rgb4(1,4)
imwrite(rgb{i},['c:\kinect rgb and depth\rgb\rgb image' num2str(i) '.jpg']);
    disp(['No of saved RGB image :   ' num2str(i) ]);
end;
%save as mat file
save('c:\kinect rgb and depth\rgb\rgb.mat','rgb');
%%%%%%%%%%%%%%%%%%%%%%%%%%%%
% Saving image to drive c(DEPTH)
%first delete previews files from specific folder
delete('c:\kinect rgb and depth\depth\*.jpg');
%then saving new files
for i = 1 : depth4(1,4)
imwrite(depth{i},['c:\kinect rgb and depth\depth\depth image' num2str(i)
'.png']);
    disp(['No of saved DEPTH image :    ' num2str(i) ]);
end;
%save as mat file
save('c:\kinect rgb and depth\depth\depth.mat','depth');
```

Following lines of codes converts frames to a single video. These frames belong to a recording of IKFDB dataset in depth mode. Frames are into the folder "frames" the video will be saved into book main folder under the title of "Sample.avi".

```
% Converting image frames to video
% Code name : "c.4.2.m"
clear;
path='frames';
fileinfo = dir(fullfile(path,'*.png'));
filesnumber=size(fileinfo);
fileinformation = struct2cell(fileinfo);
fileinformation=fileinformation';
for i=1:filesnumber(1,1)
names{i,1}=fileinformation{i,1};
end;
```

```
outputVideo = VideoWriter(fullfile('Sample.avi'));
open(outputVideo)
for i = 1:filesnumber(1,1)
    img = imread(fullfile(path,names{i}));
    writeVideo(outputVideo,img)
        disp(['Joining IMG :   ' num2str(i) ]);
end
close(outputVideo)
```

### 4.7    Face Detection and Extraction from Color and Depth Images

In order to recognizing facial expressions, first it is needed to detect and extract the face out of the image or removing the outliers. Outliers are unwanted data in a matrix. As an image is a matrix, it is possible to remove outliers of the image by cropping process. Two main face detection and extraction methods from color and depth images. For color images, Viola $ Jones [56] face detection algorithm has higher efficiency and for depth images the [13] is the best method. This algorithm [56] is almost one of the most accurate and fastest face detection algorithms and has been around for a long period of time in facial image processing. This algorithm is so robust and could be employed for depth images with lower accuracy as well. This algorithm is an object detection algorithm and could be used for any learned object but that is mainly used for face. The algorithm has four stages of 1. Haar Feature Selection, 2. Creating an Integral Image, 3. Adaboost Training and finally Cascading Classifiers.

A new interesting method to extracting face out of depth images is as follow. Following shows the steps: 1. Finding closest pixel to the sensor with smallest value (nose tip). 2. Rectangular face cropping to a specific threshold and saving it into a temporary and second matrixes. 3. Applying standard deviation filter to second matrix and using ellipse fitting technique to fit ellipse on face. 4. Selecting pixels inside the temporary matrix based on second matrix calculations and removing other pixels. 5. Finally, some percentage of the image sides could be removed to increase accuracy [13]. Figure 4.8 shows the process of depth face detection and extraction algorithm process.

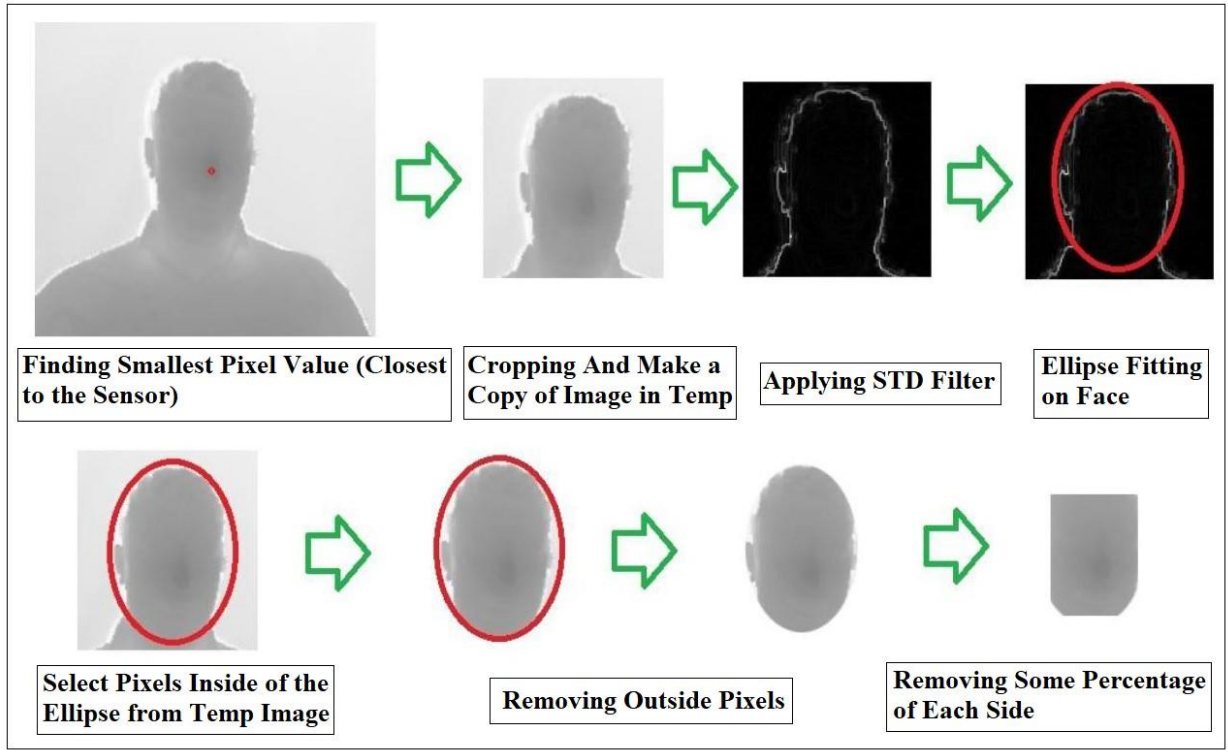

Figure 4.8 Depth face detection and extraction [13]

Following lines of codes, could detect any facial parts from image frames, but it has to be determined which part. For example, here, Viola & Jones detects and extracts face out of image as 'FrontalFaceCART' is used for vision.CascadeObjectDetector. Obviously, it is possible to change to nose or mouth to extract and show them. It will extract face from "viola" folder in the book main folder. Frames belongs to IKFDB database. Also, the algorithm shows good performance on depth images too. Figure 4.9 shows the performance of the code 'c.4.3.m".

```matlab
% Viola and Jones face detection
% Code name : "c.4.3.m"
clear;
%Detect objects using Viola-Jones Algorithm
%To detect Face use 'FrontalFaceCART', to detect nose and mouth use their
names like 'nose' or 'mouth'
FDetect =
vision.CascadeObjectDetector('FrontalFaceCART','MergeThreshold',16');
%Read the input images
path='viola';
fileinfo = dir(fullfile(path,'*.jpg'));
filesnumber=size(fileinfo);
for i = 1 : filesnumber(1,1)
images{i} = imread(fullfile(path,fileinfo(i).name));
    disp(['Loading image No :   ' num2str(i) ]);
end;
%Returns Bounding Box values based on number of objects
for i = 1 : filesnumber(1,1)
BB{i}=step(FDetect, (imread(fullfile(path,fileinfo(i).name))));
    disp(['BB :   ' num2str(i) ]);
end;
```

```matlab
% find number of empty BB and index of them
c=0;
for  i = 1 : filesnumber(1,1)
    if  isempty(BB{i})
        c=c+1;
        indexempty(c)=i;
    end;
end;
% replace the empty cells with bounding box
for  i = 1 : c
BB{indexempty(i)}=[40 60 180 180];
end;
%Removing other founded faces and keep just frist face or box
for  i = 1 : filesnumber(1,1)
    BB{i}=BB{i}(1,:);
end;
%% Croping the Bounding box(face)
for i = 1 : filesnumber(1,1)
croped{i}=imcrop(images{i},BB{i});
    disp(['Croped :   ' num2str(i) ]);
end;
%% Enhance the contrast of an intensity image using histogram equalization.
%rgb to gray convertion
for i = 1 : filesnumber(1,1)
hist{i}=rgb2gray(croped{i});
    disp(['To Gray :   ' num2str(i) ]);
end;
%imadjust
for i = 1 : filesnumber(1,1)
adjusted{i}=imadjust(hist{i});
    disp(['Image Adjust :   ' num2str(i) ]);
end;
%% resize the final image size
for i = 1 : filesnumber(1,1)
resized{i}=imresize(croped{i}, [30 40]);
    disp(['Image Resized :   ' num2str(i) ]);
end;
%% montage plot
montage(images); title('Originals');
figure;
montage(resized); title('Cropped');
```

---

Note:

- *Viola & Jones algorithm could be used for detecting almost anything by learning decent number of images from that specific object to the system.*

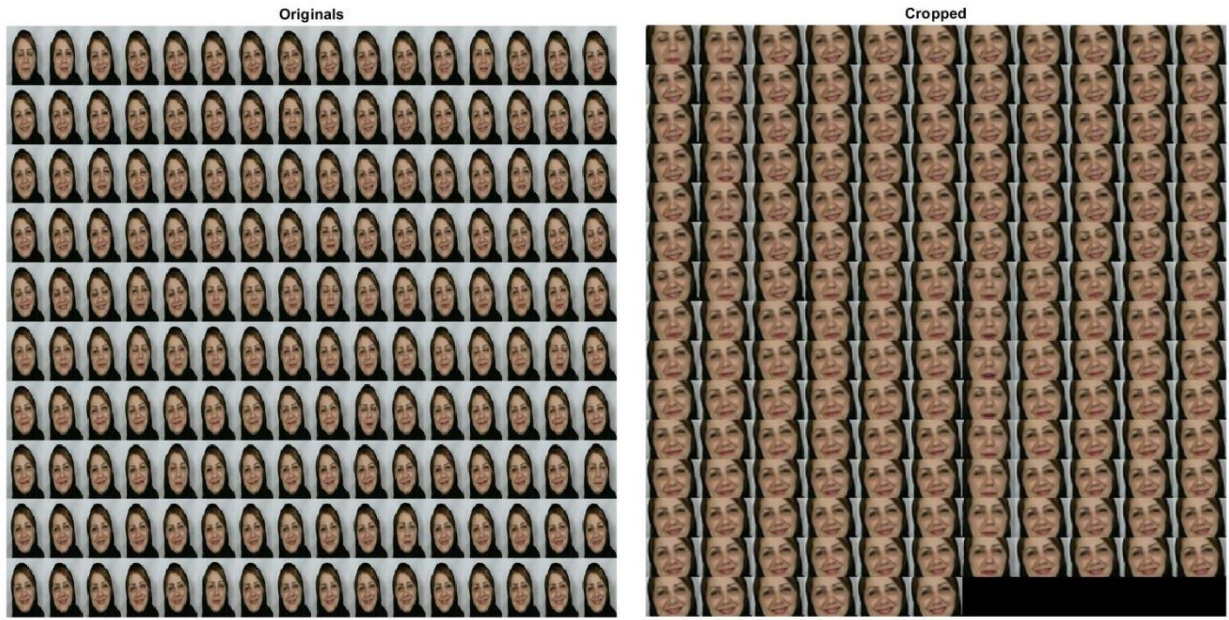

Figure 4.9 Viola and Jones face detection and extraction result

## 4.8 Exercises

1. (P1): Name the muscles which controls sadness and anger and also their related action units from FACS.

2. (P2): What is the exact different between FER and FMER? And how exactly depth sensors could help to capture micro expressions changes.

3. (P3): Write a code that code detect and extract mouth, eyes and face from "viola" folder of the book folder and finally show them before and after process. In relation to face detection, the classifier of Viola & Jones should be "Local Binary Pattern".

(Help: https://uk.mathworks.com/help/vision/ref/vision.cascadeobjectdetector-system-object.html)

4. (P4): Write a code to detect just pens.
Help:
https://uk.mathworks.com/help/vision/ref/vision.cascadeobjectdetector-system-object.html
https://uk.mathworks.com/help/vision/ref/traincascadeobjectdetector.html
https://uk.mathworks.com/help/vision/ug/train-a-cascade-object-detector.html

# *Chapter 5*

# *Feature Extraction and Selection*

# *Chapter  5*

# *Feature Extraction and Selection*

## Contents

This chapter is all about features. Features in both spatial and frequency domains for color and depth images will explained and how to combined different features' result to have a specific feature vector for each sample of dataset. Finally, outliers or unwanted or less desirable features should be removed to have smaller feature matrix which increases the classification speed and decreases classification error.

### 5.1 Feature Detection and Extraction in Spatial Domain (Color and Depth)

This book mentions and explains most importance features of spatial domain which could be extracted from an image file and provides more valuable data. Features are main distinguish factor for detecting or recognizing in the scene. There are different types of them which are working based on different parameter and sensitive to different factors on the scene. Factors such as changing in contrast and illumination. Changing in edge, blurring, sharpening and other factors. Feature extraction method selects based on type of image and environment in the image. Therefore, selecting best feature extraction method aids recognition accuracy. As it mentioned before, a nice pre-processing could remove outliers and increases the chance of getting best features possible.

- LBP

Local Binary Pattern (LBP) [57] is a color and texture-based feature and it is very nice feature for texture analysis. This feature introduced as a 3*3 rectangle for start and has good resistance against different illumination levels. So, it is used to reduce the effect of illumination changes in the experiment.

In dealing with face analysis that each face is different with another, local and texture-based features like LBP are so useful. Obviously with adding more features to the final feature vector, learning time increases which a solution is made for this purpose to have as higher accuracy as possible along with as lowest runtime speed as possible. LBP is one of the most famous local features which is using in different illumination conditions. That is why this feature is used as all databases does not have the same illumination levels. Using "extractLBPFeatures "function, it is possible to extract LBP features out of color and depth images in Matlab. Figure 5.1 shows the performance steps of LBP algorithm.

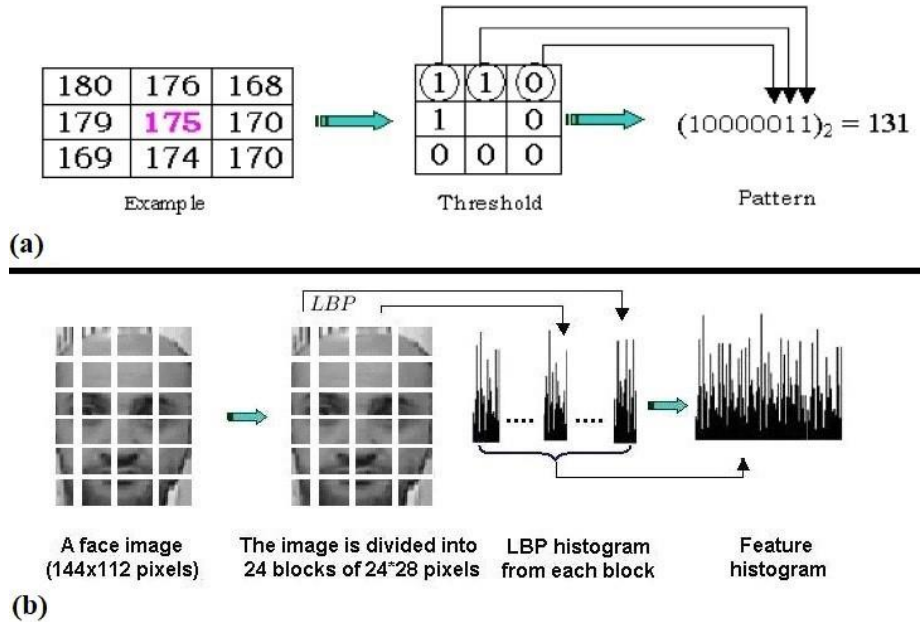

Figure 5.1 LBP algorithm workflow (a) and applying on a sample

Following lines of codes extracts LBP features from some samples of VAP RGBDT database [58]. Here Thermal images are used but color or depth images could be used too. VAP database samples are available in the book folder. After process, a matrix under the name of "lbpfeature" will be created which each row shows extracted features of an image and columns shows number of features based on "CellSize" parameter.

```
% LBP features
% Code name : "c.5.1.m"
%% Read the input images
clear;
path='VAP\thermal';
fileinfo = dir(fullfile(path,'*.jpg'));
filesnumber=size(fileinfo);
for i = 1 : filesnumber(1,1)
images{i} = imread(fullfile(path,fileinfo(i).name));
    disp(['Loading image No :   ' num2str(i) ]);
end;
% RGB to gray convertion
for i = 1 : filesnumber(1,1)
gray{i}=rgb2gray(images{i});
    disp(['To Gray :   ' num2str(i) ]);
end;
%% Extract lbp feature
for i = 1 : filesnumber(1,1)
% Decreasing "CellSize" increases matrix size but more features extracts
lbp{i} = extractLBPFeatures(gray{i},'CellSize',[8 8]);
    disp(['Extract LBP :   ' num2str(i) ]);
```

```
end;
% Converting feature vectors to a feature matrix
for i = 1 : filesnumber(1,1)
    lbpfeature(i,:)=lbp{i};
    disp(['to matrix :    ' num2str(i) ]);
end;
```

---

*Important Note:*

- *Feature vector is a 1\*n array for extracted features form a sample image. The value of "n" is variable by changing the parameter of the feature.*

- *Feature matrix is combination of feature vectors. In a feature matrix, each row shows extracted features from an image. So, and for example a feature matrix with 40\*80000 dimensions has 40 samples or images and for each sample or images, there are 80000 extracted features based on specific feature's parameters.*

- *Decreasing "CellSize" amount for example to [4 4], increases accuracy and extracts more features as increasing final feature matrix size and computational speed.*

---

- HOG

There is another type of features which are based on edge, place and angle of the pixels. It is possible to extract these features using image gradients. They are Histogram of Oriented Gradient (HOG) [59] features. These features are local, just like LBP. As these features are perfect to extract face wrinkles edges, it is rationale to employed it.

In edge-based features which are possible to get by gradient of the image, useful information is extracted from angles and position of the connected pixels. HOG features are in horizontal, vertical and diagonal directions. HOG features are extracting from blocks with different sizes. These blocks have two values of magnitude and direction. Magnitude determines the scale of the block and direction determines the path which that specific edge follows. Figure 5.2 shows HOG feature gradient magnitude and direction on a sample.

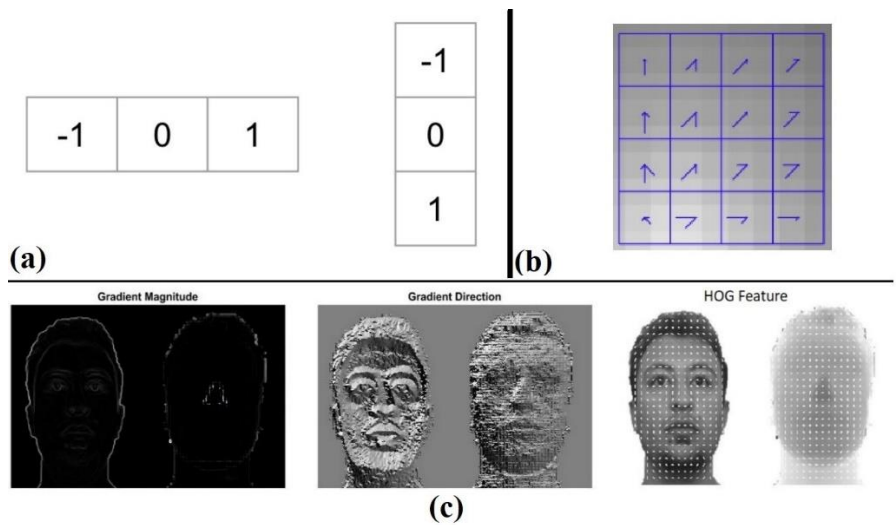

(a) (b) (c)

Figure 5.2 Gradient kernel (a), gradient directions (b) and gradient magnitude and directions on a sample (c)

Following lines of codes extracts LBP features from some samples of VAP RGBDT database [58]. Here color images are used. "extractHOGFeatures" function is responsible to extract hog features and store it in feature matrix of 'hogfeature'.

```matlab
% HOG features
% Code name : "c.5.2.m"
%% Read the input images
clear;
path='VAP\color';
fileinfo = dir(fullfile(path,'*.jpg'));
filesnumber=size(fileinfo);
for i = 1 : filesnumber(1,1)
images{i} = imread(fullfile(path,fileinfo(i).name));
    disp(['Loading image No :   ' num2str(i) ]);
end;
% RGB to gray convertion
for i = 1 : filesnumber(1,1)
gray{i}=rgb2gray(images{i});
    disp(['To Gray :   ' num2str(i) ]);
end;
%% Extract hog feature
for i = 1 : filesnumber(1,1)
% Decreasing "CellSize" increases matrix size but more features extracts
hog{i} = extractHOGFeatures(gray{i},'CellSize',[16 16]);
    disp(['Extract HOG :   ' num2str(i) ]);
end;
% Converting feature vectors to a feature matrix
for i = 1 : filesnumber(1,1)
    hogfeature(i,:)=hog{i};
    disp(['to matrix :   ' num2str(i) ]);
end;
```

- SURF

Speeded Up Robust Features (SURF) [60] is a feature detector and descriptor algorithm. SURF is so fast algorithm and has great resistance against rotation. First, image integral calculates just like Harr method. Then, feature points using Hessian algorithm [61] will be found. Making scale space is the third step. Determining maximum point is next step. Finally feature vector will be make using preview step. SURF is the advanced version of SIFT [] but it is faster multiple times.

Following lines of codes extracts SURF features out of samples from depth images of VAP RGBDT database and store it in "SURF" feature matrix.

```matlab
% SURF features
% Code name : "c.5.3.m"
clear;
% Load image data
imset = imageSet('VAP/depth','recursive');
% Extracting SURF features
% Create a bag-of-features from the image database
bag = bagOfFeatures(imset,'VocabularySize',200,...
    'PointSelection','Detector');
% Encode the images as new features
```

```
SURF = encode(bag,imset);
```

**5.2 Feature Detection and Extraction in Frequency Domain (Color and Depth)**

3.3   LPQ

If the image has unwanted smoothing, it is needed to use frequency domain features. As a lot of images in different databases have a lot of blurring or average filtering, it is rational to add frequency domain features to fix blurring effect in final feature vector. Local Phase Quantization (LPQ) [62] feature is used on depth images in this research as this feature is just like HOG on color images and extracts huge amount of edge features.

LPQ is a local feature in frequency domain based on Fourier transform system [63]. Blurring effect in magnitude and phase of frequency domain has different effect. Phase channel could deactivate low pass smoothing filters which might be in some of the images. This feature is perfect to use on depth images. Figure 5.3 represents the process of LPQ algorithm workflow. As it is clear in the Figure 5.4, LPQ method has robust performance in dealing with low pass gaussian smoothing filter. In the figure, standard deviation is (left side) and 0.5-1.5 (right side).

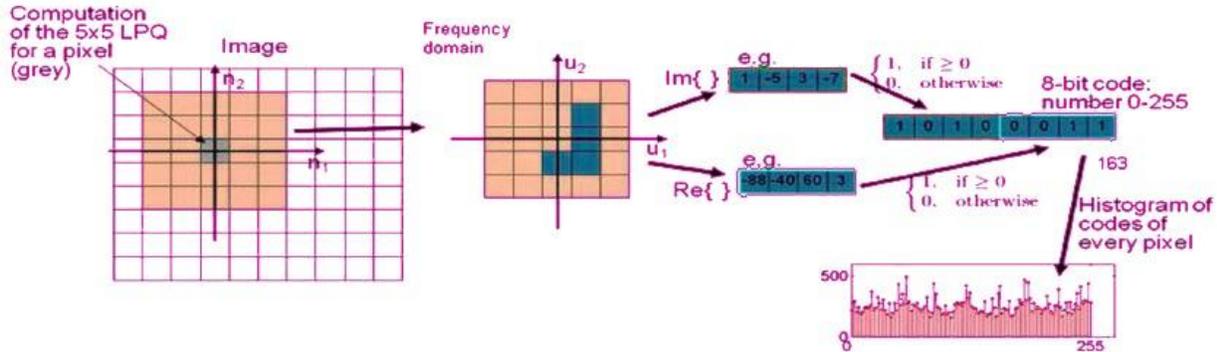

Figure 5.3 LPQ algorithm workflow

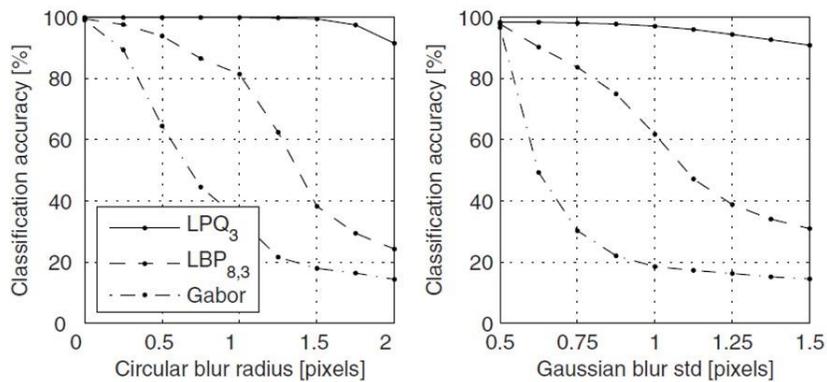

Figure 5.4 Blurring effect with different sigma on Gabor filter, LBP and LPQ algorithms

Following lines of codes, takes some depth samples from VAP RGBDT dataset and calls "lpq.m" function to extract these features from frequency domain for 40 images and saves them in "lpq" feature matrix.

```matlab
% Calling "lpq.m" function
% Code name : "c.5.4.m"
%% Read the input images
clear;
path='VAP\depth';
fileinfo = dir(fullfile(path,'*.png'));
filesnumber=size(fileinfo);
for i = 1 : filesnumber(1,1)
images{i} = imread(fullfile(path,fileinfo(i).name));
    disp(['Loading image No :   ' num2str(i) ]);
end;
%% calculating Local phase quantization feature
% winsize= should be an odd number greater than 3.
% the bigger the number the more accuracy
winsize=39;
for i = 1 : filesnumber(1,1)
f{i}=lpq(images{i},winsize);
    disp(['No of LPQ :   ' num2str(i) ]);
end;
for i = 1 : filesnumber(1,1)
    lpq(i,:)=f{i};
end;
```

Function is:

```matlab
% "lpq.m" function which extracts LPQ features
% Code name : "lpq.m"
function LPQdesc = lpq(img,winSize,decorr,freqestim,mode)
% Defaul parameters
% Local window size
if nargin<2 || isempty(winSize)
    winSize=3; % default window size 3
end
% Decorrelation
if nargin<3 || isempty(decorr)
    decorr=1; % use decorrelation by default
end
rho=0.90; % Use correlation coefficient rho=0.9 as default
% Local frequency estimation (Frequency points used [alpha,0], [0,alpha],
[alpha,alpha], and [alpha,-alpha]
if nargin<4 || isempty(freqestim)
    freqestim=1; %use Short-Term Fourier Transform (STFT) with uniform window
by default
end
STFTalpha=1/winSize;  % alpha in STFT approaches (for Gaussian derivative
alpha=1)
sigmaS=(winSize-1)/4; % Sigma for STFT Gaussian window (applied if
freqestim==2)
sigmaA=8/(winSize-1); % Sigma for Gaussian derivative quadrature filters
(applied if freqestim==3)
% Output mode
if nargin<5 || isempty(mode)
    mode='nh'; % return normalized histogram as default
end
% Other
```

```matlab
convmode='valid'; % Compute descriptor responses only on part that have full
neigborhood. Use 'same' if all pixels are included (extrapolates image with
zeros).
%% Check inputs
if size(img,3)~=1
    error('Only gray scale image can be used as input');
end
if winSize<3 || rem(winSize,2)~=1
    error('Window size winSize must be odd number and greater than equal to
3');
end
if sum(decorr==[0 1])==0
    error('decorr parameter must be set to 0->no decorrelation or 1-
>decorrelation. See help for details.');
end
if sum(freqestim==[1 2 3])==0
    error('freqestim parameter must be 1, 2, or 3. See help for details.');
end
if sum(strcmp(mode,{'nh','h','im'}))==0
    error('mode must be nh, h, or im. See help for details.');
end
%% Initialize
img=double(img); % Convert image to double
r=(winSize-1)/2; % Get radius from window size
x=-r:r; % Form spatial coordinates in window
u=1:r; % Form coordinates of positive half of the Frequency domain (Needed
for Gaussian derivative)
%% Form 1-D filters
if freqestim==1 % STFT uniform window
    % Basic STFT filters
    w0=(x*0+1);
    w1=exp(complex(0,-2*pi*x*STFTalpha));
    w2=conj(w1);
elseif freqestim==2 % STFT Gaussian window (equals to Gaussian quadrature
filter pair)
    % Basic STFT filters
    w0=(x*0+1);
    w1=exp(complex(0,-2*pi*x*STFTalpha));
    w2=conj(w1);
    % Gaussian window
    gs=exp(-0.5*(x./sigmaS).^2)./(sqrt(2*pi).*sigmaS);
    % Windowed filters
    w0=gs.*w0;
    w1=gs.*w1;
    w2=gs.*w2;
    % Normalize to zero mean
    w1=w1-mean(w1);
    w2=w2-mean(w2);
elseif freqestim==3 % Gaussian derivative quadrature filter pair
    % Frequency domain definition of filters
    G0=exp(-x.^2*(sqrt(2)*sigmaA)^2);
    G1=[zeros(1,length(u)),0,u.*exp(-u.^2*sigmaA^2)];
    % Normalize to avoid small numerical values (do not change the phase
response we use)
    G0=G0/max(abs(G0));
    G1=G1/max(abs(G1));
    % Compute spatial domain correspondences of the filters
```

```
    w0=real(fftshift(ifft(ifftshift(G0)))));
    w1=fftshift(ifft(ifftshift(G1)));
    w2=conj(w1);
    % Normalize to avoid small numerical values (do not change the phase
response we use)
    w0=w0/max(abs([real(max(w0)),imag(max(w0))]));
    w1=w1/max(abs([real(max(w1)),imag(max(w1))]));
    w2=w2/max(abs([real(max(w2)),imag(max(w2))]));
end
%% Run filters to compute the frequency response in the four points. Store
real and imaginary parts separately
% Run first filter
filterResp=conv2(conv2(img,w0.',convmode),w1,convmode);
% Initilize frequency domain matrix for four frequency coordinates (real and
imaginary parts for each frequency).
freqResp=zeros(size(filterResp,1),size(filterResp,2),8);
% Store filter outputs
freqResp(:,:,1)=real(filterResp);
freqResp(:,:,2)=imag(filterResp);
% Repeat the procedure for other frequencies
filterResp=conv2(conv2(img,w1.',convmode),w0,convmode);
freqResp(:,:,3)=real(filterResp);
freqResp(:,:,4)=imag(filterResp);
filterResp=conv2(conv2(img,w1.',convmode),w1,convmode);
freqResp(:,:,5)=real(filterResp);
freqResp(:,:,6)=imag(filterResp);
filterResp=conv2(conv2(img,w1.',convmode),w2,convmode);
freqResp(:,:,7)=real(filterResp);
freqResp(:,:,8)=imag(filterResp);
% Read the size of frequency matrix
[freqRow,freqCol,freqNum]=size(freqResp);

%% If decorrelation is used, compute covariance matrix and corresponding
whitening transform
if decorr == 1
    % Compute covariance matrix (covariance between pixel positions x_i and
x_j is rho^||x_i-x_j||)
    [xp,yp]=meshgrid(1:winSize,1:winSize);
    pp=[xp(:) yp(:)];
    dd=dist(pp,pp');
    C=rho.^dd;
    % Form 2-D filters q1, q2, q3, q4 and corresponding 2-D matrix operator M
(separating real and imaginary parts)
    q1=w0.'*w1;
    q2=w1.'*w0;
    q3=w1.'*w1;
    q4=w1.'*w2;
    u1=real(q1); u2=imag(q1);
    u3=real(q2); u4=imag(q2);
    u5=real(q3); u6=imag(q3);
    u7=real(q4); u8=imag(q4);
    M=[u1(:)';u2(:)';u3(:)';u4(:)';u5(:)';u6(:)';u7(:)';u8(:)'];
    % Compute whitening transformation matrix V
    D=M*C*M';
    A=diag([1.000007 1.000006 1.000005 1.000004 1.000003 1.000002 1.000001
1]); % Use "random" (almost unit) diagonal matrix to avoid multiple
eigenvalues.
```

```matlab
    [U,S,V]=svd(A*D*A);
    % Reshape frequency response
    freqResp=reshape(freqResp,[freqRow*freqCol,freqNum]);
    % Perform whitening transform
    freqResp=(V.'*freqResp.').';
    % Undo reshape
    freqResp=reshape(freqResp,[freqRow,freqCol,freqNum]);
end
%% Perform quantization and compute LPQ codewords
LPQdesc=zeros(freqRow,freqCol); % Initialize LPQ code word image (size
depends whether valid or same area is used)
for i=1:freqNum
    LPQdesc=LPQdesc+(double(freqResp(:,:,i))>0)*(2^(i-1));
end
%% Switch format to uint8 if LPQ code image is required as output
if strcmp(mode,'im')
    LPQdesc=uint8(LPQdesc);
end
%% Histogram if needed
if strcmp(mode,'nh') || strcmp(mode,'h')
    LPQdesc=hist(LPQdesc(:),0:255);
end
%% Normalize histogram if needed
if strcmp(mode,'nh')
    LPQdesc=LPQdesc/sum(LPQdesc);
end
```

- Gabor Filter

Gabor filters [64] are so common in face analysis applications. This feature reveal face wrinkles very well. These features are not sensitive to rotation, resizing and illumination changes. Gabor filter is based on texture just like LBP and is robust against low pass filters. Gabor filters widely used in texture analysis and edge detection. This filter is linear and local. Gabor filter convolution core is based on an exponential linear function in a gaussian one. If they be adjusted very well, they could have very precise performance. They have great response into sudden changes which makes these filters very good in face analysis. Their main advantages are in change in illumination, rotation and resizing. Below re Gabor filter parameters.

- Sigma

Standard deviation which is used in gaussian function. Sigma shows the changes width in the wave form.

- Theta

Wavelet direction angle. The most important parameter and determines to which features should be responded.

- Lambda

The size of sine wave length.

- Gamma

How much elliptical wavelet is and 1 means a circular gaussian function.

- Psi

Phase changes during time.

Gabor filter complex, real and imaginary equations are as follow:

$$Complex = \ g(x,y;\lambda,\theta,\psi,\sigma,\gamma) = exp\left(-\frac{x'^2 + \gamma^2 y'^2}{2\sigma^2}\right) exp\left(i\left(2\pi\frac{x'}{\lambda} + \psi\right)\right) \qquad (1)$$

$$Real = g(x, y; \lambda, \theta, \psi, \sigma, \gamma) = exp\left(-\frac{x'^2 + \gamma^2 y'^2}{2\sigma^2}\right)\cos\left(2\pi\frac{x'}{\lambda} + \psi\right) \tag{2}$$

$$Imaginary = g(x, y; \lambda, \theta, \psi, \sigma, \gamma) = exp\left(-\frac{x'^2 + \gamma^2 y'^2}{2\sigma^2}\right)\sin\left(2\pi\frac{x'}{\lambda} + \psi\right) \tag{3}$$

Where,

$$x' = x \cos\theta + y \sin\theta \tag{4}$$
$$y' = x \sin\theta + y \cos\theta \tag{5}$$

Figure 5.5 show two images in color and depth form and in the frequency domain which a gaussian filter with sigma=3 is applied on them. As it is clear in this figure, low pass filtering in depth image and in frequency domain has weaker effect compared with color image. Amplitude and phase spectrums in frequency domain has slight change in depth image in the figure, which means using frequency domain features on depth image is a rational effort. Figure 5.6 shows Gabor filter in different frequency and directions in 2-Dimentional (2-D) and 3-Dimentional (3-D) forms (a), Gaussian kernel in Gabor filter with 30 Degree of sin (b) and Gabor filter with wave length of 8 and in directions of 0, 22, 45, 67, 90, 112, 135 and 180 degrees. Figure 5.7 represents LBP, LPQ, HOG, SURF and Gabor filter features (machine understanding) on a sample in color and depth forms.

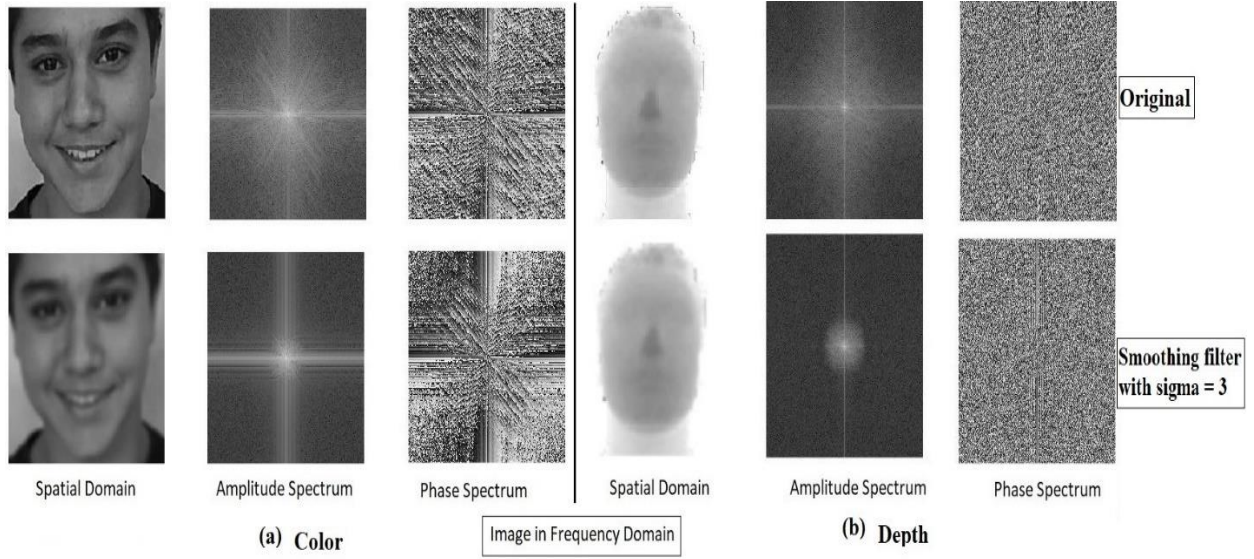

Figure 5.5 Blurring effect on color (a) and depth (b) images in the frequency domain with similar sigma amount.

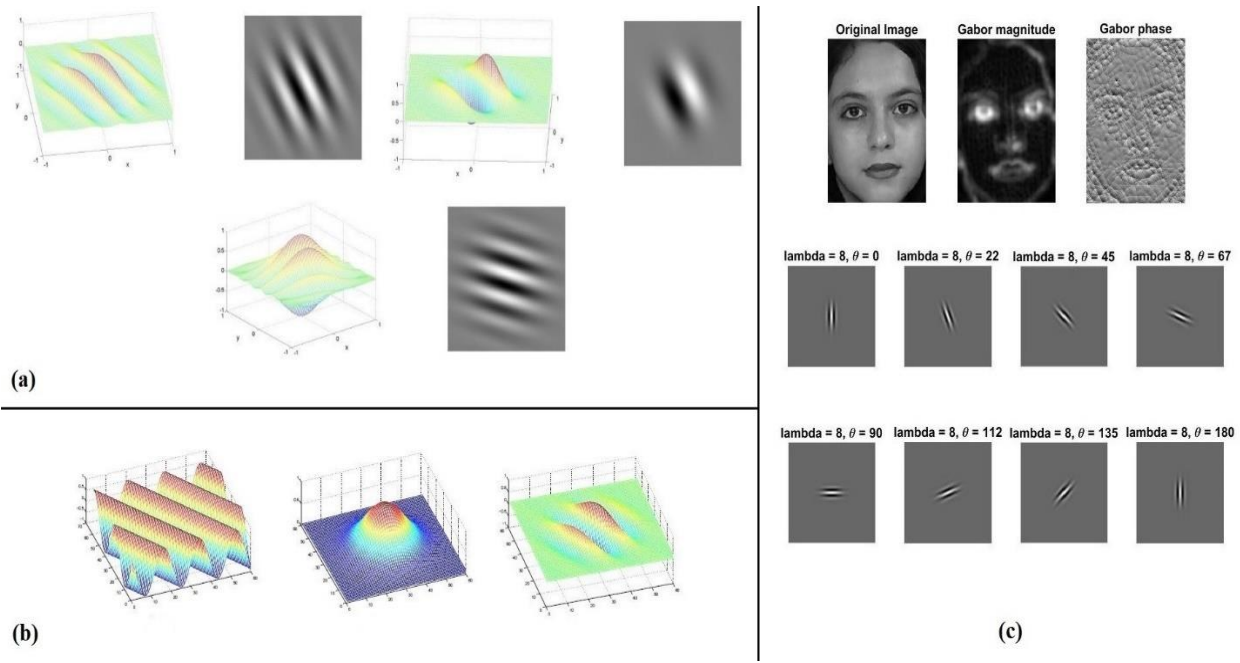

Figure 5.6 Gabor filter in different freqency and directions (a), Gabor filter in sin waveform in 30 degree (b), Gabor filter with wveleanght of 8 in different directions on a sample color image

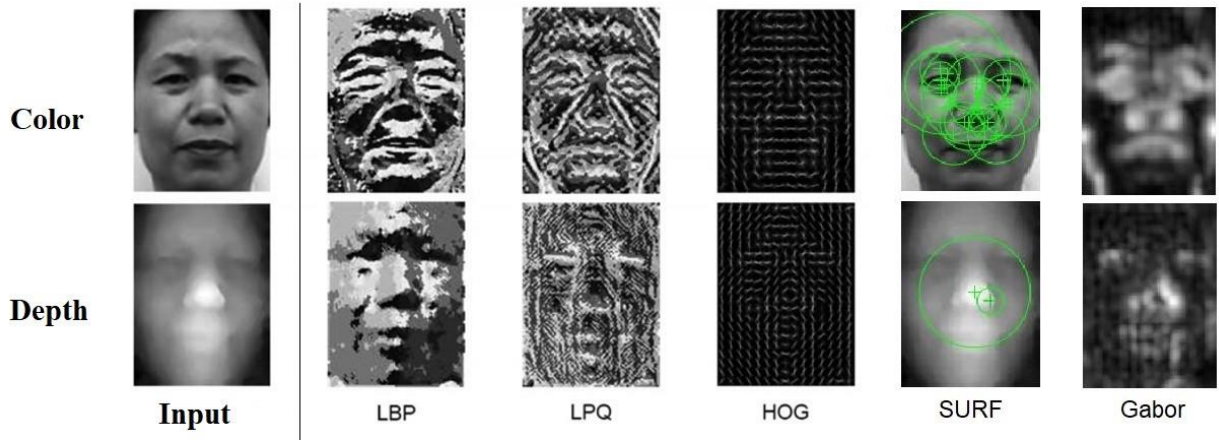

Figure 5.7 Appling 5 main features on a sample face image in color and depth forms, showing machine understanding of different features in spatial and frequency domains [65]

Following lines of codes calls "gaborFeatures.m" and "gaborFilterBank.m" functions to extract Gabor features out of some thermal samples from VAP RGBDT database. Files are in "VAP" folder in the book main folder. Final features stores in a matrix called "Gaborvector".

```
% Calling "gaborFeatures.m" and "gaborFilterBank.m" functions
% Code name : "c.5.5.m"
%% Read the input images
clear;
```

```
path='VAP\thermal';
fileinfo = dir(fullfile(path,'*.jpg'));
filesnumber=size(fileinfo);
for i = 1 : filesnumber(1,1)
images{i} = imread(fullfile(path,fileinfo(i).name));
    disp(['Loading image No :   ' num2str(i) ]);
end;
% RGB to gray convertion
for i = 1 : filesnumber(1,1)
gray{i}=rgb2gray(images{i});
    disp(['To Gray :   ' num2str(i) ]);
end;
%% extract gabor features
for i = 1 : filesnumber(1,1)
gaborArray = gaborFilterBank(5,8,39,39);  % Generates the Gabor filter bank
featureVector{i} = gaborFeatures(gray{i},gaborArray,8,8);   % Extracts Gabor
feature vector, 'featureVector', from the image, 'img'.
disp(['Extracting Gabor Vector :   ' num2str(i) ]);
end;
for i = 1 : filesnumber(1,1)
    Gaborvector(i,:)=featureVector{i};
    disp(['to matrix :   ' num2str(i) ]);
end;
```

"gaborFeatures.m" function is:

```
% Gabor Features
% Code name : "gaborFeatures.m"
function featureVector = gaborFeatures(img,gaborArray,d1,d2)

% GABORFEATURES extracts the Gabor features of the image.
% It creates a column vector, consisting of the image's Gabor features.
% The feature vectors are normalized to zero mean and unit variance.
%
%
% Inputs:
%       img       :   Matrix of the input image
%       gaborArray  :   Gabor filters bank created by the function
gaborFilterBank
%       d1        :   The factor of downsampling along rows.
%                     d1 must be a factor of n if n is the number of rows
in img.
%       d2        :   The factor of downsampling along columns.
%                     d2 must be a factor of m if m is the number of
columns in img.
%
% Output:
%       featureVector  :   A column vector with length (m*n*u*v)/(d1*d2).
%                          This vector is the Gabor feature vector of an
%                          m by n image. u is the number of scales and
%                          v is the number of orientations in 'gaborArray'.
%
%
% Sample use:
%
% % img = imread('cameraman.tif');
```

```matlab
% % gaborArray = gaborFilterBank(5,8,39,39);  % Generates the Gabor filter
bank
% % featureVector = gaborFeatures(img,gaborArray,4,4);   % Extracts Gabor
feature vector, 'featureVector', from the image, 'img'.
if (nargin ~= 4)     % Check correct number of arguments
    error('Use correct number of input arguments!')
end
if size(img,3) == 3 % Check if the input image is grayscale
    img = rgb2gray(img);
end
img = double(img);
%% Filtering
% Filter input image by each Gabor filter
[u,v] = size(gaborArray);
gaborResult = cell(u,v);
for i = 1:u
    for j = 1:v
        gaborResult{i,j} = conv2(img,gaborArray{i,j},'same');
        % J{u,v} = filter2(G{u,v},I);
    end
end

%% Feature Extraction
% Extract feature vector from input image
[n,m] = size(img);
s = (n*m)/(d1*d2);
l = s*u*v;
featureVector = zeros(l,1);
c = 0;
for i = 1:u
    for j = 1:v

        c = c+1;
        gaborAbs = abs(gaborResult{i,j});
        gaborAbs = downsample(gaborAbs,d1);
        gaborAbs = downsample(gaborAbs.',d2);
        gaborAbs = reshape(gaborAbs.',[],1);
        % Normalized to zero mean and unit variance. (if not applicable,
please comment this line)
        gaborAbs = (gaborAbs-mean(gaborAbs))/std(gaborAbs,1);

        featureVector(((c-1)*s+1):(c*s)) = gaborAbs;
    end
end
%% Show filtered images
% % Show real parts of Gabor-filtered images
% figure('NumberTitle','Off','Name','Real parts of Gabor filters');
% for i = 1:u
%     for j = 1:v
%         subplot(u,v,(i-1)*v+j)
%         imshow(real(gaborResult{i,j}),[]);
%     end
% end
%
%
% % Show magnitudes of Gabor-filtered images
% figure('NumberTitle','Off','Name','Magnitudes of Gabor filters');
% for i = 1:u
```

```
%      for j = 1:v
%            subplot(u,v,(i-1)*v+j)
%            imshow(abs(gaborResult{i,j}),[]);
%      end
% end
```

"gaborFilterBank.m" function is:

```
% Generating Gabor Filters
% Code name : "gaborFilterBank.m"
function gaborArray = gaborFilterBank(u,v,m,n)

% GABORFILTERBANK generates a custum Gabor filter bank.
% It creates a u by v array, whose elements are m by n matries;
% each matrix being a 2-D Gabor filter.
%
%
% Inputs:
%       u   :  No. of scales (usually set to 5)
%       v   :  No. of orientations (usually set to 8)
%       m   :  No. of rows in a 2-D Gabor filter (an odd integer number
usually set to 39)
%       n   :  No. of columns in a 2-D Gabor filter (an odd integer number
usually set to 39)
%
% Output:
%       gaborArray: A u by v array, element of which are m by n
%                   matries; each matrix being a 2-D Gabor filter
%
%
% Sample use:
%
% gaborArray = gaborFilterBank(5,8,39,39);

if (nargin ~= 4)      % Check correct number of arguments
    error('There should be four inputs.')
end
%% Create Gabor filters
% Create u*v gabor filters each being an m*n matrix
gaborArray = cell(u,v);
fmax = 0.25;
gama = sqrt(2);
eta = sqrt(2);

for i = 1:u

    fu = fmax/((sqrt(2))^(i-1));
    alpha = fu/gama;
    beta = fu/eta;

    for j = 1:v
        tetav = ((j-1)/v)*pi;
        gFilter = zeros(m,n);

        for x = 1:m
            for y = 1:n
                xprime = (x-((m+1)/2))*cos(tetav)+(y-((n+1)/2))*sin(tetav);
```

```matlab
                yprime = -(x-((m+1)/2))*sin(tetav)+(y-((n+1)/2))*cos(tetav);
                gFilter(x,y) = (fu^2/(pi*gama*eta))*exp(-
((alpha^2)*(xprime^2)+(beta^2)*(yprime^2)))*exp(1i*2*pi*fu*xprime);
            end
        end
        gaborArray{i,j} = gFilter;
    end
end
%% Show Gabor filters
% Show magnitudes of Gabor filters:
% figure('NumberTitle','Off','Name','Magnitudes of Gabor filters');
% for i = 1:u
%     for j = 1:v
%         subplot(u,v,(i-1)*v+j);
%         imshow(abs(gaborArray{i,j}),[]);
%     end
% end
%
% Show real parts of Gabor filters:
% figure('NumberTitle','Off','Name','Real parts of Gabor filters');
% for i = 1:u
%     for j = 1:v
%         subplot(u,v,(i-1)*v+j);
%         imshow(real(gaborArray{i,j}),[]);
%     end
% end
```

### 5.3 Feature Selection or Dimensionality Reduction

There are multiple methods and techniques in order to feature selection or dimensionality reduction, but in this book, two of best algorithms that are PCA and Lasso are selected and used.

- PCA

Principal Component Analysis (PCA) [66] is a quantitatively rigorous method for achieving this simplification. The method generates a new set of variables, called principal components. Each principal component is a linear combination of the original variables. All the principal components are orthogonal to each other, so there is no redundant information. The principal components as a whole form an orthogonal basis for the space of the data. "coeff = pca(X)" returns the principal component coefficients, also known as loadings, for the n-by-p data matrix X. Rows of X correspond to observations and columns correspond to variables. The coefficient matrix is p-by-p. Each column of "coeff" contains coefficients for one principal component, and the columns are in descending order of component variance. By default, "pca" centers the data and uses the Singular Value Decomposition (SVD) algorithm [67]. "[coeff,score,latent] = pca(___)" also returns the principal component scores in score and the principal component variances in latent. You can use any of the input arguments in the previous syntaxes. Principal component scores are the representations of X in the principal component space. Rows of score correspond to observations, and columns correspond to components. The principal component variances are the eigenvalues of the covariance matrix of X.

Below code selects principal components out of "hogfeature" matrix. This matrix is extracted hog features out of "jaffe" dataset. The matrix is labeled for classification task with seven main facial expressions using numbers 1 to 7 placed in the last column of the matrix. This matrix is available under the name of "hogfeatures.mat' as a mat file in the book main folder.

```
% PCA feature selection
% Code name : "c.5.6.m"
clear;
feature = load('hogfeature.mat');
hogfeature=feature.hogfeature;
[coeff,score,latent] = pca(hogfeature)
classificationLearner
```

There is app in Matlab in the APP tab which is called "classification learner". This app explains in the next chapter but totally, you can use PCA here too. As it is clear in Figure 5.8, just by using 93 % of data or using 89 features out of 211, there was 2% more accuracy using SVM classifier. It has to be mentioned that classifiers are explained in the next chapter.

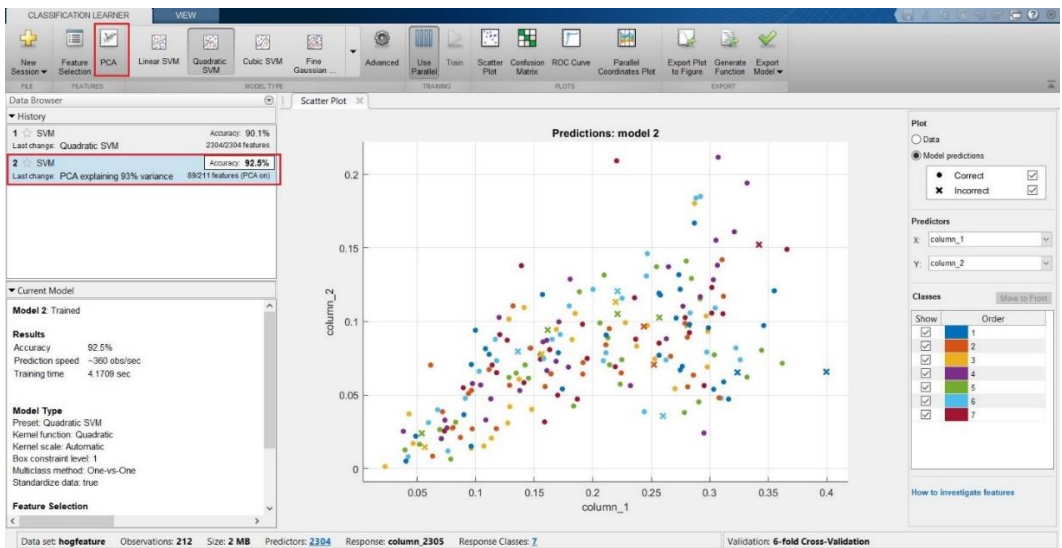

Figure 5.8 Using PCA in the classification learner

• Lasso

Lasso [68] is a regularization [68] technique for estimating generalized linear models. Lasso includes a penalty term that constrains the size of the estimated coefficients. Therefore, it resembles Ridge Regression [69]. Lasso is a shrinkage estimator: it generates coefficient estimates that are biased to be small. Nevertheless, a lasso estimator can have smaller error than an ordinary maximum likelihood estimator when you apply it to new data. Unlike ridge regression, as the penalty term increases, the lasso technique sets more coefficients to zero. This means that the lasso estimator is a smaller model, with fewer predictors. As such, lasso is an alternative to stepwise regression and other model selection and dimensionality reduction techniques. Standard linear regression works by estimating a set of coefficients that minimize the sum of the squared error between the observed values and the fitted values from the model. Regularization techniques like ridge regression, lasso, and the elastic net introduce an additional term to this minimization problem.

Regularization methods and feature selection techniques both have unique strengths and weaknesses. Let's close this blog posting with some practical guidance regarding pros and cons for the various techniques. Regularization techniques have two major advantages compared to feature selection.

Regularization techniques are able to operate on much larger datasets than feature selection methods. Lasso and ridge regression can be applied to datasets that contains thousands - even tens of thousands of variables. Even sequential feature selection is usually too slow to cope with this many possible predictors.

Regularization algorithms often generate more accurate predictive models than feature selection. Regularization operates over a continuous space while feature selection operates over a discrete space. As a result, regularization is often able to fine tune the model and produce more accurate estimates. However, feature selection methods also have their advantages

Regularization techniques are only available for a small number of model types. Notably, regularization can be applied to linear regression and logistic regression. However, if you're working some other modeling technique - say a boosted decision tree - you'll typically need to apply feature selection techniques.

Feature selection is easier to understand and explain to third parties. Never underestimate the importance of being able to describe your methods when sharing your results. With this said and done, each of the three regularization techniques also offers its own unique advantages and disadvantages. Because lasso uses an L1 norm it tends to force individual coefficient values completely towards zero. As a result, lasso works very well as a feature selection algorithm. It quickly identifies a small number of key variables.

In contrast, ridge regression uses an L2 norm for the coefficients (you're minimizing the sum of the squared errors). Ridge regression tends to spread coefficient shrinkage across a larger number of coefficients. If you think that your model should contain a large number of coefficients, ridge regression is probably a better choice than lasso. Last, but not least, we have the elastic net which is able to compensate for a very specific limitation of lasso. Lasso is unable to identify more predictors than you have coefficients.

Following lines of codes, reads "jaffe" dataset and extracts faces. Then, extract hog features and makes it ready for lasso feature selection by labeling each row for all seven main expressions. Finally, lasso selects most valuable features out of "hogfeatures" matrix and labels it for classification task. Finally, result is stores in lasso matrix. As it is clear from comparing of "lasso" and "hogfeatures" matrixes, number of features decreases significantly in "lasso" matrix without considerable accuracy loss. Final matrix is stored in book main folder under the name of "lasso.mat". Figure 5.9 shows the performance of lasso on "jaffe" dataset.

```matlab
% Lasso feature selection
% Code name : "c.5.7.m"
clear;
%Detect objects using Viola-Jones Algorithm
%To detect Face
FDetect = vision.CascadeObjectDetector;
%Read the input images
path='jaffe';
fileinfo = dir(fullfile(path,'*.jpg'));
filesnumber=size(fileinfo);
for i = 1 : filesnumber(1,1)
images{i} = imread(fullfile(path,fileinfo(i).name));
    disp(['Loading image No :   ' num2str(i) ]);
end;
```

```matlab
%% Extract hog feature
for i = 1 : filesnumber(1,1)
hog{i} = extractHOGFeatures(images{i},'CellSize',[16 16]);
    disp(['Extract hog :   ' num2str(i) ]);
end;
for i = 1 : filesnumber(1,1)
    hogfeature(i,:)=hog{i};
    disp(['to matrix :   ' num2str(i) ]);
end;
%% Lasso Regularization Algorithm
% Labeling for lasso
label(1:31,1)=1;
label(31:61,1)=2;
label(62:90,1)=3;
label(91:122,1)=4;
label(123:152,1)=5;
label(153:182,1)=6;
label(183:212,1)=7;
% Construct the lasso fit by using 5-fold cross-validation
% Set of coefficients B that models label as a function of hogfeature
% By default, lasso will create 100 different models. Each model was
%estimated
% using a slightly larger lambda. All of the model coefficients are stored in
%array
% B. The rest of the information about the model is stored in a structure
named Stats.
[B Stats] = lasso(hogfeature,label, 'CV', 5);
disp(B(:,1:5))
disp(Stats)
% lassoPlot generates a plot that displays the relationship between lambda
and the
% cross validated mean square error (MSE) of the resulting model. Each of the
% red dots show the MSE for the resulting model. The vertical line segments
% stretching out from each dot are error bars for each estimate.
lassoPlot(B, Stats, 'PlotType', 'CV')
ds.Lasso = B(:,Stats.IndexMinMSE);
disp(ds)
sizemfcc=size(hogfeature);
temp=1;
for i=1:sizemfcc(1,2)
if ds.Lasso(i)~=0
lasso(:,temp)=hogfeature(:,i);
temp=temp+1;
end;
end;
% Labeling feature extracted matrix for classification
sizefinal=size(lasso);
sizefinal=sizefinal(1,2);
lasso(1:31,sizefinal+1)=1;
lasso(31:61,sizefinal+1)=2;
lasso(62:90,sizefinal+1)=3;
lasso(91:122,sizefinal+1)=4;
lasso(123:152,sizefinal+1)=5;
lasso(153:182,sizefinal+1)=6;
lasso(183:212,sizefinal+1)=7;
```

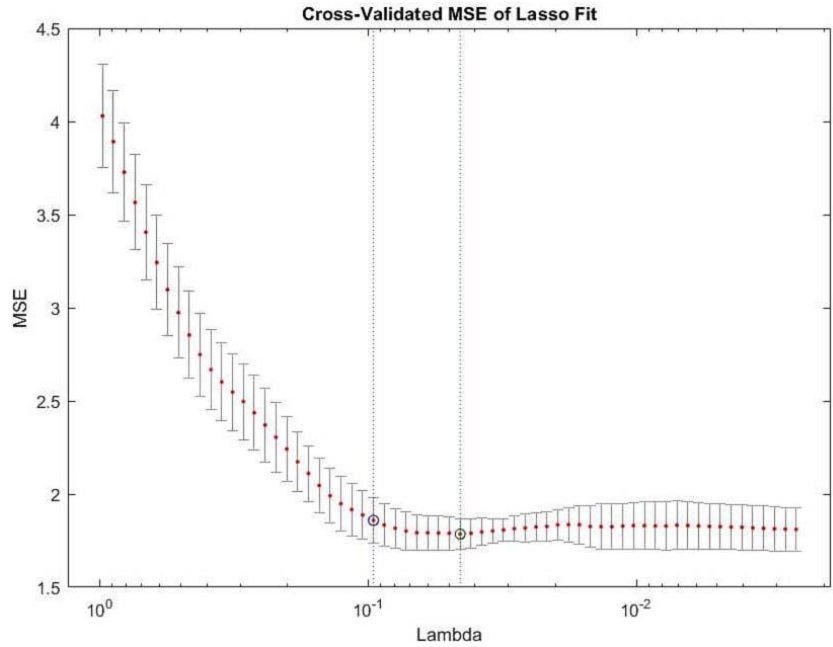

Figure 5.9 the performance of lasso on "jaffe" dataset using cross-validated MSE

## 5.4 Exercises

1. (P3): Write a code to extract Gabor and SURF feature from depth images of VAP database in the main book folder and combine their feature matrixes.

2. (P4): Write a code to extract SIFT features out of "jaffe" dataset and use any evolutionary feature selection to reduce dimensions size.
Help:
https://uk.mathworks.com/matlabcentral/fileexchange/50319-sift-feature-extreaction
https://uk.mathworks.com/matlabcentral/fileexchange/52973-evolutionary-feature-selection

# *Chapter   6*

# *Classification*

# *Chapter 6*

# *Classification*

## Contents

This chapter includes covering 6 famous and main classification classifiers and how to validate inputs and targets in train and test phases. In supervised classification, inputs are main data and targets are class labels or simply classes. The division between train and test data, normally is 70% train and 30% test. Also, different plots for final train a test result are considered which shows the final recognition in a better way.

### 6.1 Classification Classifiers

Machine learning (ML) [17, 18, 19, 20] is the study of computer algorithms that improve automatically through experience and by the use of data. Machine learning algorithms build a model based on sample data, known as "training data", in order to make predictions or decisions without being explicitly programmed. A classifier is a machine learning model that is used to discriminate different objects based on certain features. Most useful and commonly used classification classifiers are decision tree [70], discriminant analysis [71], naive bayes [72], Support Vector Machines [73], K- Nearest Neighborhood (K-NN) and ensemble classifiers [74]. It has to be mentioned that these are known as classifier classes which the most effective of each one is used in the book. A supervised machine learning algorithm is one that relies on labeled input data to learn a function that produces an appropriate output when given new unlabeled data [29]. In unsupervised learning, data is not labeled, so it is not classification anymore and it is called clustering [29]. In semi-supervised learning, part of data is labeled and part is not. Here, just supervised learning is used as facial expressions are known to us. So, it is possible to label data with its related expression label like, sadness, joy, anger and more. Figure 6.1 shows the general form these classifiers.

## 1. *Decision tree*

Decision Trees (DTs) are a non-parametric supervised learning method used for classification and regression [69]. The goal is to create a model that predicts the value of a target variable by learning simple decision rules inferred from the data features. A tree can be seen as a piecewise constant approximation. Some advantages of decision trees are:

- Simple to understand and to interpret. Trees can be visualized.
- Requires little data preparation. Other techniques often require data normalization, dummy variables need to be created and blank values to be removed. Note however that this module does not support missing values.
- The cost of using the tree is logarithmic in the number of data points used to train the tree.
- Able to handle both numerical and categorical data.
- Able to handle multi-output problems.
- Uses a white box model. If a given situation is observable in a model, the explanation for the condition is easily explained by Boolean logic.

The disadvantages of decision trees include:

- Decision-tree learners can create over-complex trees that do not generalize the data well. This is called overfitting. Mechanisms such as pruning, setting the minimum number of samples required at a leaf node or setting the maximum depth of the tree are necessary to avoid this problem.
- Decision trees can be unstable because small variations in the data might result in a completely different tree being generated. This problem is mitigated by using decision trees within an ensemble.
- There are concepts that are hard to learn because decision trees do not express them easily, such as XOR, parity or multiplexer problems.
- Decision tree learners create biased trees if some classes dominate. It is therefore recommended to balance the dataset prior to fitting with the decision tree.

## 2. *Linear discriminant analysis*

Linear discriminant analysis (LDA), Normal Discriminant Analysis (NDA), or discriminant function analysis is a generalization of Fisher's linear discriminant [75], a method used in statistics and other fields, to find a linear combination of features that characterizes or separates two or more classes of objects or events. The resulting combination may be used as a linear classifier, or, more commonly, for dimensionality reduction before later classification.

Quadratic Discriminant Analysis (QDA) [76] is a variant of LDA in which an individual covariance matrix is estimated for every class of observations. QDA is particularly useful if there is prior knowledge that individual classes exhibit distinct covariances. A disadvantage of QDA is that it cannot be used as a dimensionality reduction technique.

## 3. *Support vector machines*

Support vector machines (SVMs) [86] are a set of supervised learning methods used for classification, regression and outliers detection.

The advantages of support vector machines are:

- Effective in high dimensional spaces.
- Effective in cases where number of dimensions is greater than the number of samples.
- Uses a subset of training points in the decision function (called support vectors), so it is also memory efficient.

- Versatile: different Kernel functions can be specified for the decision function. Common kernels are provided, but it is also possible to specify custom kernels.

The disadvantages of support vector machines include:

- If the number of features is much greater than the number of samples, avoid over-fitting in choosing Kernel functions and regularization term is crucial.
- SVMs do not directly provide probability estimates, these are calculated using an expensive five-fold cross-validation.

### 4.   *K- Nearest Neighborhood*

The k-nearest neighbors (K-NN) algorithm is a simple, easy-to-implement supervised machine learning algorithm that can be used to solve both classification and regression problems. In both cases, the input consists of the k closest training examples in data set. The output depends on whether K-NN is used for classification or regression.

In K-NN, classification is computed from a simple majority vote of the nearest neighbors of each point: a query point is assigned the data class which has the most representatives within the nearest neighbors of the point.

**Pros:**
- Easy to use.
- Quick calculation time.
- Does not make assumptions about the data.

**Cons:**
- Accuracy depends on the quality of the data.
- Must find an optimal k value (number of nearest neighbors).
- Poor at classifying data points in a boundary where they can be classified one way or another.

### 5.   *Naive Bayes*

Naive Bayes methods are a set of supervised learning algorithms based on applying Bayes' theorem [77] with the "naive" assumption of conditional independence between every pair of features given the value of the class variable. The calculation of Bayes Theorem can be simplified by making some assumptions, such as each input variable is independent of all other input variables. Although a dramatic and unrealistic assumption, this has the effect of making the calculations of the conditional probability tractable and results in an effective classification model referred to as Naive Bayes.

Naive Bayes classifiers are highly scalable, requiring a number of parameters linear in the number of variables (features/predictors) in a learning problem.

Naive Bayes algorithms are mostly used in sentiment analysis, spam filtering, recommendation systems etc. They are fast and easy to implement but their biggest disadvantage is that the requirement of predictors to be independent.

### 6.   *Ensemble classifiers*

The word ensemble is a Latin-derived word which means "union of parts". The regular classifiers that are used often are prone to make errors. As much as these errors are inevitable, they can be reduced with the proper construction of a learning classifier. Ensemble learning is a way of generating various base classifiers from which a new classifier is derived which performs better than any constituent classifier.

These base classifiers may differ in the algorithm used, hyperparameters, representation or the training set [78]. The key objective of the ensemble methods is to reduce bias and variance.

Matlab provides boosted tree, bagged tree, subspace discriminant, subspace K-NN, and RUSBossted trees ensemble classifiers [78], which some of them will be employed.

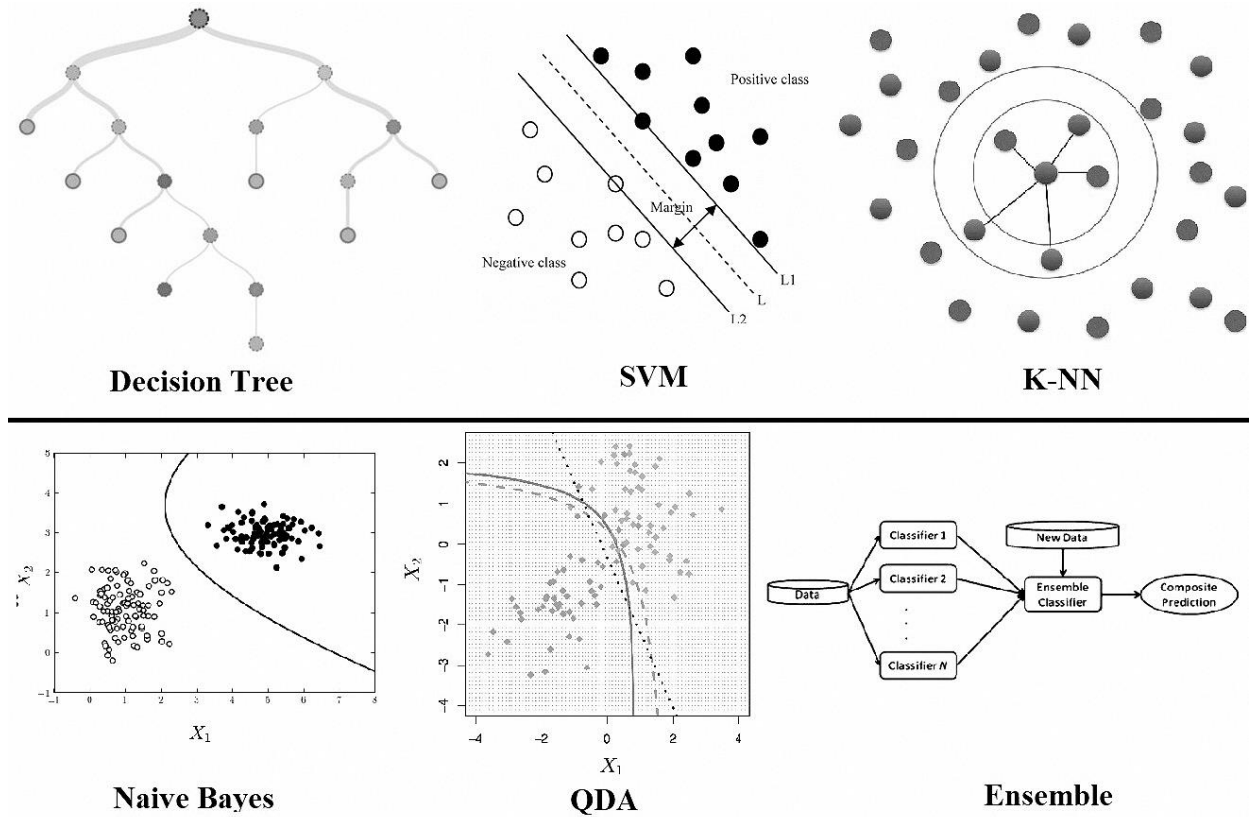

**Decision Tree**        **SVM**        **K-NN**

**Naive Bayes**        **QDA**        **Ensemble**

Figure 6.1 General forms of commonly use classifiers

### 6.1.1    Cross Validation, Inputs and Targets

Cross-validation is a statistical method used to estimate the skill of machine learning models [79]. It is commonly used in applied machine learning to compare and select a model for a given predictive modeling problem because it is easy to understand, easy to implement, and results in skill estimates that generally have a lower bias than other methods.

Cross-validation is a resampling procedure used to evaluate machine learning models on a limited data sample. The procedure has a single parameter called k that refers to the number of groups that a given data sample is to be split into. As such, the procedure is often called k-fold cross-validation. When a specific value for k is chosen, it may be used in place of k in the reference to the model, such as k=10 becoming 10-fold cross-validation. Cross-validation is primarily used in applied machine learning to estimate the skill of a machine learning model on unseen data. That is, to use a limited sample in order to estimate how the model is expected to perform in general when used to make predictions on data not used during the training of the model. It is a popular method because it is simple to understand and because it generally results in a less biased or less optimistic estimate of the model skill than other methods, such as a simple train/test split.

The general procedure is as follows:

1.  Shuffle the dataset randomly.
2.  Split the dataset into k groups

3.  For each unique group:
    1.  Take the group as a hold out or test data set
    2.  Take the remaining groups as a training data set
    3.  Fit a model on the training set and evaluate it on the test set
    4.  Retain the evaluation score and discard the model
4.  Summarize the skill of the model using the sample of model evaluation scores

Now, inputs and targets are mostly used in supervised learning. After feature extraction and feature selection (if needed), there is a feature matrix which normally is consists of multiple rows and columns. Each row presents a sample in dataset and the last value in the row is target or label of that row which defines the class of that sample. In other words, input is main data and target is class label. Sometimes input and targets are in different array or matrix which has the same effect of being in the same array or matrix.

### 6.1.2 Train and Test

In or der to create a model to validate the dataset or database, it is needed to divide the data into train and test parts. This section runs after feature extraction, feature selection, labeling and cross validation. Normally 70% of the final feature matrix belongs to train set and remaining 30% for test set, but it could be changed, but surely more percentage belongs to train set. Test set creates just for validation which means to make the system ready in facing with any new data in real life. If the system learns with a good machine learning algorithm or the classifier and train data return decent recognition accuracy in that stage, most probably it returns nice accuracy in dealing with test data. Final recognition which is considered for validation evaluation, assessment and also, considered as recognition accuracy or final result is test recognition accuracy, not trained one.

## 6.2 Scatter Plot, Confusion Matrix and ROC Curve

A scatter plot uses dots to represent values for two different numeric variables. The position of each dot on the horizontal and vertical axis indicates values for an individual data point. Scatter plots are used to observe relationships between variables.

A confusion matrix [19, 32] is a performance measurement technique for Machine learning classification. It is a kind of table which helps you to the know the performance of the classification model on a set of test data for that the true values are known. The term confusion matrix itself is very simple, but its related terminology can be a little confusing.

In Machine Learning, performance measurement is an essential task. So, when it comes to a classification problem, we can count on a Receiver Operating Characteristics (ROC) Curve. When we need to check or visualize the performance of the multi-class classification problem, we use the ROC curve. It is one of the most important evaluation metrics for checking any classification model's performance. ROC curves typically feature true positive rate on the Y axis, and false positive rate on the X axis. This means that the top left corner of the plot is the "ideal" point - a false positive rate of zero, and a true positive rate of one [19, 32, 80].

Following lines of code, reads some samples from KDEF face dataset [35]. Expressions are reduced from seven expressions to four and number of samples are reduced significantly to just 200 sampled for all expressions. Each expression has 50 samples. It has to be mentioned, using whole samples of any dataset or database require permission from data provider which we earned it and they let us to download it, so just few samples from any dataset could be provided in the book folder as we do. IKFDB is an exception as the owner is the same author as this book. After reading the data, all images resize to 256*256 dimensions and LBP features extract form them. Finally, it labels for lasso and after feature selection, last column of labels is added for four expressions numerically. The final feature matrix is "lasso". The final feature matrix under the name of "kdef lbp.mat" is stored in the book main folder. Now it is ready to be used in the "classificationLearner" app.

```matlab
% Making data ready for classification
% Code name : "c.6.1.m"
clc;
clear;
%% Read the input images
path='Four KDEF';
fileinfo = dir(fullfile(path,'*.jpg'));
filesnumber=size(fileinfo);
for i = 1 : filesnumber(1,1)
images{i} = rgb2gray(imread(fullfile(path,fileinfo(i).name)));
    disp(['Loading image No :   ' num2str(i) ]);
end;
%% Resize images
for i = 1 : filesnumber(1,1)
resized{i}=imresize(images{i}, [256 256]);
    disp(['Image Resized :   ' num2str(i) ]);
end;
%% Extract LBP features
for i = 1 : filesnumber(1,1)
    % the less cell size the more accuracy
lbp{i} = extractLBPFeatures(resized{i},'CellSize',[32 32]);
    disp(['Extract LBP :   ' num2str(i) ]);
end;
for i = 1 : filesnumber(1,1)
    lbpfeature(i,:)=lbp{i};
    disp(['to matrix :   ' num2str(i) ]);
end;

%% Lasso Regularization Algorithm
% Labeling for lasso
label(1:50,1)=1;
label(51:100,1)=2;
label(101:150,1)=3;
label(151:200,1)=4;

[B Stats] = lasso(lbpfeature,label, 'CV', 5);
disp(B(:,1:5))
disp(Stats)

lassoPlot(B, Stats, 'PlotType', 'CV')
ds.Lasso = B(:,Stats.IndexMinMSE);
disp(ds)
sizemfcc=size(lbpfeature);
temp=1;
for i=1:sizemfcc(1,2)
if ds.Lasso(i)~=0
lasso(:,temp)=lbpfeature(:,i);
temp=temp+1;
end;
end;
```

```
% Labeling feature extracted matrix for classification
% As we have four expressions of joy, anger, surprise and disgust-
% - So, four classes of 1, 2, 3 and 4 are defined for samples. -
% - Each 50 samples or feature vectors covers one class.
sizefinal=size(lasso);
sizefinal=sizefinal(1,2);
lasso(1:50,sizefinal+1)=1;
lasso(51:100,sizefinal+1)=2;
lasso(101:150,sizefinal+1)=3;
lasso(151:200,sizefinal+1)=4;
classificationLearner
```

In order to use the "classificationLearner" app, run the app from "APPS" tab under "Machine Learning and Deep Learning" section and select yellow plus button at top left corner and choose "from workspace". Then under "workspace variable" select "lasso" matrix as input data. Target, response or label selects automatically as the last column. From right, select desire amount of cross validation and push "Start Session" button. In the next window, it is possible to select your preferred classifier and select run to train. It returns training accuracy along with classifier parameters and different plots such as confusion matrix. Figure 6.2 represents "classification learner app" parts after loading data. Figure 6.3 shows classification result and related plots in the training stage for code "c.6.1.m".

---

*Note:*

- *Classification learner app could be run by compiling "classificationLearner" in the command window.*
- *Having big feature matrix leads to higher training run time. So, be sure to reduce the number of dimensions using feature selection methods.*
- *In order to acquire any dataset, contacting the dataset owner is required.*
- *Please just use datasets content which is provided in the book folder just for research purposes.*
- *IKFDB dataset owner is the writer of this book.*

---

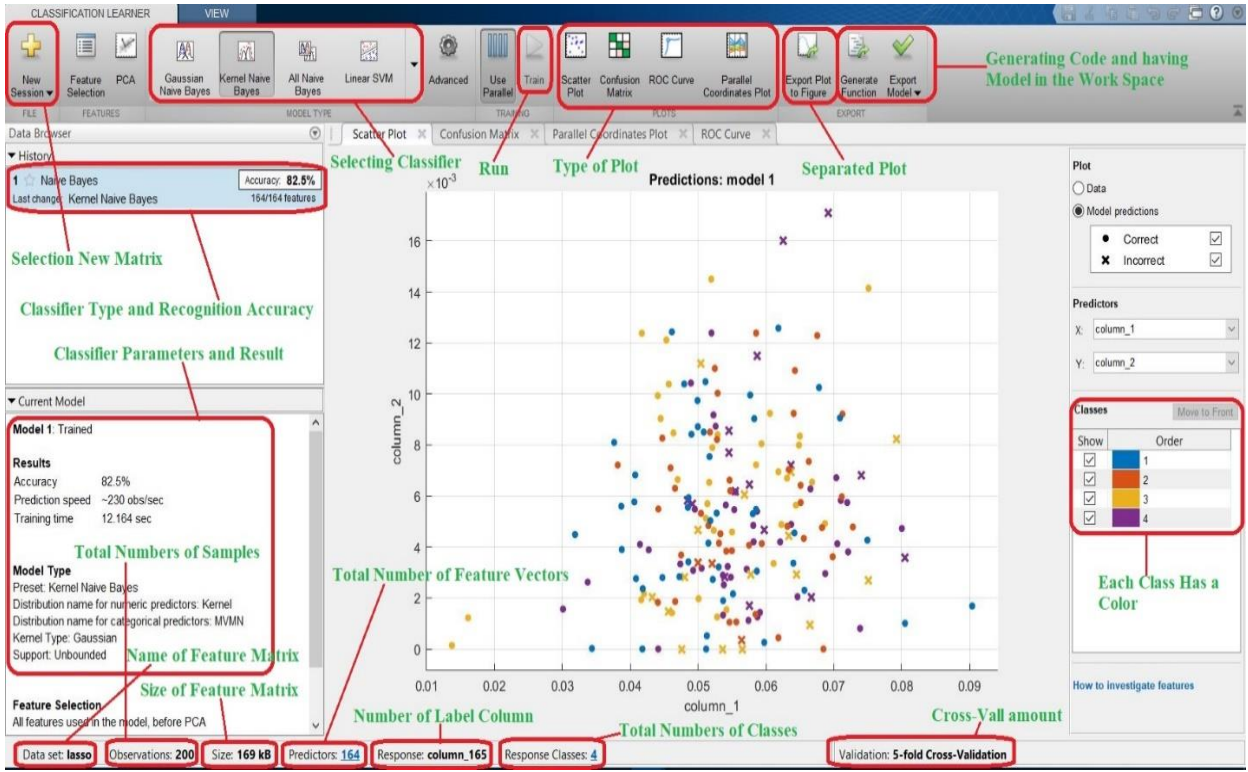

Figure 6.2 Classification learner app parts after loading data

| **Scatter Plot** | **Confusion Matrix** | **ROC Curve** | |
|---|---|---|---|

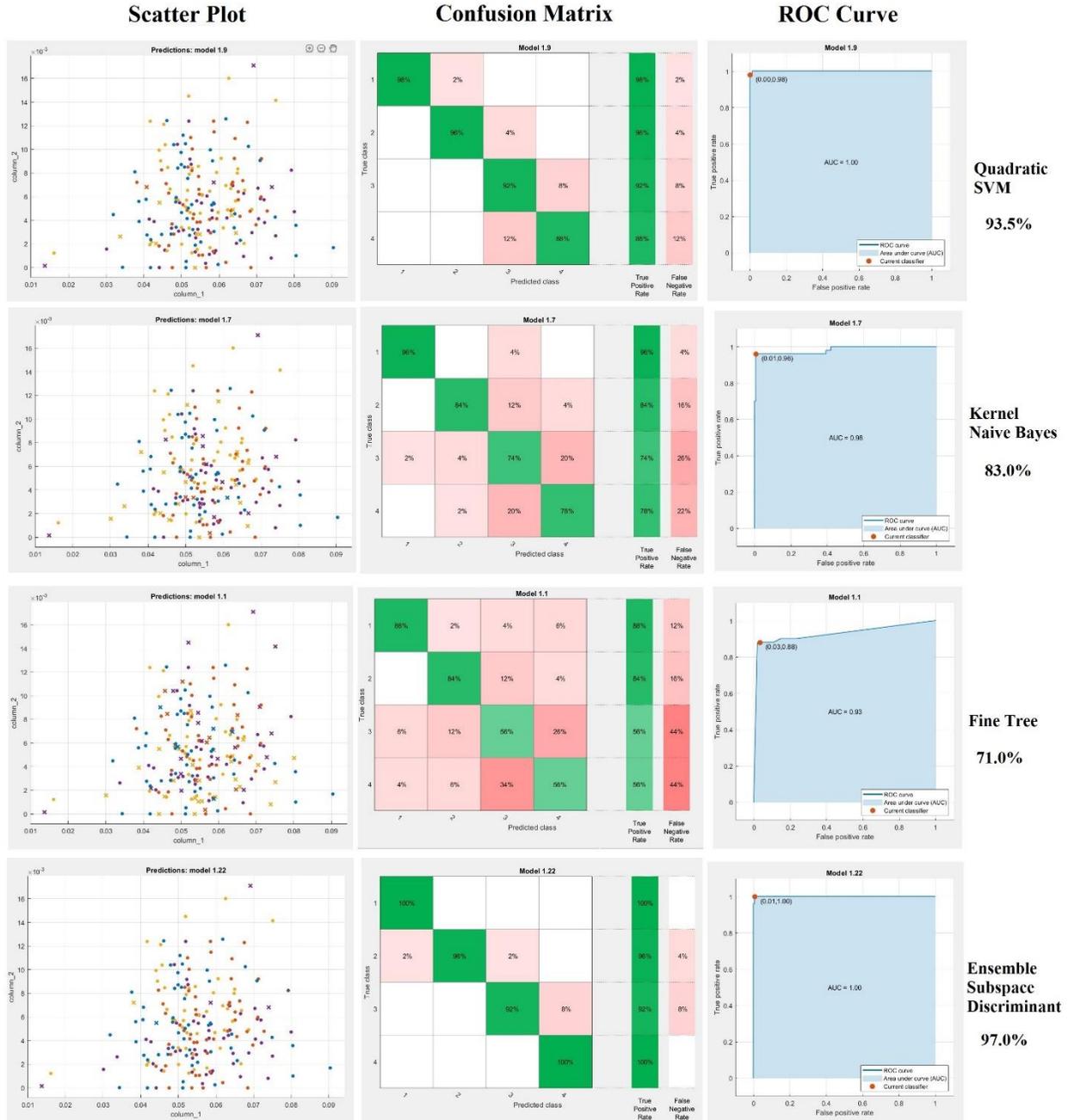

Figure 6.3 Classification result and related plots in the training stage for code "c.6.1.m"

Now, in order to test the accuracy of the system with new data or real-life data, it is needed to split data to 70% train and 30% test which is not happened in the previous code. Previous code is considered 100 % of data in the training stage. As the ready to classify data is needed, so running "c.6.1.m" is needed which provides "lasso" variable containing all classes and labels. This file is already available in the book folder and you can just load it into the workspace as the next code will do it. Code "c.6.2.m", load the feature matrix "lasso" or "kdef lbp" and trained model of data which is for shuffled 70 % of the feature matrix "lasso". If you want to do it yourself, execute following code and train the "dataTrain" matrix with any desire classifier using classification learner app. After training stage, you must export the model into the workspace in the classification learner at top right corner of the app. Here ensemble subspace discriminant with 97.5 % recognition accuracy is exported to the workspace. Normally it will suggest "trainedModel" name for saving into the workspace. Remember if you changed the models name, then it should be affected in the code "c.6.2.m".

```matlab
% Reading ready to classify feature matrix
kdeflbp=load('kdef lbp.mat');
kdeflbp=kdeflbp.lasso;
% Cross varidation (train: 70%, test: 30%)
% Data shuffles each time. So, the result is slightly different with any run
cv = cvpartition(size(kdeflbp,1),'HoldOut',0.3);
idx = cv.test;
% Separate to training and test data
dataTrain = kdeflbp(~idx,:);
dataTest  = kdeflbp(idx,:);
```

After reading data, it splits into 70 % train and 30% test and test matrix tests using trained model. Finally, final testing accuracy calculates and prints. As data shuffles in any split, recognition might have a slight change for test. At the end, loss function for train and test is calculated too. Also, confusion matrix for test data based on real facial expressions displays.

```matlab
% Testing final model using trained model
% Code name : "c.6.2.m"
clear;
% Reading ready to classify feature matrix
kdeflbp=load('kdef lbp.mat');
kdeflbp=kdeflbp.lasso;
% Reading trained model
trainedModel=load('trainedModel.mat');
trainedModel=trainedModel.trainedModel;
% Cross varidation (train: 70%, test: 30%)
% Data shuffles each time. So, the result is slightly different with any run
cv = cvpartition(size(kdeflbp,1),'HoldOut',0.3);
idx = cv.test;
% Separate to training and test data
dataTrain = kdeflbp(~idx,:);
dataTest  = kdeflbp(idx,:);
% Separating data from label in test data (removing label column)
tstdat=dataTest (:,1:end-1);
% Labels for calculating error
tstlbl=dataTest(:,end);
% Determining new test data classes
tstres = trainedModel.predictFcn(tstdat)
% Calculating final test error
err=[tstlbl tstres];
serr=size(err); serr=serr(1,1);
count=0;
for i=1 : serr
        if err(i,1) ~= err(i,2)
            count=count+1;
        end
end
Finalerr=100-((count*100)/serr)
disp(['Data shuffles each time. So, the result is slightly different with any run']);
```

```matlab
disp(['The Test Accuracy is :   ' num2str(Finalerr) '%']);
% Converting classes names from number to real facial expressions for
% confusion matrix
for i=1 : serr
        if err(i,1) == 1
            lbl{i}='Joy';
        elseif err(i,1) == 2
            lbl{i}='Surprise';
        elseif err(i,1) == 3
            lbl{i}='Anger';
        elseif err(i,1) == 4
            lbl{i}='Disgust';
        end
end
for i=1 : serr
        if err(i,2) == 1
            lbl2{i}='Joy';
        elseif err(i,2) == 2
            lbl2{i}='Surprise';
        elseif err(i,2) == 3
            lbl2{i}='Anger';
        elseif err(i,2) == 4
            lbl2{i}='Disgust';
        end
end
% Plotting Confusion matrix
cm = confusionchart(lbl,lbl2,'RowSummary','row-
normalized','ColumnSummary','column-normalized');
%Loss function (train and test)
model=trainedModel.ClassificationSVM;
Losstrain = loss(model,dataTrain(:,1:138),dataTrain(:,139))
Losstest = loss(model,dataTest(:,1:138),dataTest(:,139))
```

---

*Important Note:*

- *Achieved accuracy in this chapter is not so precise, as pre-processing stage did not take place. So, a lot of outliers were recognized as features which is not so suitable. The main goal of chapter 6 was to show how classification task works on any feature matrix. Next chapter goes the same way as this chapter but using neural networks and deep learning approaches. But, chapter 8 which is the final chapter covers all necessary processing aspects to have the most precises accuracy on a real-world data.*

---

Figure 6.4 Test confusion matrix for KDEF dataset samples (four expressions)

**6.3 Exercises**

1. (P3): Try to classify 3 classes of Joy, Anger and Neutral expressions from JAFFE dataset (available in the book folder). Feature extraction should be done with LPQ algorithm. Use lasso to for feature selection and label data accordingly. Divide data into 80% of train and 20% of test. Train the 80% training data and export the model into the work space. Use classification learner app for Naïve Baye, Cubic K-NN and Quadratic SVM with cross validation of 10. After training, just export quadratic SVM into the workspace and try to test the model with remaining 20% of data. Finally try to have, recognition accuracy, confusion and loss functions values just for the test data.

# *Chapter   7*
# *Neural   Networks   and Deep                Leaning Classification*

# *Chapter  7*

# *Neural  Networks  and  Deep  Leaning Classification*

## Contents

This chapter pays to classification using neural network algorithms. Chapter covers both shallow and deep neural network algorithms for classification. Matlab has apps to do these tasks which makes processing easier.

### 7.1 Neural Networks Classification

Neural Networks (NN) [19, 32] take inspiration from the learning process occurring in human brains. They consist of an artificial network of functions, called parameters, which allows the computer to learn, and to fine tune itself, by analyzing new data. Each parameter, sometimes also referred to as neurons, is a function which produces an output, after receiving one or multiple inputs. Those outputs are then passed to the next layer of neurons, which use them as inputs of their own function, and produce further outputs. Those outputs are then passed on to the next layer of neurons, and so it continues until every layer of neurons have been considered, and the terminal neurons have received their input. Those terminal neurons then output the final result for the model. An alternative way of thinking about a neural net is to think of it as one massive function which takes inputs and arrives at a final output. The intermediary functions, which are done by the neurons in their many layers, are usually unobserved, and thankfully automated. The mathematics behind them is as interesting as it is complex, and deserves a further look.

the neurons within the network interact with the neurons in the next layer, with every output acting as an input for a future function. Every function, including the initial neuron receives a numeric input, and produces a numeric output, based on an internalized function, which includes the addition of a bias term, which is unique for every neuron. That output is then converted to the numeric input for the function in the next layer, by being multiplied with an appropriate weight. This continues until one final output for the network is produced.

The difficulty lies in determining the optimal value for each bias term, as well as finding the best weighted value for each pass in the neural network. To accomplish this, one must choose a cost function. A cost function is a way of calculating how far a particular solution is from the best possible solution. There are many different possible cost functions, each with advantages and drawbacks, each best suited under certain conditions. Thus, the cost function should be tailored and selected based on individual research needs. Once a cost function has been determined, the neural net can be altered in a way to minimize that cost function. A simple way of optimizing the weights and bias, is therefore to simply run the network multiple times. On the first try, the predictions will by necessity be random. After each iteration, the cost function will be analyzed, to determine how the model performed, and how it can be improved. The information gotten from the cost function is then passed onto the optimizing function, which calculates new weight values, as well as new bias values. With those new values integrated into the model, the model is rerun. This is continued until no alteration improves the cost function.

There are three methods of learning: supervised, unsupervised, and reinforcement learning. The simplest of these learning paradigms is supervised learning, where the neural net is given labelled inputs. The labelled examples, are then used to infer generalizable rules which can be applied to unlabeled cases. It is the simplest learning method, since it can be thought of operating with a 'teacher', in the form of a function that allows the net to compare its predictions to the true, and desired results. Figure 7.1 shows the basic structure of artifactual neural networks and similarity to human brain neurons. Figure 7.2 shows the structure of the most useful neural networks, especially in image processing.

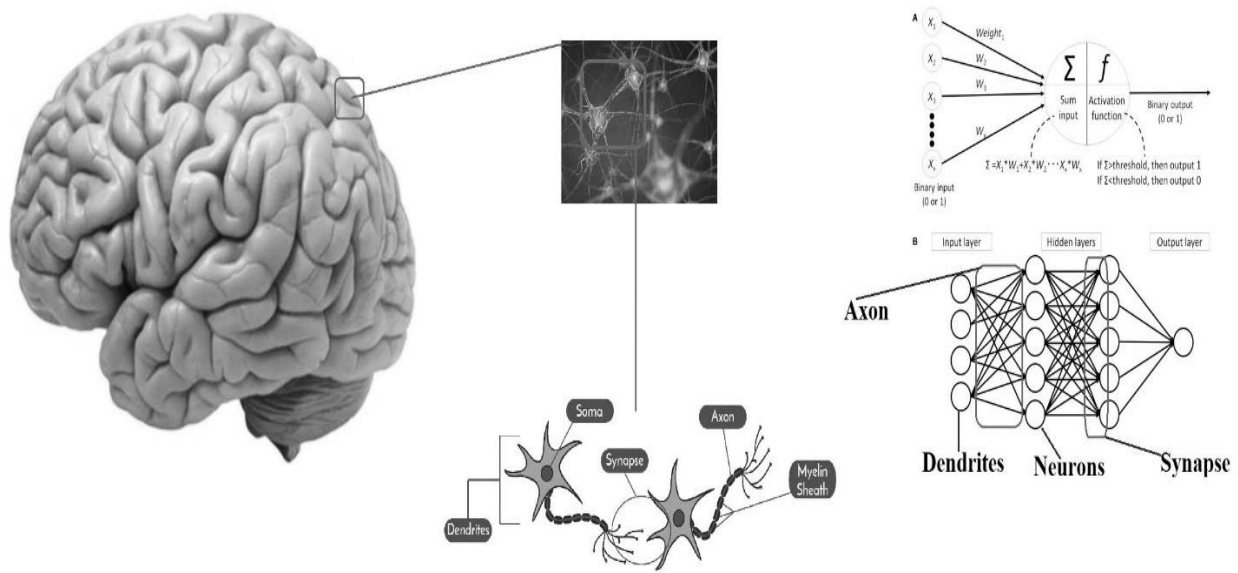

Figure 7.1 Basic structure of artifactual neural networks and similarity to human brain neurons

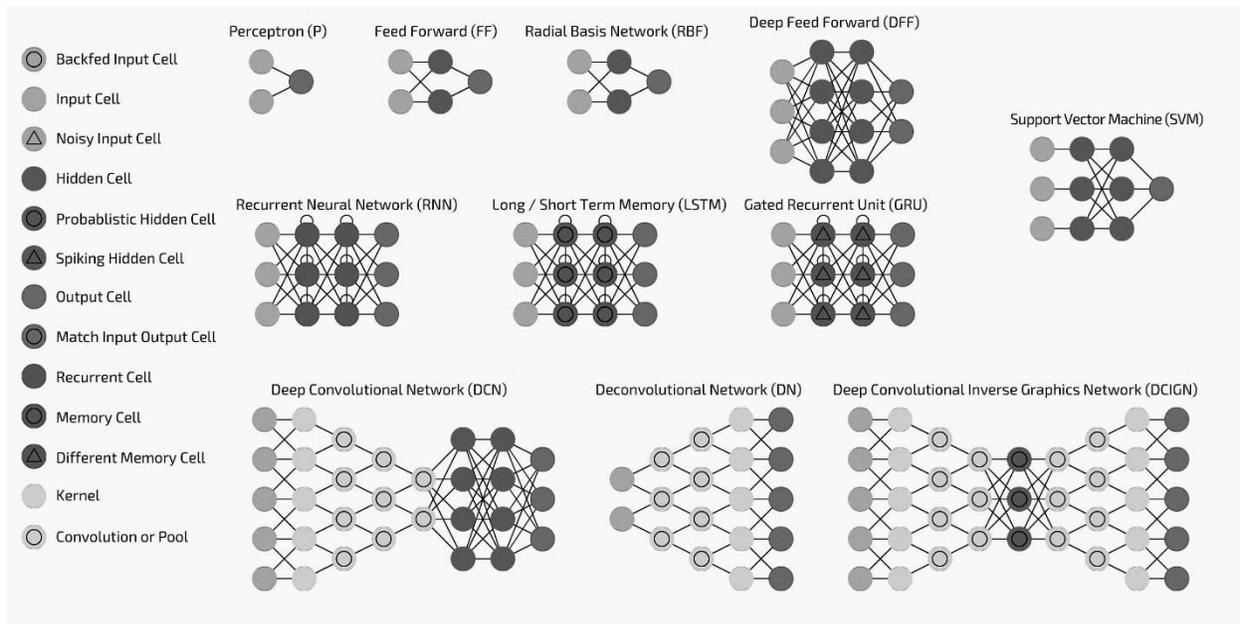

Figure 7.2 Structure of the most common neural networks

There is app in Matlab software called "Neural Net Pattern Recognition" which is for classification in shallow neural network form. It is possible to run it by opening "APPS" tab in software and selecting it under the section "Machine Learning and Deep Learning" or just compiling "nprtool" in the command window. Following lines of code, reads depth sample for 5 expressions of surprise, anger, disgust, neutral and sadness from IKFDB [32] database and after extracting LPQ features, makes it ready for neural network classification. Depth images are stored in the "Network" folder. Extracted features are stored in the book main folder under the titles of "lpq.mat", "lpqlbl2.mat" and "lpqdata.mat".

```matlab
% Making data ready for NN classification
% Code name : "c.7.1.m"
clc;
clear;
%% Read the input images
path='Network';
fileinfo = dir(fullfile(path,'*.png'));
filesnumber=size(fileinfo);
for i = 1 : filesnumber(1,1)
images{i} = imread(fullfile(path,fileinfo(i).name));
    disp(['Loading image No :  ' num2str(i) ]);
end;
%% Contrast Adjustment
for i = 1 : filesnumber(1,1)
adjusted{i}=imadjust(images{i});
    disp(['Image Adjust :  ' num2str(i) ]);
end;
%% Resize Images
for i = 1 : filesnumber(1,1)
resized{i}=imresize(adjusted{i}, [256 256]);
    disp(['Image Resized :  ' num2str(i) ]);
end;
%% Calculating Local Phase Quantization Features
% winsize= should be an odd number greater than 3.
```

```matlab
% the bigger the number the more accuracy
winsize=19;
for i = 1 : filesnumber(1,1)
tmp{i}=lpq(resized{i},winsize);
    disp(['No of LPQ :   ' num2str(i) ]);
end;
for i = 1 : filesnumber(1,1)
    lpq(i,:)=tmp{i};
end;
%% Labeling For Classification (Classification Learner APP)
sizefinal=size(lpq);
sizefinal=sizefinal(1,2);
lpq(1:50,sizefinal+1)=1;
lpq(51:100,sizefinal+1)=2;
lpq(101:150,sizefinal+1)=3;
lpq(151:200,sizefinal+1)=4;
lpq(201:250,sizefinal+1)=5;

%% Labeling For Classification (Neural Network Pattern Recognition APP)
lpqdata=lpq(:,1:end-1);
lpqlbl=lpq(:,end);
sizelpq=size(lpq);
sizelpq=sizelpq(1,1);
for i=1 : sizelpq
            if lpqlbl(i) == 1
                lpqlbl2(i,1)=1;
        elseif lpqlbl(i) == 2
                lpqlbl2(i,2)=1;
        elseif lpqlbl(i) == 3
                lpqlbl2(i,3)=1;
        elseif lpqlbl(i) == 4
                lpqlbl2(i,4)=1;
        elseif lpqlbl(i) == 5
                lpqlbl2(i,5)=1;
        end
end
nprtool
```

Figure 7.3 shows the whole process of giving data to the "Neural Net Pattern Recognition" app and performing train and test. Finally, system returns error and different validation plots such as confusion matrix. According to figure 7.3 and in step 2, "lpqdata" and "lpqlbl2" is given as inputs and targets. The splits value for train and test should be determined in the step 3. Step 4 determines number of hidden neurons which 15 is enough here, as our model is small. In the step 5, you have to push train button to achieve training result, error and validation plots.

Figure 7.4 represents some of the validation metrics results. Also, Figure 7.5 presents confusion matrixes and ROC curves for train, validation and test stages. As it is clear from figures, there is low error value and high recognition accuracy of 98 % alongside with optimized system performance with having 5 classes of facial expressions; feature extracted just by on feature from depth samples. It is quiet promising. It has to be mentioned that in a real experiment, more features alongside with pre-processing stage and more samples are involved which contains micro expressions that is hard to recognize. This matter is covered in the chapter 8, fully.

*Note:*

- *Classes names in the book folder are named with uppercases English alphabets in order to have ease of use in image reading process for input. Actually, each uppercased alphabet is one facial expression. This alphabets changes to number later for classification. For example, samples in the "Network" folder are names with A to E alphabets which later changed to numbers of 1 to 5 for classification. So, we have 1= Surprise, 2= Anger, ... and 5= Sadness.*
- *Using this book's codes without the book folder is a little hard. So, please download the book folder alongside with the book. Obviously, you can replace your images instead of book folder contents but having the contents brings ease of use.*

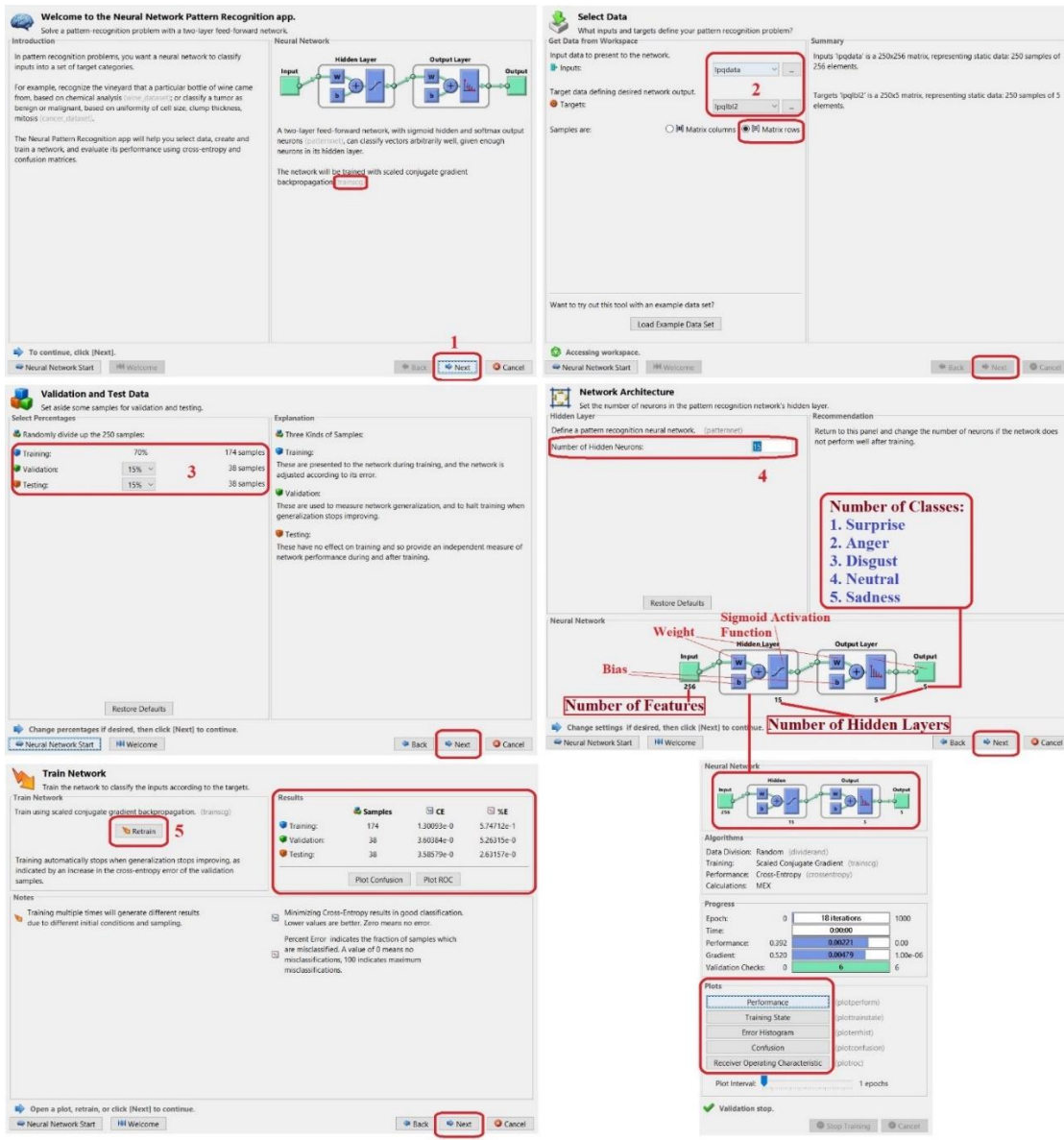

Figure 7.3 Working with "Neural Net Pattern Recognition" app, step by step

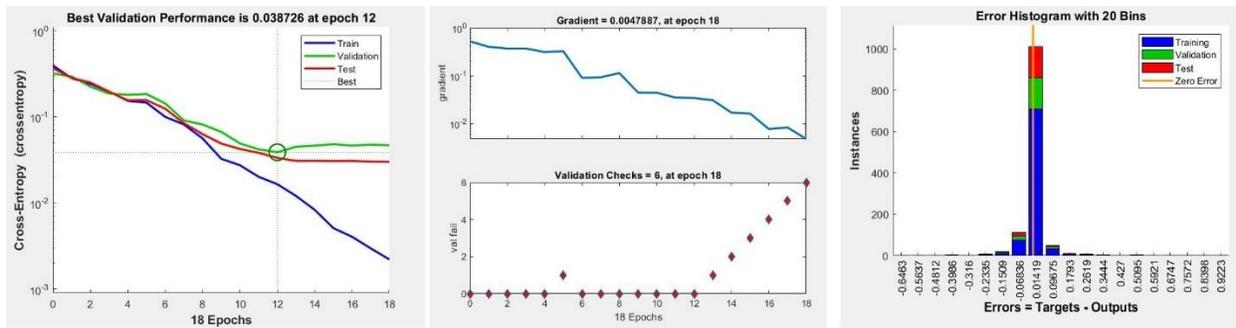

Figure 7.4 From left to right, system performance, number of epochs to achieve best result and error histogram for train, validation and test stages

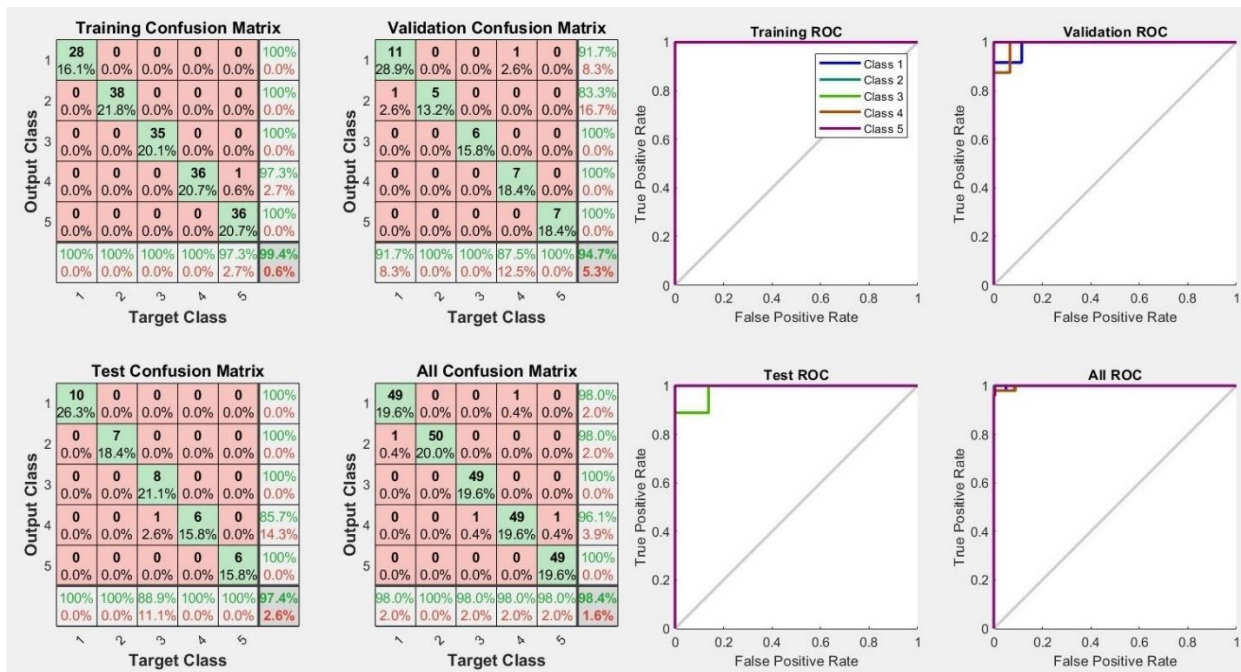

Figure 7.5 Confusion matrixes and ROC curves for all stages of the classification

Following lines of code, train and test extracted features of depth samples from last experiment but this time using five different shallow neural network algorithms of Levenberg-Marquardt [81, 82], One Step Secant [81, 82], Gradient Descent [81, 82], Scaled Conjugate Gradient and Resilient Backpropagation [81, 82] and plots their confusion matrixes and Training States respectively. In this method every this is done using code and not app. Figure 7.6 shows five algorithms confusion matrixes and Figure 7.7 shows their related training states plots.

```
% Different Shallow Neural Network Algorithms (Train and Test)
% Code name : "c.7.2.m"
clear;
% Inputs
lpqdata=load('lpqdata.mat');
lpqdata=lpqdata.lpqdata;
lpqdata=lpqdata'; % changing data shape from rows to columns
% Labels (targets)
lpqlbl2=load('lpqlbl2.mat');
```

```matlab
lpqlbl2=lpqlbl2.lpqlbl2;
lpqlbl2=lpqlbl2'; % changing data shape from rows to columns
% Defining input and target variables
inputs = lpqdata;
targets = lpqlbl2;
% Create a Pattern Recognition Network
hiddenLayerSize = 14;
net1 = patternnet(hiddenLayerSize);
net2 = patternnet(hiddenLayerSize);
net3 = patternnet(hiddenLayerSize);
net4 = patternnet(hiddenLayerSize);
net5 = patternnet(hiddenLayerSize);
% Set up Division of Data for Training, Validation, Testing
net1.divideParam.trainRatio = 70/100;
net1.divideParam.valRatio = 15/100;
net1.divideParam.testRatio = 15/100;
net2.divideParam.trainRatio = 70/100;
net2.divideParam.valRatio = 15/100;
net2.divideParam.testRatio = 15/100;
net3.divideParam.trainRatio = 70/100;
net3.divideParam.valRatio = 15/100;
net3.divideParam.testRatio = 15/100;
net4.divideParam.trainRatio = 70/100;
net4.divideParam.valRatio = 15/100;
net4.divideParam.testRatio = 15/100;
net5.divideParam.trainRatio = 70/100;
net5.divideParam.valRatio = 15/100;
net5.divideParam.testRatio = 15/100;
%% Train the Network
% Levenberg-Marquardt
net1 = feedforwardnet(11, 'trainlm');
% One Step Secant
net2 = feedforwardnet(10, 'trainoss');
% Gradient Descent
net3 = feedforwardnet(11, 'traingd');
% Scaled Conjugate Gradient
net4 = feedforwardnet(12, 'trainscg');
% Resilient Backpropagation
net5 = feedforwardnet(19, 'trainrp');
%
[net1,tr1] = train(net1,inputs,targets);
[net2,tr2] = train(net2,inputs,targets);
[net3,tr3] = train(net3,inputs,targets);
[net4,tr4] = train(net4,inputs,targets);
[net5,tr5] = train(net5,inputs,targets);
% Test the Network
outputs1 = net1(inputs);
outputs2 = net2(inputs);
outputs3 = net3(inputs);
outputs4 = net4(inputs);
outputs5 = net5(inputs);
errors1 = gsubtract(targets,outputs1);
errors2 = gsubtract(targets,outputs2);
errors3 = gsubtract(targets,outputs3);
errors4 = gsubtract(targets,outputs4);
errors5 = gsubtract(targets,outputs5);
performance1 = perform(net1,targets,outputs1)
```

```matlab
performance2 = perform(net2,targets,outputs2)
performance3 = perform(net3,targets,outputs3)
performance4 = perform(net4,targets,outputs4)
performance5 = perform(net5,targets,outputs5)
% View the Network
view(net1)
view(net2)
view(net3)
view(net4)
view(net5)
% Plot Training States
%Levenberg-Marquardt
figure, plottrainstate(tr1)
% One Step Secant
figure, plottrainstate(tr2)
% Gradient Descent
figure, plottrainstate(tr3)
% Scaled Conjugate Gradient
figure, plottrainstate(tr4)
% Resilient Backpropagation
figure, plottrainstate(tr5)
%% Plot Confusion Matrixes
figure, plotconfusion(targets,outputs1); title('Levenberg-Marquardt');
figure, plotconfusion(targets,outputs2);title('One Step Secant');
figure, plotconfusion(targets,outputs3);title('Gradient Descent');
figure, plotconfusion(targets,outputs4);title('Scaled Conjugate Gradient');
figure, plotconfusion(targets,outputs5);title('Resilient Backpropagation');
% figure, plotperform(tr1)
% figure, ploterrhist(errors1)
```

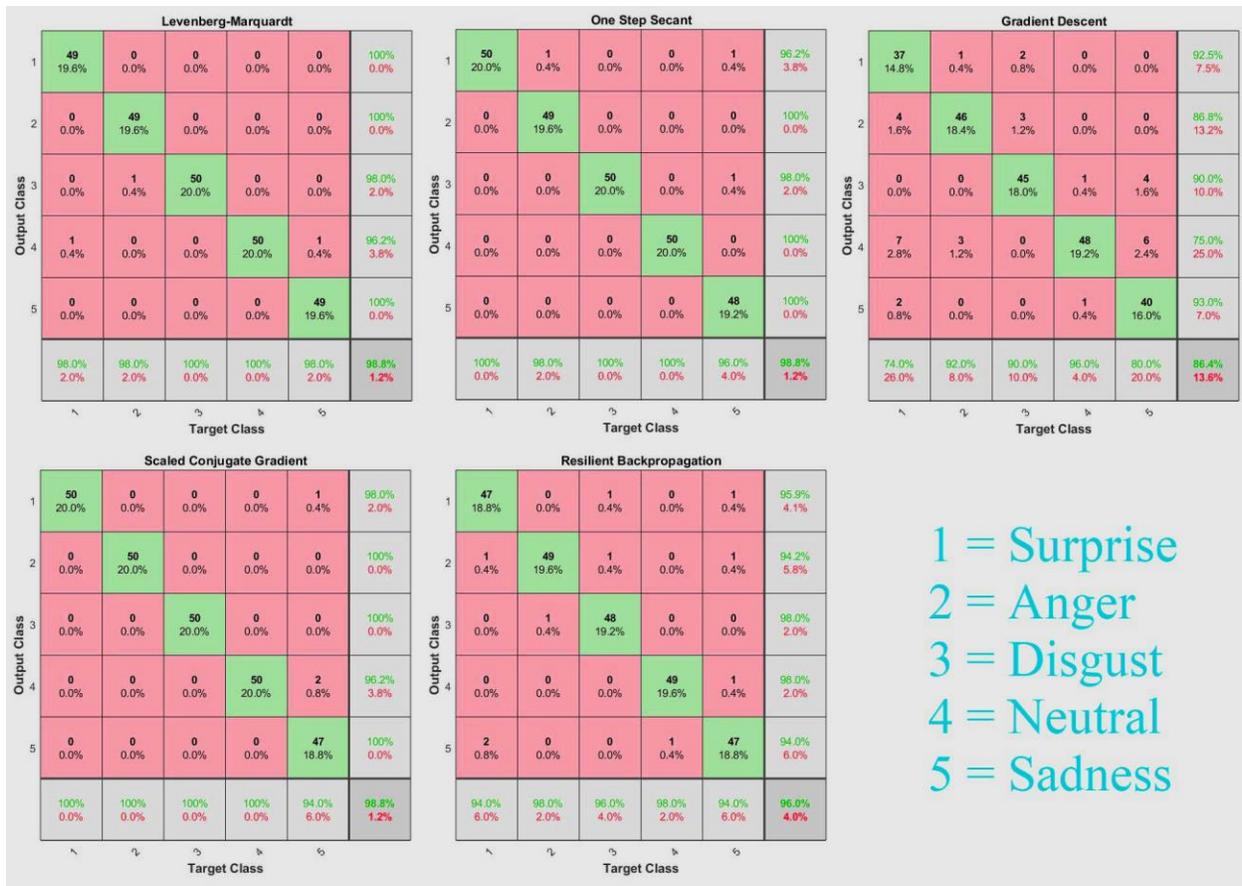

Figure 7.6 Confusion matrixes for five different shallow NN algorithms over five facial expressions (IKFDB depth samples)

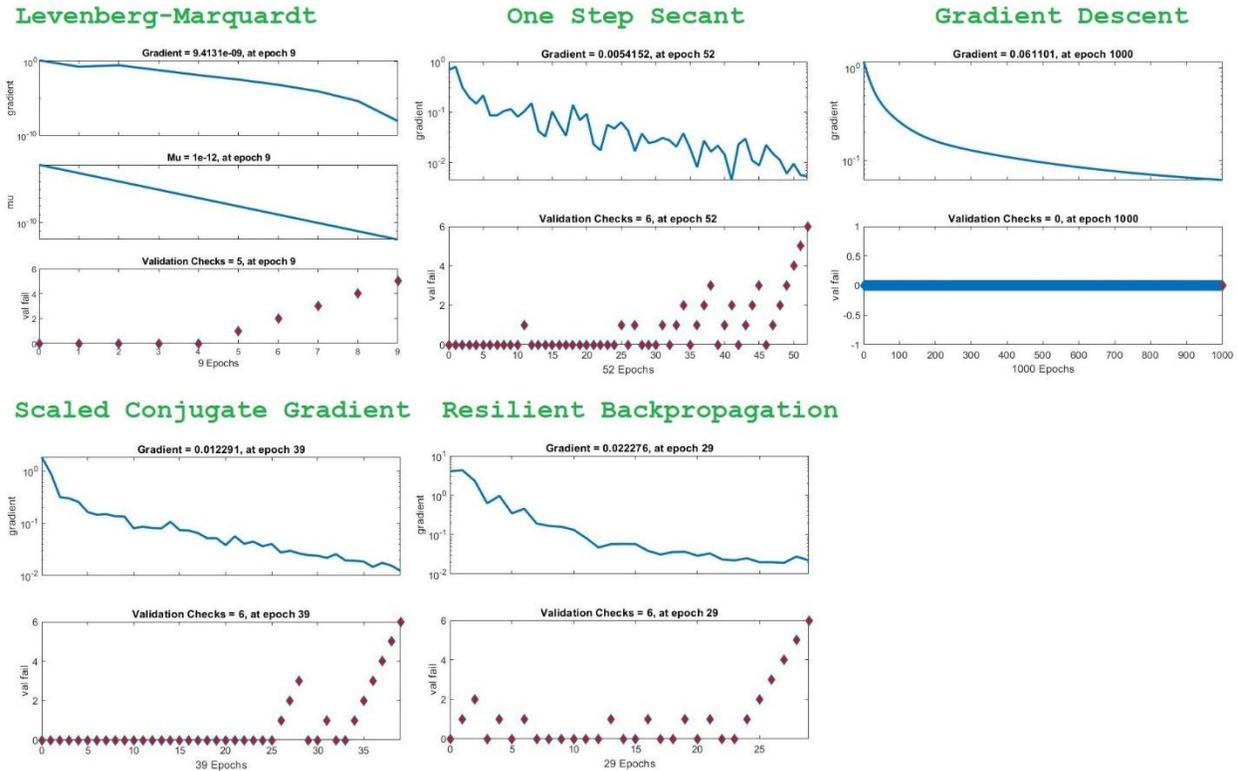

Figure 7.7 Training states plots for five different shallow NN algorithms over five facial expressions (IKFDB depth samples)

## 7.2 Deep Learning Classification

Deep learning [83, 32] uses artificial neural networks to perform sophisticated computations on large amounts of data. It is a type of machine learning that works based on the structure and function of the human brain. Deep learning algorithms train machines by learning from examples. Industries such as health care, eCommerce, entertainment, and advertising commonly use deep learning.

While deep learning algorithms feature self-learning representations, they depend upon ANNs [19, 32, 87] that mirror the way the brain computes information. During the training process, algorithms use unknown elements in the input distribution to extract features, group objects, and discover useful data patterns. Much like training machines for self-learning, this occurs at multiple levels, using the algorithms to build the models. Deep learning models make use of several algorithms. While no one network is considered perfect, some algorithms are better suited to perform specific tasks. To choose the right ones, it's good to gain a solid understanding of all primary algorithms.

Deep learning algorithms work with almost any kind of data and require large amounts of computing power and information to solve complicated issues. Types of Deep Learning Algorithms are [83, 84]:

1. Convolutional Neural Networks (CNNs) [85, 32]
2. Long Short-Term Memory Networks (LSTMs)
3. Recurrent Neural Networks (RNNs)

4. Generative Adversarial Networks (GANs)
5. Radial Basis Function Networks (RBFNs)
6. Multilayer Perceptrons (MLPs)
7. Self-Organizing Maps (SOMs)
8. Deep Belief Networks (DBNs)
9. Restricted Boltzmann Machines (RBMs)
10. Autoencoders

Convolutional Neural Networks (CNNs) [85, 32] also known as ConvNets, consist of multiple layers and are mainly used for image processing and object detection. Yann LeCun developed the first CNN in 1988 when it was called LeNet. It was used for recognizing characters like ZIP codes and digits. CNNs are widely used to identify satellite images, process medical images, forecast time series, and detect anomalies. CNN's have multiple layers that process and extract features from data:

1. Convolution Layer:

CNN has a convolution layer that has several filters to perform the convolution operation.

2.Rectified Linear Unit (ReLU):

CNN's have a ReLU layer to perform operations on elements. The output is a rectified feature map.

3.Pooling Layer:

The rectified feature map next feeds into a pooling layer. Pooling is a down-sampling operation that reduces the dimensions of the feature map. The pooling layer then converts the resulting two-dimensional arrays from the pooled feature map into a single, long, continuous, linear vector by flattening it.

4.Fully Connected Layer:

A fully connected layer forms when the flattened matrix from the pooling layer is fed as an input, which classifies and identifies the images. Figure 7.8 shows the processing structure in the CNN.

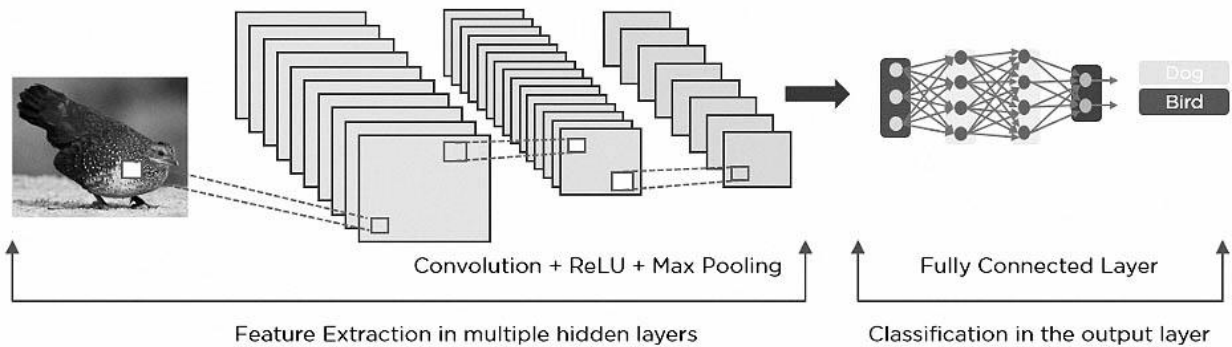

Figure 7.8 Processing structure in the CNN

Following lines of code, classifies 5 classes of facial expressions using CNN algorithm (train and test). Data is depth images from IKFDB [32] database exactly as previews experiment. Each section of code is explained in the code as comment. Figure 7.9 presents the performance of the system over 25 epochs. Figure 7.10 illustrates 250 depth samples which is used in the chapter 7.

```
% CNN Deep Neural Network Algorithm (Train and Test)
% Code name : "c.7.3.m"
% CNN Classification
clear;
% Load the deep sample data as an image datastore. imageDatastore
automatically
% labels the images based on folder names and stores the data as an
ImageDatastore object.
```

```matlab
%  The class labels are sourced from the subfolder names.
deepDatasetPath = fullfile('Deep');
imds = imageDatastore(deepDatasetPath, ...
    'IncludeSubfolders',true, ...
    'LabelSource','foldernames');
% Divide the data into training and validation data sets, so that each
category in
% the training set contains 40 images, and the validation set contains the
remaining
% images from each label. splitEachLabel splits the datastore Data into two
new datastores,
% trainDigitData and valDigitData.
numTrainFiles = 40;
[imdsTrain,imdsValidation] = splitEachLabel(imds,numTrainFiles,'randomize');
% Define the convolutional neural network architecture.
layers = [
% Image Input Layer An imageInputLayer is where you specify the image size,
% which, in this case, is 200-by-128-by-1. These numbers correspond to the
height,
% width, and the channel size. The digit data consists of grayscale images,
so
% the channel size (color channel) is 1. For a color image, the channel size
is 3
    imageInputLayer([200 128 1])
% Convolutional Layer In the convolutional layer, the first argument is
filterSize,
% which is the height and width of the filters the training function uses
while
% scanning along the images. In this example, the number 3 indicates that the
filter
% size is 3-by-3. You can specify different sizes for the height and width of
the
% filter. The second argument is the number of filters, numFilters, which is
the
% number of neurons that connect to the same region of the input. This
parameter
% determines the number of feature maps. Use the 'Padding' name-value pair to
add
% padding to the input feature map. For a convolutional layer with a default
stride
% of 1, 'same' padding ensures that the spatial output size is the same as
the input size.
    convolution2dLayer(3,8,'Padding','same')
% Batch Normalization Layer Batch normalization layers normalize the
activations
% and gradients propagating through a network, making network training an
easier
% optimization problem. Use batch normalization layers between convolutional
layers
% and nonlinearities, such as ReLU layers, to speed up network training and
reduce
% the sensitivity to network initialization.
    batchNormalizationLayer
% ReLU Layer The batch normalization layer is followed by a nonlinear
activation
% function. The most common activation function is the rectified linear unit
(ReLU).
```

```
    reluLayer
% Max Pooling Layer Convolutional layers (with activation functions) are
sometimes
% followed by a down-sampling operation that reduces the spatial size of the
feature
% map and removes redundant spatial information. Down-sampling makes it
possible to
% increase the number of filters in deeper convolutional layers without
increasing
% the required amount of computation per layer. One way of down-sampling is
using a
% max pooling, which you create using maxPooling2dLayer. The max pooling
layer
% returns the maximum values of rectangular regions of inputs, specified by
the
% first argument, poolSize. In this example, the size of the rectangular
region is [2,2].
    maxPooling2dLayer(2,'Stride',2)
    convolution2dLayer(3,16,'Padding','same')
    batchNormalizationLayer
    reluLayer
    maxPooling2dLayer(2,'Stride',2)
    convolution2dLayer(3,32,'Padding','same')
    batchNormalizationLayer
    reluLayer
% Fully Connected Layer The convolutional and down-sampling layers are
followed by
% one or more fully connected layers. As its name suggests, a fully connected
layer
% is a layer in which the neurons connect to all the neurons in the preceding
layer.
% This layer combines all the features learned by the previous layers across
the
% image to identify the larger patterns. The last fully connected layer
combines
% the features to classify the images. Therefore, the OutputSize parameter in
the
% last fully connected layer is equal to the number of classes in the target
data.
% In this example, the output size is 5, corresponding to the 5 classes.
    fullyConnectedLayer(5)
% Softmax Layer The softmax activation function normalizes the output of the
fully
% connected layer. The output of the softmax layer consists of positive
numbers that
% sum to one, which can then be used as classification probabilities by the
% classification layer.
    softmaxLayer
% Classification Layer The final layer is the classification layer. This
layer uses
% the probabilities returned by the softmax activation function for each
input to
% assign the input to one of the mutually exclusive classes and compute the
loss.
    classificationLayer];
% After defining the network structure, specify the training options. Train
the
```

```matlab
% network using stochastic gradient descent with momentum (SGDM) with an initial
% learning rate of 0.01. Set the maximum number of epochs to 25. An epoch is a full
% training cycle on the entire training data set. Monitor the network accuracy
% during training by specifying validation data and validation frequency. Shuffle
% the data every epoch. The software trains the network on the training data and
% calculates the accuracy on the validation data at regular intervals during
% training. The validation data is not used to update the network weights. Turn on
% the training progress plot, and turn off the command window output.
options = trainingOptions('sgdm', ...
    'InitialLearnRate',0.01, ...
    'MaxEpochs',25, ...
    'Shuffle','every-epoch', ...
    'ValidationData',imdsValidation, ...
    'ValidationFrequency',30, ...
    'Verbose',false, ...
    'Plots','training-progress');
% Train the network using the architecture defined by layers, the training data,
% and the training options.The training progress plot shows the mini-batch loss and
% accuracy and the validation loss and accuracy.
net = trainNetwork(imdsTrain,layers,options);
% Predict the labels of the validation data using the trained network, and
% calculate the final validation accuracy. Accuracy is the fraction of labels
% that the network predicts correctly.
YPred = classify(net,imdsValidation);
YValidation = imdsValidation.Labels;
accuracy = sum(YPred == YValidation)/numel(YValidation)
```

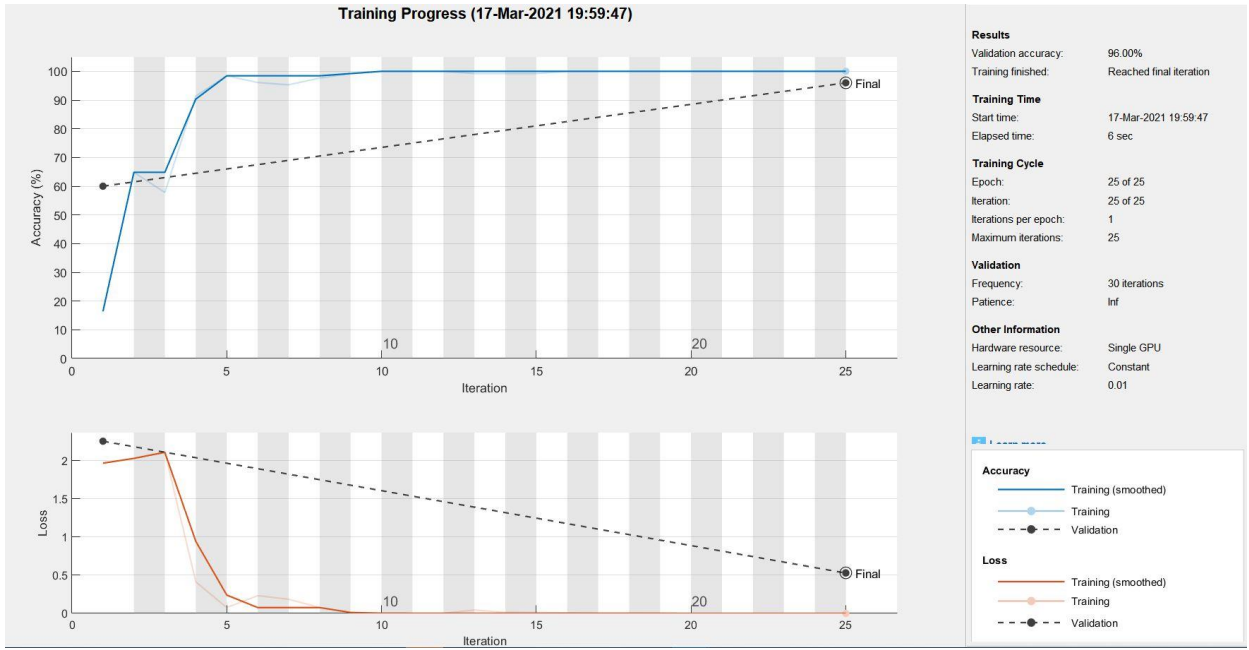

Figure 7.9 The performance of the system over 25 epochs

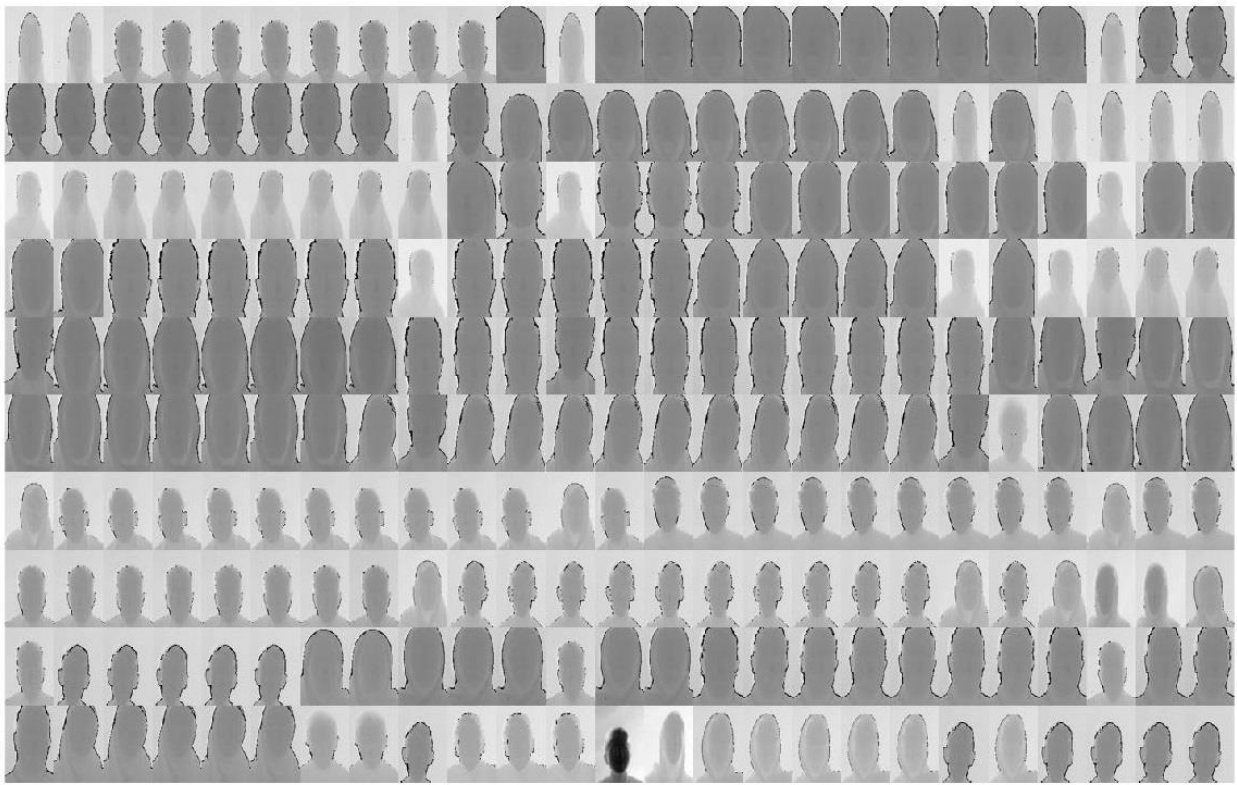

Figure 7.10 250 depth samples which is used in this chapter

**7.3 Exercises**

1. (P3): Try to classify 4 classes of expressions in the "Four KDEF" folder (available in the book folder) with shallow and deep neural network algorithms and plot the results in train and test stages.

# *Chapter   8*

## *A     Full     Experiment (FMER on IKFDB)*

# *Chapter 8*

## *A Full Experiment (FMER on IKFDB)*

## Contents

This chapter uses previews chapters' experiments results, in order to perform a full facial expressions and micro facial expressions recognition project on color and depth data using different methods. Basically, by learning this chapter you should be able to perform any image-based recognition system with any number of classes and any type of image data. For example, by replacing your data, you can run a face recognition project instead of FER project.

**8.1 The project**

System starts with reading (2500 to 5000 samples of IKFDB) and preparing data as follow : 1. Reading, 2. Detecting and extracting faces (Viola and Jones algorithm), 3. Removing any non-faces and cropping the remaining for processing, 4. Gray level conversion, 5. Resizing dimensions, 6. Contrast adjustment, 7. Edge detection and low-high pass filtering (Sharp polished algorithm), 8. LBP features for color data from edge detected images alongside with lasso feature selection (data destination : book folder\IKFDB ClassLearner) , 9. HOG features from color data alongside with lasso feature selection(data destination : book folder\IKFDB ClassLearner) , 10. Extracting SURF features from color data (data destination : book folder\IKFDB SURF), 11. Extracting LPQ features from depth data (data destination : book folder\IKFDB ClassLearner), 12. Extracting Gabor features from depth data and feature selection with lasso (data destination : book folder\IKFDB ClassLearner), 13. Combining all extracted feature matrixes into one feature matrix of test, 14. Labeling data for seven main expressions of Anger, Disgust, Fear, Joy, Neutral, Sadness and Surprise. Feature matrix is stored under the title of "matlab.mat" in the book folder and the last column is representing 7 classes with numbers of 1 to 7 which 1 is Anger and 7 is Surprise. These labels are for classification learner app (Fine-KNN, Fine Tree, Subspace K-NN, Cubic SVM, Linear Discriminant and Kernel Naïve Bayes algorithms are used), 15. Also, system configures and labels data for shallow neural network of "Polak-Ribiére Conjugate Gradient" and returns the result in train and test stages, 16. Deep neural network of CNN and returning results in train and test stages during 30 epochs (data destination: book folder\IKFDB Chap8deep).

Following 300 lines of code runs the previews paragraph comments as a full FMER project. You can evaluate the code in the Matlab software and it took around 30 minutes for full processing with a system with a Core i7 CPU. Please be sure to have the book folder downloaded and determine Matlab path into the folder. Figure 8.1 shows some of the samples after pre-processing and edge detection. Figure 8.2 presents ROC and confusion matrixes for 6 classification methods of Fine-KNN, Fine Tree, Subspace K-NN, Cubic SVM, Linear Discriminant and Kernel Naïve Bayes alongside with recognition accuracy in the training stage. Also, Figure 8.3 illustrates training states and confusion matrix plots for "Polak-Ribiére Conjugate Gradient" shallow neural network algorithm along-side with recognition accuracy for test stage. Finally, Figure 8.4 presents the performance of CNN deep learning algorithm on data in 30 epochs (630 iterations) along-side with final test recognition accuracy and Loss value.

```matlab
% Final Project (Train and Test)
% Code name : "c.8.1.m"
clc;
clear;
%% Reading and Face Detection
FDetect =
vision.CascadeObjectDetector('FrontalFaceCART','MergeThreshold',16');
%Read the input images
path='IKFDB ClassLearner';
fileinfo = dir(fullfile(path,'*.jpg'));
filesnumber=size(fileinfo);
for i = 1 : filesnumber(1,1)
images{i} = imread(fullfile(path,fileinfo(i).name));
    disp(['Loading image No :  ' num2str(i) ]);
end;
% Returns Bounding Box values based on number of objects
for i = 1 : filesnumber(1,1)
BB{i}=step(FDetect,(imread(fullfile(path,fileinfo(i).name))));
    disp(['BB :  ' num2str(i) ]);
end;
% Find number of empty BB and index of them
c=0;
for  i = 1 : filesnumber(1,1)
    if  isempty(BB{i})
        c=c+1;
        indexempty(c)=i;
    end;
end;
% Replace the empty cells with bounding box
for  i = 1 : c
BB{indexempty(i)}=[40 60 180 180];
end;
% Removing other founded faces and keep just frist face or box
for  i = 1 : filesnumber(1,1)
    BB{i}=BB{i}(1,:);
end;
% Croping the Bounding box(face)
for i = 1 : filesnumber(1,1)
croped{i}=imcrop(images{i},BB{i});
    disp(['Croped :   ' num2str(i) ]);
end;
%% Color to Gray Conversion
for i = 1 : filesnumber(1,1)
gray{i}=rgb2gray(croped{i});
```

```matlab
    disp(['To Gray :   ' num2str(i) ]);
end;
%% Resize Images
for i = 1 : filesnumber(1,1)
resized{i}=imresize(gray{i}, [256 256]);
    disp(['Image Resized :   ' num2str(i) ]);
end;
%% Contrast Adjustment
for i = 1 : filesnumber(1,1)
adjusted{i}=imadjust(resized{i});
    disp(['Image Adjust :   ' num2str(i) ]);
end;
%% Sharp Polished Edge Detection
for i = 1 : filesnumber(1,1)
[edge{i},polished{i}]=sharppolished(adjusted{i});
    disp(['Sharp Polished Edge Detection and Pre-Processing :   '
num2str(i)]);
end;
%% Extract LBP Features (just color)
for i = 1 : filesnumber(1,1)
    % The less cell size the more accuracy
lbp{i} = extractLBPFeatures(polished{i},'CellSize',[32 32]);
    disp(['Extract LBP :   ' num2str(i) ]);
end;
for i = 1 : filesnumber(1,1)
    lbpfeature(i,:)=lbp{i};
    disp(['LBP To Matrix :   ' num2str(i) ]);
end;
    disp(['Working On Lasso For LBP (Please Wait) ...']);
lbpmain=lbpfeature;
% Labeling for lasso
clear lasso;clear B;clear Stats;clear ds;
label(1:160,1)=1;
label(161:320,1)=2;label(321:480,1)=3;label(481:640,1)=4;
label(641:800,1)=5;label(801:960,1)=6;label(961:1120,1)=7;
% clear lasso;
[B Stats] = lasso(lbpfeature,label, 'CV', 5);
disp(B(:,1:5))
disp(Stats)
%
lassoPlot(B, Stats, 'PlotType', 'CV')
ds.Lasso = B(:,Stats.IndexMinMSE);
disp(ds)
sizemfcc=size(lbpfeature);
temp=1;
for i=1:sizemfcc(1,2)
if ds.Lasso(i)~=0
lasso(:,temp)=lbpfeature(:,i);
temp=temp+1;end;end;lbpfeature=lasso;
%% Extract HOG Features (just color)
for i = 1 : filesnumber(1,1)
    % The less cell size the more accuracy
hog{i} = extractHOGFeatures(adjusted{i},'CellSize',[16 16]);
    disp(['Extract HOG :   ' num2str(i) ]);
end;
for i = 1 : filesnumber(1,1)
    hogfeature(i,:)=hog{i};
```

```matlab
        disp(['HOG To Matrix :   ' num2str(i) ]);
end;
        disp(['Working On Lasso For HOG (Please Wait) ...']);
hogmain=hogfeature;
% Labeling for lasso
clear lasso;clear B;clear Stats;clear ds;
label(1:160,1)=1;
label(161:320,1)=2;label(321:480,1)=3;label(481:640,1)=4;
label(641:800,1)=5;label(801:960,1)=6;label(961:1120,1)=7;
% clear lasso;
[B Stats] = lasso(hogfeature,label, 'CV', 5);
disp(B(:,1:5))
disp(Stats)
%
lassoPlot(B, Stats, 'PlotType', 'CV')
ds.Lasso = B(:,Stats.IndexMinMSE);
disp(ds)
sizemfcc=size(hogfeature);
temp=1;
for i=1:sizemfcc(1,2)
if ds.Lasso(i)~=0
lasso(:,temp)=hogfeature(:,i);
temp=temp+1;end;end;hogfeature=lasso;
%% Extract SURF Features (just color)
imset = imageSet('IKFDB SURF','recursive');
% Create a bag-of-features from the image database
bag = bagOfFeatures(imset,'VocabularySize',200,'PointSelection','Detector');
% Encode the images as new features
surf = encode(bag,imset);
%% Extract LPQ Features (just depth)
% Read the input images
path='IKFDB ClassLearner';fileinfo = dir(fullfile(path,'*.png'));
filesnumber=size(fileinfo);
for i = 1 : filesnumber(1,1)
images{i} = imread(fullfile(path,fileinfo(i).name));
        disp(['Loading image No :   ' num2str(i) ]);end;
% Contrast
for i = 1 : filesnumber(1,1)
adjusted2{i}=imadjust(images{i}); disp(['Contrast Adjust :   ' num2str(i)
]);end;
% Resize
for i = 1 : filesnumber(1,1)
resized2{i}=imresize(adjusted2{i}, [256 256]); disp(['Image Resized :   '
num2str(i) ]);end;
% LPQ
winsize=39;
for i = 1 : filesnumber(1,1)
tmp{i}=lpq(resized2{i},winsize);disp(['Extract LPQ :   ' num2str(i) ]);end;
for i = 1 : filesnumber(1,1)lpq(i,:)=tmp{i};end;

%% Extract Gabor Features (just depth)
for i = 1 : filesnumber(1,1)
gaborArray = gaborFilterBank(3,8,29,29);  % Generates the Gabor filter bank
featureVector{i} = gaborFeatures(resized2{i},gaborArray,16,16);
disp(['Extracting Gabor Vector :   ' num2str(i) ]);
end;
for i = 1 : filesnumber(1,1)
```

```matlab
    Gaborvector(i,:)=featureVector{i};
    disp(['To Matrix :   ' num2str(i) ]);end;
    disp(['Working On Lasso For Gabor (Please Wait) ...']);
%
clear lasso;clear B;clear Stats;clear ds;
[B Stats] = lasso(Gaborvector,label, 'CV', 5);
disp(B(:,1:5))
disp(Stats)
%
lassoPlot(B, Stats, 'PlotType', 'CV')
ds.Lasso = B(:,Stats.IndexMinMSE);
disp(ds)
sizemfcc=size(Gaborvector);temp=1;
%
for i=1:sizemfcc(1,2)
if ds.Lasso(i)~=0
lasso(:,temp)=Gaborvector(:,i);temp=temp+1;end;end;
%
Gabor=lasso;
%% Labeling for Classification
clear test;
% Combining Feature Matrixes
test=[lbpfeature hogfeature surf lpq Gabor];
% Labels
sizefinal=size(test);
sizefinal=sizefinal(1,2);
test(1:160,sizefinal+1)=1;
test(161:320,sizefinal+1)=2;
test(321:480,sizefinal+1)=3;
test(481:640,sizefinal+1)=4;
test(641:800,sizefinal+1)=5;
test(801:960,sizefinal+1)=6;
test(961:1120,sizefinal+1)=7;
% Reading ready to classify feature matrix
clear ikfdb;clear lbl;clear sizeikfdb;
ikfdb=load('matlab.mat');
ikfdb=ikfdb.test;
classificationLearner

%% Shallow Neural Network
% clear;
netdata=load('matlab.mat');
netdata=netdata.test;
% Labeling For Classification
network=netdata(:,1:end-1);
netlbl=netdata(:,end);
sizenet=size(network);
sizenet=sizenet(1,1);
for i=1 : sizenet
            if netlbl(i) == 1
                netlbl2(i,1)=1;
                netlbl2(i,2)=1;
        elseif netlbl(i) == 3
                netlbl2(i,3)=1;
        elseif netlbl(i) == 4
                netlbl2(i,4)=1;
        elseif netlbl(i) == 5
```

```matlab
                netlbl2(i,5)=1;
        elseif netlbl(i) == 6
                netlbl2(i,6)=1;
        elseif netlbl(i) == 7
                netlbl2(i,7)=1;
        end
end
% Changing data shape from rows to columns
network=network';
% Changing data shape from rows to columns
netlbl2=netlbl2';
% Defining input and target variables
inputs = network;
targets = netlbl2;
% Create a Pattern Recognition Network
hiddenLayerSize = 100;
net = patternnet(hiddenLayerSize);
% Set up Division of Data for Training, Validation, Testing
net.divideParam.trainRatio = 70/100;
net.divideParam.valRatio = 15/100;
net.divideParam.testRatio = 15/100;
% Train the Network
% Polak-Ribiére Conjugate Gradient
net = feedforwardnet(20, 'traincgp');
%
[net,tr] = train(net,inputs,targets);
% Test the Network
outputs = net(inputs);
%
errors = gsubtract(targets,outputs);
%
performance = perform(net,targets,outputs)
% Polak-Ribiére Conjugate Gradient
figure, plottrainstate(tr)
% Plot Confusion Matrixes
figure, plotconfusion(targets,outputs);
title('Polak-Ribiére Conjugate Gradient');

%% CNN Deep Neural Network Algorithm (Train and Test)
%  clear;
% Load the deep sample data as an image datastore.
deepDatasetPath = fullfile('Chap8deep');
imds = imageDatastore(deepDatasetPath, ...
    'IncludeSubfolders',true, ...
    'LabelSource','foldernames');
% Divide the data into training and validation data sets
numTrainFiles = 400;
[imdsTrain,imdsValidation] = splitEachLabel(imds,numTrainFiles,'randomize');
% Define the convolutional neural network architecture.
layers = [
% Image Input Layer An imageInputLayer
    imageInputLayer([128 128 1])
% Convolutional Layer
convolution2dLayer(3,8,'Padding','same')
% Batch Normalization
    batchNormalizationLayer
% ReLU Layer The batch
```

```matlab
    reluLayer
% Max Pooling Layer
    maxPooling2dLayer(3,'Stride',3)
    convolution2dLayer(3,8,'Padding','same')
    batchNormalizationLayer
    reluLayer
    maxPooling2dLayer(3,'Stride',3)
    convolution2dLayer(3,8,'Padding','same')
    batchNormalizationLayer
    reluLayer
% Fully Connected Layer
    fullyConnectedLayer(7)
% Softmax Layer
    softmaxLayer
% Classification Layer The final layer
    classificationLayer];
% Specify the training options
options = trainingOptions('sgdm', ...
    'InitialLearnRate',0.01, ...
    'MaxEpochs',30, ...
    'Shuffle','every-epoch', ...
    'ValidationData',imdsValidation, ...
    'ValidationFrequency',15, ...
    'Verbose',false, ...
    'Plots','training-progress');
% Train the network
net = trainNetwork(imdsTrain,layers,options);
% Predict the labels
YPred = classify(net,imdsValidation);
YValidation = imdsValidation.Labels;
accuracy = sum(YPred == YValidation)/numel(YValidation)
```

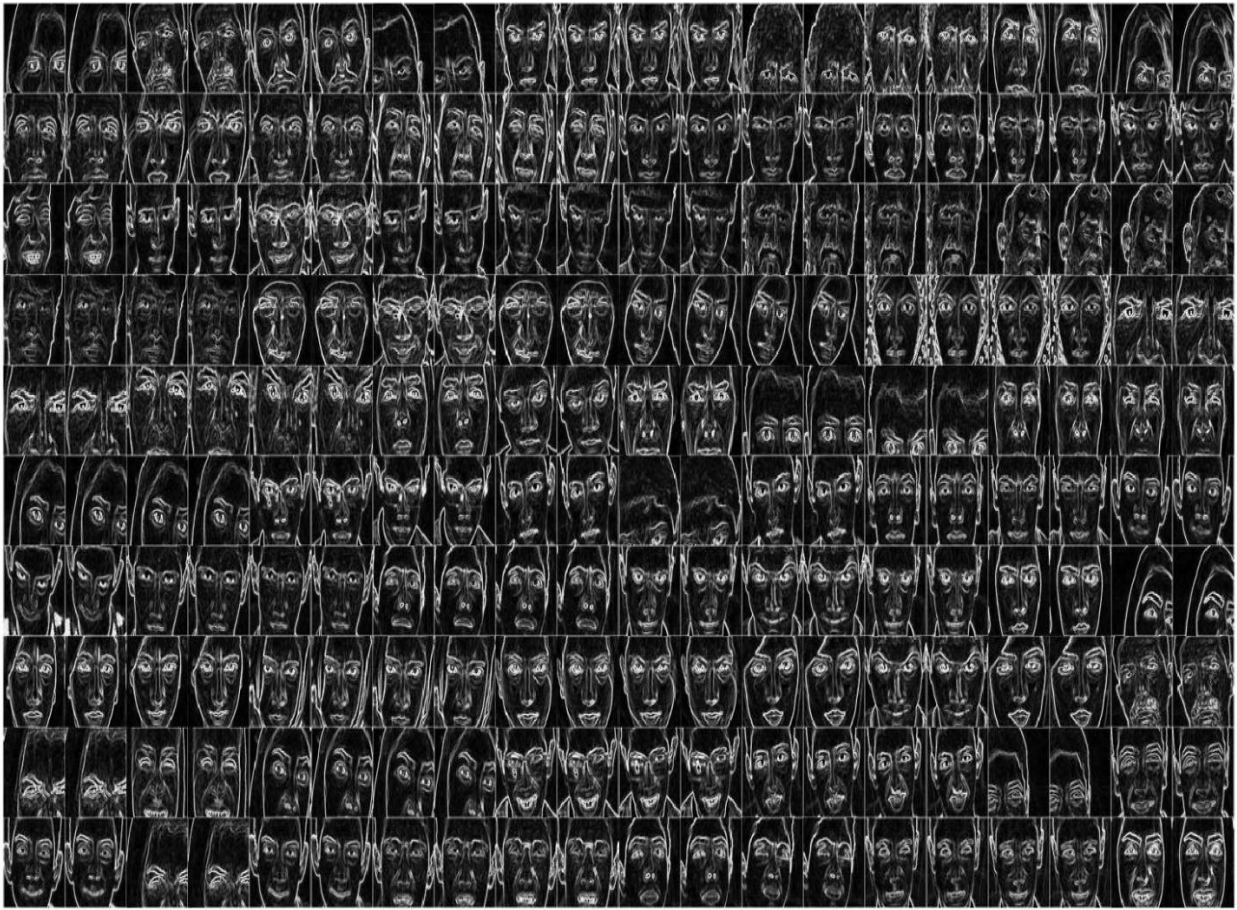

Figure 8.1 Some IKFDB samples after pre-processing and edge detection

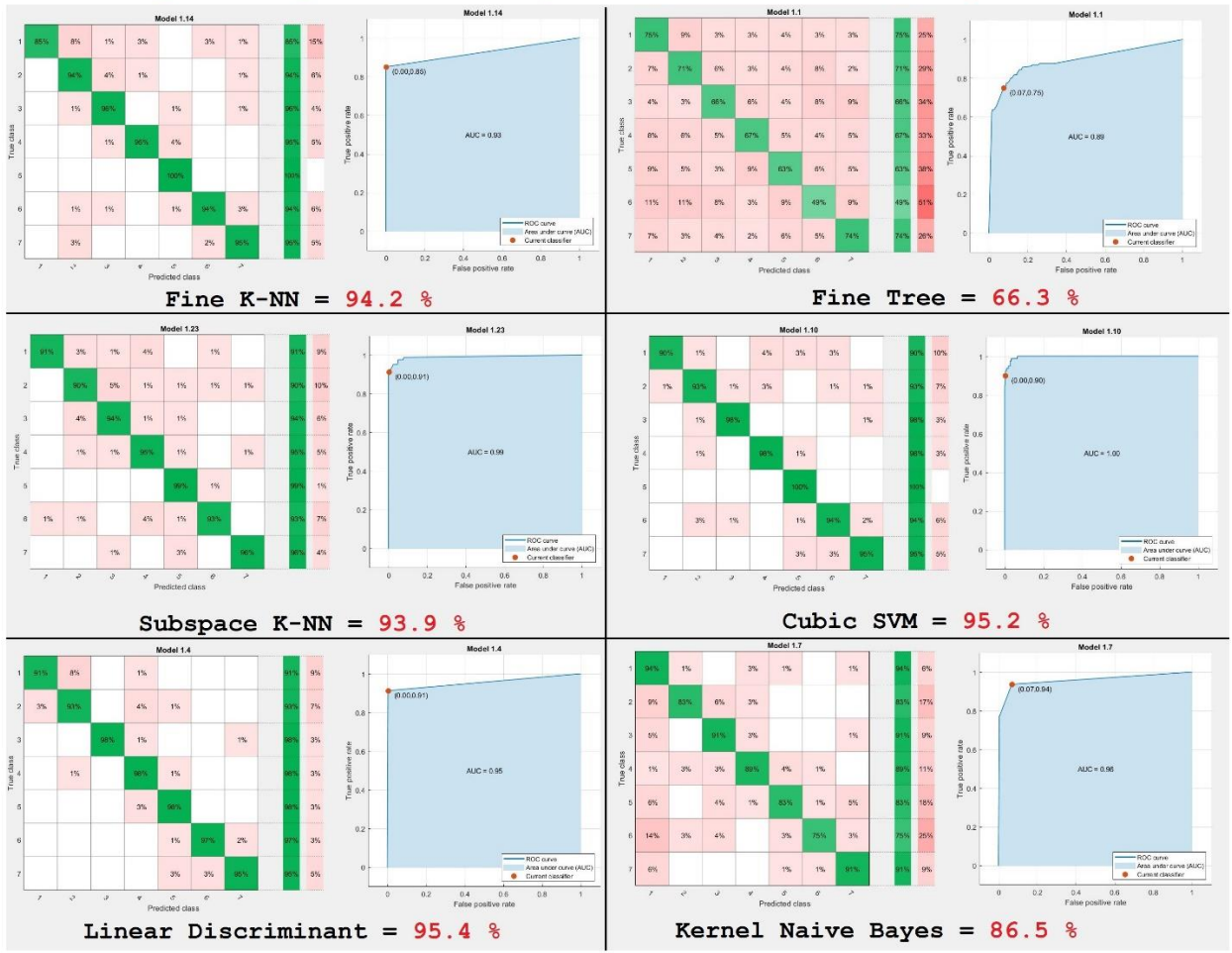

Figure 8.2 ROC and confusion matrix for six classification algorithms on data (7 classes)

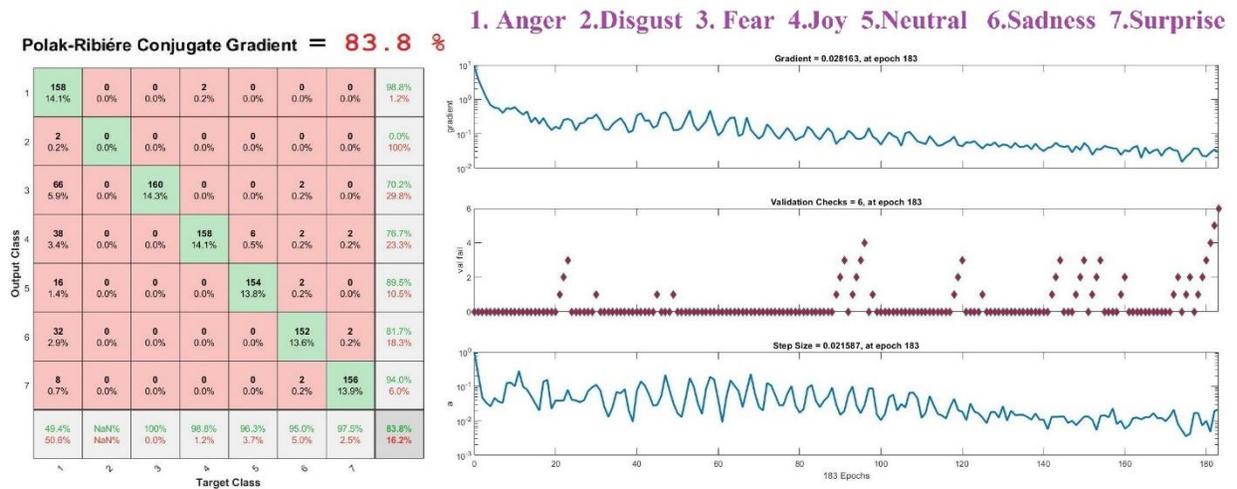

Figure 8.3 Training states plot and confusion matrix plot for shallow neural network classification on data (7 classes)

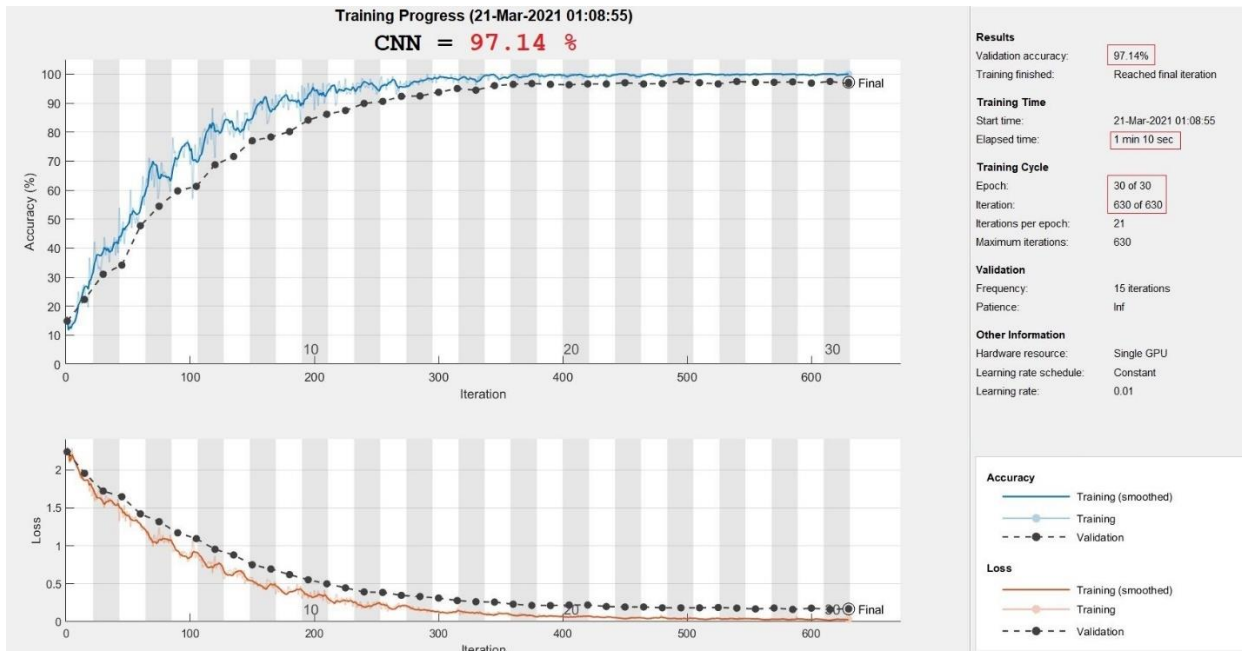

Figure 8.4 Training states in 30 epochs (630 iterations) for CNN deep neural network classification on data (7 classes)

## 8.2 Exercises

1. (P4): Try to classify 8 classes of faces as a face recognition project from IKFDB samples exist in the book folder\Chap8deep. Please use Coarse Tree, Gaussian Naïve Bayes, Quadratic SVM, Cosine K-NN, Ensemble Subspace Discriminant, Levenberg-Marquardt, Bayesian Regularization, BFGS Quasi-Newton and CNN algorithms and return related confusion matrixes. Be sure to use most important pre-processing steps such as face extraction, contrast adjustment, resizing and more. (in selecting 8 subjects out of 40, it is important to use both color and depth images for the whole process).

Help:

https://uk.mathworks.com/help/stats/train-classification-models-in-classification-learner-app.htmlv
https://uk.mathworks.com/help/deeplearning/ref/feedforwardnet.html;jsessionid=6126143adf2eb8fb99e28c7be800
https://uk.mathworks.com/help/deeplearning/gs/create-simple-deep-learning-classification-network.html
https://uk.mathworks.com/help/deeplearning/ug/create-simple-deep-learning-network-for-classification.html
https://uk.mathworks.com/help/vision/ug/image-category-classification-using-deep-learning.html

# APPENDIX  *A*

## The Matlab Image Processing and Analysis Library

### Import, Export, and Conversion

#### Read and Write Image Data from Files

##### Work with DICOM Files

| | |
|---|---|
| dicominfo | Read metadata from DICOM message |
| dicomread | Read DICOM image |
| dicomwrite | Write images as DICOM files |
| dicomreadVolume | Create 4-D volume from set of DICOM images |
| dicomCollection | Gather details about related series of DICOM files |
| dicomContours | Extract ROI data from DICOM-RT structure set |
| dicomanon | Anonymize DICOM file |
| dicomdict | Get or set active DICOM data dictionary |
| dicomdisp | Display DICOM file structure |
| dicomlookup | Find attribute in DICOM data dictionary |
| dicomuid | Generate DICOM globally unique identifier |
| images.dicom.decodeUID | Get information about DICOM unique identifier |
| images.dicom.parseDICOMDIR | Extract metadata from DICOMDIR file |

##### Work with RAW Files

| | |
|---|---|
| rawinfo | Read metadata from RAW file |
| rawread | Read Color Filter Array (CFA) image from RAW file |
| raw2rgb | Transform Color Filter Array (CFA) image in RAW file into RGB image |
| planar2raw | Combine planar sensor images into full Bayer pattern CFA |
| raw2planar | Separate Bayer pattern Color Filter Array (CFA) image into sensor element images |

##### Work with Specialized File Formats

| | |
|---|---|
| tiffreadVolume | Read volume from TIFF file |
| niftiinfo | Read metadata from NIfTI file |
| niftiwrite | Write volume to file using NIfTI format |
| niftiread | Read NIfTI image |
| analyze75info | Read metadata from header file of Analyze 7.5 data set |
| analyze75read | Read image data from image file of Analyze 7.5 data set |
| interfileinfo | Read metadata from Interfile file |
| interfileread | Read images in Interfile format |
| nitfinfo | Read metadata from National Imagery Transmission Format (NITF) file |
| nitfread | Read image from NITF file |
| isnitf | Check if file is National Imagery Transmission Format (NITF) file |
| dpxinfo | Read metadata from DPX file |
| dpxread | Read DPX image |

#### High Dynamic Range Images

| | |
|---|---|
| hdrread | Read high dynamic range (HDR) image |
| hdrwrite | Write high dynamic range (HDR) image file |
| makehdr | Create high dynamic range image |
| tonemap | Render high dynamic range image for viewing |
| tonemapfarbman | Convert HDR image to LDR using edge-preserving multiscale decompositions |
| localtonemap | Render HDR image for viewing while enhancing local contrast |
| blendexposure | Create well-exposed image from images with different exposures |
| camresponse | Estimate camera response function |

## Blocked Images

| | |
|---|---|
| blockedImage | Image made from discrete blocks |
| blockedImageDatastore | Datastore for use with blocks from `blockedImage` objects |
| blockLocationSet | List of block locations in large images |
| selectBlockLocations | Select blocks from blocked images |
| bigimageshow | Display 2-D `blockedImage` object |

### Other

| | |
|---|---|
| images.blocked.Adapter | Adapter interface for `blockedImage` objects |
| images.blocked.BINBlocks | Default adapter for writing blocked image data to disk |
| images.blocked.GenericImage | Store blocks in a single image file |
| images.blocked.GenericImageBlocks | Store each block as image file in folder |
| images.blocked.H5 | Store blocks in a single HDF5 file |
| images.blocked.H5Blocks | Store each block as HDF5 file in folder |
| images.blocked.InMemory | Store blocks in memory |
| images.blocked.JPEGBlocks | Store each block as JPEG file in folder |
| images.blocked.MATBlocks | Store each block as MAT-file in folder |
| images.blocked.PNGBlocks | Store each block as PNG file in folder |
| images.blocked.TIFF | Store blocks in single TIFF file |

## Image Type Conversion

### Convert Between Image Types

| | |
|---|---|
| gray2ind | Convert grayscale or binary image to indexed image |
| ind2gray | Convert indexed image to grayscale image |
| mat2gray | Convert matrix to grayscale image |
| rgb2lightness | Convert RGB color values to lightness values |
| label2rgb | Convert label matrix into RGB image |
| demosaic | Convert Bayer pattern encoded image to truecolor image |
| imsplit | Split multichannel image into its individual channels |

### Convert to Binary Image Using Thresholding

| | |
|---|---|
| imbinarize | Binarize 2-D grayscale image or 3-D volume by thresholding |
| adaptthresh | Adaptive image threshold using local first-order statistics |
| otsuthresh | Global histogram threshold using Otsu's method |
| graythresh | Global image threshold using Otsu's method |

### Convert to Indexed Image Using Quantization

| | |
|---|---|
| imquantize | Quantize image using specified quantization levels and output values |
| multithresh | Multilevel image thresholds using Otsu's method |
| grayslice | Convert grayscale image to indexed image using multilevel thresholding |

### Convert Between Data Types

| | |
|---|---|
| im2int16 | Convert image to 16-bit signed integers |
| im2single | Convert image to single precision |
| im2uint16 | Convert image to 16-bit unsigned integers |
| im2uint8 | Convert image to 8-bit unsigned integers |

## Image Sequences and Batch Processing

| | |
|---|---|
| implay | Play movies, videos, or image sequences |
| montage | Display multiple image frames as rectangular montage |

## Color

### Color Space Conversion

| | |
|---|---|
| rgb2lab | Convert RGB to CIE 1976 L*a*b* |
| rgb2ntsc | Convert RGB color values to NTSC color space |
| rgb2xyz | Convert RGB to CIE 1931 XYZ |
| rgb2ycbcr | Convert RGB color values to YCbCr color space |
| rgbwide2ycbcr | Convert wide-gamut RGB color values to YCbCr color values |
| rgbwide2xyz | Convert wide-gamut RGB color values to CIE 1931 XYZ color values |
| lab2rgb | Convert CIE 1976 L*a*b* to RGB |
| lab2xyz | Convert CIE 1976 L*a*b* to CIE 1931 XYZ |
| ntsc2rgb | Convert NTSC values to RGB color space |
| xyz2lab | Convert CIE 1931 XYZ to CIE 1976 L*a*b* |
| xyz2rgb | Convert CIE 1931 XYZ to RGB |

| xyz2rgbwide | Convert CIE 1931 XYZ color values to wide-gamut RGB color values |
|---|---|
| ycbcr2rgb | Convert YCbCr color values to RGB color space |
| ycbcr2rgbwide | Convert YCbCr color values to wide-gamut RGB color values |
| colorcloud | Display 3-D color gamut as point cloud in specified color space |

### Color Values to Data Type Conversion

| lab2double | **Convert L*a*b* color values to double** |
|---|---|
| lab2uint16 | Convert L*a*b color values to `uint16` |
| lab2uint8 | Convert L*a*b color values to `uint8` |
| xyz2double | Convert XYZ color values to `double` |
| xyz2uint16 | Convert XYZ color values to `uint16` |

### International Color Consortium (ICC) Profile

| iccfind | **Find ICC profiles** |
|---|---|
| iccread | Read ICC profile |
| iccroot | Find system default ICC profile repository |
| iccwrite | Write ICC color profile data |
| isicc | Check for valid ICC profile data |

### Color Transformation

| makecform | **Create color transformation structure** |
|---|---|
| applycform | Apply device-independent color space transformation |

### Automatic White Balance

| chromadapt | **Adjust color balance of RGB image with chromatic adaptation** |
|---|---|
| illumgray | Estimate illuminant using gray world algorithm |
| illumpca | Estimate illuminant using principal component analysis (PCA) |
| illumwhite | Estimate illuminant using White Patch Retinex algorithm |
| lin2rgb | Apply gamma correction to linear RGB values |
| rgb2lin | Linearize gamma-corrected RGB values |
| whitepoint | XYZ color values of standard illuminants |

### Measure Color Differences

| colorangle | **Angle between two RGB vectors** |
|---|---|
| deltaE | Color difference based on CIE76 standard |
| imcolordiff | Color difference based on CIE94 or CIE2000 standard |

## Synthetic Images

| checkerboard | **Create checkerboard image** |
|---|---|
| phantom | Create head phantom image |
| imnoise | Add noise to image |

# Display and Exploration

## Basic Display

### Display Images and Image Sequences

| imshow | **Display image** |
|---|---|
| imfuse | Composite of two images |
| imshowpair | Compare differences between images |
| montage | Display multiple image frames as rectangular montage |
| immovie | Make movie from multiframe image |
| implay | Play movies, videos, or image sequences |
| warp | Display image as texture-mapped surface |

### Display Volumes

| sliceViewer | **Browse image slices** |
|---|---|
| orthosliceViewer | Browse orthogonal slices in grayscale or RGB volume |
| volshow | Display volume |
| labelvolshow | Display labeled volume |
| obliqueslice | Extract oblique slice from 3-D volumetric data |

### Set Image Display Preferences

| iptgetpref | **Get values of Image Processing Toolbox preferences** |
|---|---|
| iptprefs | Display Image Processing Toolbox Preferences dialog box |
| iptsetpref | Set Image Processing Toolbox preferences or display valid values |

## Interactive Exploration with Image Viewer App

| imtool | **Open Image Viewer app** |
|---|---|

| iptgetpref | Get values of Image Processing Toolbox preferences |
|---|---|
| iptprefs | Display Image Processing Toolbox Preferences dialog box |
| iptsetpref | Set Image Processing Toolbox preferences or display valid values |
| isrset | Check if file is valid R-Set file |
| openrset | Open R-Set file and display R-Set |
| rsetwrite | Create R-Set file from image file |

## Interactive Exploration of Volumetric Data with Volume Viewer App

| volshow | **Display volume** |
|---|---|
| labelvolshow | Display labeled volume |
| iptgetpref | Get values of Image Processing Toolbox preferences |
| iptprefs | Display Image Processing Toolbox Preferences dialog box |
| iptsetpref | Set Image Processing Toolbox preferences or display valid values |

## Build Interactive Tools

### Interactive Tools for Image Display and Exploration

| imageinfo | **Image Information tool** |
|---|---|
| imcolormaptool | Choose Colormap tool |
| imcontrast | Adjust Contrast tool |
| imcrop | Crop image |
| imcrop3 | Crop 3-D image |
| imdisplayrange | Display Range tool |
| imdistline | Distance tool |
| impixelinfo | Pixel Information tool |
| impixelinfoval | Pixel Information tool without text label |
| impixelregion | Pixel Region tool |
| impixelregionpanel | Pixel Region tool panel |
| immagbox | Magnification box for image displayed in scroll panel |
| imoverview | Overview tool for image displayed in scroll panel |
| imoverviewpanel | Overview tool panel for image displayed in scroll panel |
| imsave | Save Image Tool |
| imscrollpanel | Scroll panel for interactive image navigation |

### Get Image Properties

| getimage | **Image data from axes** |
|---|---|
| imagemodel | Image model object |
| getimagemodel | Image model object from image object |
| imattributes | Information about image attributes |
| imhandles | Get all image objects |

### Create Custom Modular Tools

| axes2pix | **Convert axes coordinates to pixel coordinates** |
|---|---|
| imgca | Get current axes containing image |
| imgcf | Get current figure containing image |
| imgetfile | Display Open Image dialog box |
| imputfile | Display Save Image dialog box |
| iptaddcallback | Add function handle to callback list |
| iptcheckmap | Check validity of colormap |
| iptcheckhandle | Check validity of handle |
| iptgetapi | Get Application Programmer Interface (API) for handle |
| iptGetPointerBehavior | Retrieve pointer behavior from graphics object |
| ipticondir | Directories containing Image Processing Toolbox and MATLAB icons |
| iptPointerManager | Create pointer manager in figure |
| iptremovecallback | Delete function handle from callback list |
| iptSetPointerBehavior | Store pointer behavior structure in graphics object |
| iptwindowalign | Align figure windows |
| makeConstrainToRectFcn | Create rectangularly bounded drag constraint function |
| truesize | Adjust display size of image |

# Geometric Transformation and Image Registration

## Common Geometric Transformations

| imcrop | **Crop image** |
|---|---|
| imcrop3 | Crop 3-D image |

| imresize | Resize image |
|---|---|
| imresize3 | Resize 3-D volumetric intensity image |
| imrotate | Rotate image |
| imrotate3 | Rotate 3-D volumetric grayscale image |
| imtranslate | Translate image |
| impyramid | Image pyramid reduction and expansion |

## Generic Geometric Transformations

| imwarp | **Apply geometric transformation to image** |
|---|---|
| affineOutputView | Create output view for warping images |
| fitgeotrans | Fit geometric transformation to control point pairs |
| findbounds | Find output bounds for spatial transformation |
| fliptform | Flip input and output roles of spatial transformation structure |
| makeresampler | Create resampling structure |
| maketform | Create spatial transformation structure (TFORM) |
| tformarray | Apply spatial transformation to N-D array |
| tformfwd | Apply forward spatial transformation |
| tforminv | Apply inverse spatial transformation |

### Image Warper Object

| Warper | **Apply same geometric transformation to many images efficiently** |
|---|---|

### Spatial Referencing Objects

| imref2d | **Reference 2-D image to world coordinates** |
|---|---|
| imref3d | Reference 3-D image to world coordinates |

### Transformation Objects

| affine2d | **2-D affine geometric transformation** |
|---|---|
| affine3d | 3-D affine geometric transformation |
| rigid2d | 2-D rigid geometric transformation |
| rigid3d | 3-D rigid geometric transformation |
| projective2d | 2-D projective geometric transformation |
| geometricTransform2d | 2-D geometric transformation object |
| geometricTransform3d | 3-D geometric transformation object |
| PiecewiseLinearTransformation2D | 2-D piecewise linear geometric transformation |
| PolynomialTransformation2D | 2-D polynomial geometric transformation |
| LocalWeightedMeanTransformation2D | 2-D local weighted mean geometric transformation |

## Image Registration

### Automatic Registration

| imregister | **Intensity-based image registration** |
|---|---|
| imregconfig | Configurations for intensity-based registration |
| imregtform | Estimate geometric transformation that aligns two 2-D or 3-D images |
| imregcorr | Estimate geometric transformation that aligns two 2-D images using phase correlation |
| imregdemons | Estimate displacement field that aligns two 2-D or 3-D images |
| imregmtb | Register 2-D images using median threshold bitmaps |
| normxcorr2 | Normalized 2-D cross-correlation |
| MattesMutualInformation | Mattes mutual information metric configuration |
| MeanSquares | Mean square error metric configuration |
| RegularStepGradientDescent | Regular step gradient descent optimizer configuration |
| OnePlusOneEvolutionary | One-plus-one evolutionary optimizer configuration |

### Control Point Registration

| cpselect | **Control Point Selection tool** |
|---|---|
| fitgeotrans | Fit geometric transformation to control point pairs |
| cpcorr | Tune control point locations using cross-correlation |
| cpstruct2pairs | Extract valid control point pairs from `cpstruct` structure |
| imwarp | Apply geometric transformation to image |

### Spatial Referencing Objects

| imref2d | **Reference 2-D image to world coordinates** |
|---|---|
| imref3d | Reference 3-D image to world coordinates |

### Geometric Transformation Objects

| affine2d | **2-D affine geometric transformation** |
|---|---|

| affine3d | 3-D affine geometric transformation |
|---|---|
| projective2d | 2-D projective geometric transformation |

# Image Filtering and Enhancement

## Image Filtering

### Design Image Filters

| fspecial | **Create predefined 2-D filter** |
|---|---|
| fspecial3 | Create predefined 3-D filter |
| convmtx2 | 2-D convolution matrix |

### Basic Image Filtering in the Spatial Domain

| imfilter | **N-D filtering of multidimensional images** |
|---|---|
| roifilt2 | Filter region of interest (ROI) in image |
| nlfilter | General sliding-neighborhood operations |
| imgaussfilt | 2-D Gaussian filtering of images |
| imgaussfilt3 | 3-D Gaussian filtering of 3-D images |
| wiener2 | 2-D adaptive noise-removal filtering |
| medfilt2 | 2-D median filtering |
| medfilt3 | 3-D median filtering |
| modefilt | 2-D and 3-D mode filtering |
| ordfilt2 | 2-D order-statistic filtering |
| stdfilt | Local standard deviation of image |
| rangefilt | Local range of image |
| entropyfilt | Local entropy of grayscale image |
| imboxfilt | 2-D box filtering of images |
| imboxfilt3 | 3-D box filtering of 3-D images |
| fibermetric | Enhance elongated or tubular structures in image |
| maxhessiannorm | Maximum of Frobenius norm of Hessian of matrix |
| padarray | Pad array |

### Edge-Preserving Filtering

| imbilatfilt | **Bilateral filtering of images with Gaussian kernels** |
|---|---|
| imdiffuseest | Estimate parameters for anisotropic diffusion filtering |
| imdiffusefilt | Anisotropic diffusion filtering of images |
| imguidedfilter | Guided filtering of images |
| imnlmfilt | Non-local means filtering of image |
| burstinterpolant | Create high-resolution image from set of low-resolution burst mode images |

### Texture Filtering

| gabor | **Create Gabor filter or Gabor filter bank** |
|---|---|
| imgaborfilt | Apply Gabor filter or set of filters to 2-D image |

### Filtering By Property Characteristics

| bwareafilt | **Extract objects from binary image by size** |
|---|---|
| bwpropfilt | Extract objects from binary image using properties |

### Integral Image Domain Filtering

| integralImage | **Calculate 2-D integral image** |
|---|---|
| integralImage3 | Calculate 3-D integral image |
| integralBoxFilter | 2-D box filtering of integral images |
| integralBoxFilter3 | 3-D box filtering of 3-D integral images |

### Frequency Domain Filtering

| freqz2 | **2-D frequency response** |
|---|---|
| fsamp2 | 2-D FIR filter using frequency sampling |
| ftrans2 | 2-D FIR filter using frequency transformation |
| fwind1 | 2-D FIR filter using 1-D window method |
| fwind2 | 2-D FIR filter using 2-D window method |

## Contrast Adjustment

| imadjust | **Adjust image intensity values or color map** |
|---|---|
| imadjustn | Adjust intensity values in *N*-D volumetric image |
| imcontrast | Adjust Contrast tool |
| imsharpen | Sharpen image using unsharp masking |
| imflatfield | 2-D image flat-field correction |
| imlocalbrighten | Brighten low-light image |

| imreducehaze | Reduce atmospheric haze |
|---|---|
| locallapfilt | Fast local Laplacian filtering of images |
| localcontrast | Edge-aware local contrast manipulation of images |
| localtonemap | Render HDR image for viewing while enhancing local contrast |
| histeq | Enhance contrast using histogram equalization |
| adapthisteq | Contrast-limited adaptive histogram equalization (CLAHE) |
| imhistmatch | Adjust histogram of 2-D image to match histogram of reference image |
| imhistmatchn | Adjust histogram of N-D image to match histogram of reference image |
| decorrstretch | Apply decorrelation stretch to multichannel image |
| stretchlim | Find limits to contrast stretch image |
| intlut | Convert integer values using lookup table |
| imnoise | Add noise to image |

## Morphological Operations

### Perform Morphological Operations

| imerode | **Erode image** |
|---|---|
| imdilate | Dilate image |
| imopen | Morphologically open image |
| imclose | Morphologically close image |
| imtophat | Top-hat filtering |
| imbothat | Bottom-hat filtering |
| imclearborder | Suppress light structures connected to image border |
| imfill | Fill image regions and holes |
| bwhitmiss | Binary hit-miss operation |
| bwmorph | Morphological operations on binary images |
| bwmorph3 | Morphological operations on binary volume |
| bwperim | Find perimeter of objects in binary image |
| bwskel | Reduce all objects to lines in 2-D binary image or 3-D binary volume |
| bwulterode | Ultimate erosion |

### Perform Morphological Reconstruction

| imreconstruct | **Morphological reconstruction** |
|---|---|
| imregionalmax | Regional maxima |
| imregionalmin | Regional minima |
| imextendedmax | Extended-maxima transform |
| imextendedmin | Extended-minima transform |
| imhmax | H-maxima transform |
| imhmin | H-minima transform |
| imimposemin | Impose minima |

### Create Structuring Elements and Connectivity Arrays

| strel | **Morphological structuring element** |
|---|---|
| offsetstrel | Morphological offset structuring element |
| conndef | Create connectivity array |
| iptcheckconn | Check validity of connectivity argument |

### Create and Use Lookup Tables

| applylut | **Neighborhood operations on binary images using lookup tables** |
|---|---|
| bwlookup | Nonlinear filtering using lookup tables |
| makelut | Create lookup table for use with `bwlookup` |

### Pack Binary Images

| bwpack | **Pack binary image** |
|---|---|
| bwunpack | Unpack binary image |

## Deblurring

| deconvblind | **Deblur image using blind deconvolution** |
|---|---|
| deconvlucy | Deblur image using Lucy-Richardson method |
| deconvreg | Deblur image using regularized filter |
| deconvwnr | Deblur image using Wiener filter |
| edgetaper | Taper discontinuities along image edges |
| otf2psf | Convert optical transfer function to point-spread function |
| psf2otf | Convert point-spread function to optical transfer function |
| padarray | Pad array |

## ROI-Based Processing
### ROI Objects

| | |
|---|---|
| AssistedFreehand | Assisted freehand region of interest |
| Circle | Circular region of interest |
| Crosshair | Crosshair region of interest |
| Cuboid | Cuboidal region of interest |
| Ellipse | Elliptical region of interest |
| Freehand | Freehand region of interest |
| Line | Line region of interest |
| Point | Point region of interest |
| Polygon | Polygonal region of interest |
| Polyline | Polyline region of interest |
| Rectangle | Rectangular region of interest |
| draw | Begin drawing ROI interactively |

### ROI Creation Convenience Functions

| | |
|---|---|
| drawassisted | Create freehand ROI on image with assistance from image edges |
| drawcircle | Create customizable circular ROI |
| drawcrosshair | Create customizable crosshair ROI |
| drawcuboid | Create customizable cuboidal ROI |
| drawellipse | Create customizable elliptical ROI |
| drawfreehand | Create customizable freehand ROI |
| drawline | Create customizable linear ROI |
| drawpoint | Create customizable point ROI |
| drawpolygon | Create customizable polygonal ROI |
| drawpolyline | Create customizable polyline ROI |
| drawrectangle | Create customizable rectangular ROI |

### ROI Object Customization

| | |
|---|---|
| reduce | Reduce density of points in ROI |
| beginDrawingFromPoint | Begin drawing ROI from specified point |
| inROI | Query if points are located in ROI |
| bringToFront | Bring ROI to front of Axes stacking order |
| wait | Block MATLAB command line until ROI operation is finished |

### Mask Creation

| | |
|---|---|
| createMask | Create binary mask image from ROI |
| roipoly | Specify polygonal region of interest (ROI) |
| poly2mask | Convert region of interest (ROI) polygon to region mask |

### ROI Filtering

| | |
|---|---|
| regionfill | Fill in specified regions in image using inward interpolation |
| inpaintCoherent | Restore specific image regions using coherence transport based image inpainting |
| inpaintExemplar | Restore specific image regions using exemplar-based image inpainting |
| roicolor | Select region of interest (ROI) based on color |
| roifilt2 | Filter region of interest (ROI) in image |
| reducepoly | Reduce density of points in ROI using Ramer–Douglas–Peucker algorithm |

## Neighborhood and Block Processing

| | |
|---|---|
| blockproc | Distinct block processing for image |
| bestblk | Determine optimal block size for block processing |
| nlfilter | General sliding-neighborhood operations |
| col2im | Rearrange matrix columns into blocks |
| colfilt | Columnwise neighborhood operations |
| im2col | Rearrange image blocks into columns |
| ImageAdapter | Interface for image I/O |

## Image Arithmetic

| | |
|---|---|
| imabsdiff | Absolute difference of two images |
| imadd | Add two images or add constant to image |
| imapplymatrix | Linear combination of color channels |
| imcomplement | Complement image |
| imdivide | Divide one image into another or divide image by constant |

| imlincomb | Linear combination of images |
| immultiply | Multiply two images or multiply image by constant |
| imsubtract | Subtract one image from another or subtract constant from image |

## Image Segmentation and Analysis

### Image Segmentation
#### Segmentation Techniques

| graythresh | **Global image threshold using Otsu's method** |
| multithresh | Multilevel image thresholds using Otsu's method |
| otsuthresh | Global histogram threshold using Otsu's method |
| adaptthresh | Adaptive image threshold using local first-order statistics |
| grayconnected | Select contiguous image region with similar gray values using flood-fill technique |
| watershed | Watershed transform |
| activecontour | Segment image into foreground and background using active contours (snakes) region growing technique |
| lazysnapping | Segment image into foreground and background using graph-based segmentation |
| grabcut | Segment image into foreground and background using iterative graph-based segmentation |
| imseggeodesic | Segment image into two or three regions using geodesic distance-based color segmentation |
| imsegfmm | Binary image segmentation using fast marching method |
| gradientweight | Calculate weights for image pixels based on image gradient |
| graydiffweight | Calculate weights for image pixels based on grayscale intensity difference |
| imsegkmeans | K-means clustering based image segmentation |
| imsegkmeans3 | K-means clustering based volume segmentation |
| superpixels | 2-D superpixel oversegmentation of images |
| superpixels3 | 3-D superpixel oversegmentation of 3-D image |

#### Display Segmentation Results

| imoverlay | **Burn binary mask into 2-D image** |
| overlay | Overlay label matrix regions on 2-D image |
| label2idx | Convert label matrix to cell array of linear indices |
| boundarymask | Find region boundaries of segmentation |

#### Evaluate Segmentation Accuracy

| jaccard | **Jaccard similarity coefficient for image segmentation** |
| dice | Sørensen-Dice similarity coefficient for image segmentation |
| bfscore | Contour matching score for image segmentation |

### Object Analysis
#### Display Boundaries

| bwboundaries | **Trace region boundaries in binary image** |
| bwtraceboundary | Trace object in binary image |
| visboundaries | Plot region boundaries |

#### Detect Circles

| imfindcircles | **Find circles using circular Hough transform** |
| viscircles | Create circle |

#### Detect Edges and Gradients

| edge | **Find edges in intensity image** |
| edge3 | Find edges in 3-D intensity volume |
| imgradient | Find gradient magnitude and direction of 2-D image |
| imgradientxy | Find directional gradients of 2-D image |
| imgradient3 | Find gradient magnitude and direction of 3-D image |
| imgradientxyz | Find directional gradients of 3-D image |

#### Detect Lines

| hough | **Hough transform** |
| houghlines | Extract line segments based on Hough transform |
| houghpeaks | Identify peaks in Hough transform |
| radon | Radon transform |
| iradon | Inverse Radon transform |

**Detect Homogenous Blocks Using Quadtree Decomposition**

| | |
|---|---|
| qtdecomp | Quadtree decomposition |
| qtgetblk | Block values in quadtree decomposition |
| qtsetblk | Set block values in quadtree decomposition |

## Region and Image Properties

**Measure Properties of Image Regions**

| | |
|---|---|
| regionprops | Measure properties of image regions |
| regionprops3 | Measure properties of 3-D volumetric image regions |
| bwarea | Area of objects in binary image |
| bwconvhull | Generate convex hull image from binary image |
| bweuler | Euler number of binary image |
| bwferet | Measure Feret properties |
| bwperim | Find perimeter of objects in binary image |

**Measure Properties of Pixels and Paths**

| | |
|---|---|
| impixel | Pixel color values |
| improfile | Pixel-value cross-sections along line segments |
| imcontour | Create contour plot of image data |

**Measure Properties of Images**

| | |
|---|---|
| bwdist | Distance transform of binary image |
| bwdistgeodesic | Geodesic distance transform of binary image |
| graydist | Gray-weighted distance transform of grayscale image |
| imhist | Histogram of image data |
| mean2 | Average or mean of matrix elements |
| std2 | Standard deviation of matrix elements |
| corr2 | 2-D correlation coefficient |

**Find, Select, and Label Objects in Binary Images**

| | |
|---|---|
| bwconncomp | Find connected components in binary image |
| bwareaopen | Remove small objects from binary image |
| bwareafilt | Extract objects from binary image by size |
| bwpropfilt | Extract objects from binary image using properties |
| bwselect | Select objects in binary image |
| bwselect3 | Select objects in binary volume |
| bwlabel | Label connected components in 2-D binary image |
| bwlabeln | Label connected components in binary image |
| labelmatrix | Create label matrix from bwconncomp structure |
| label2rgb | Convert label matrix into RGB image |
| poly2label | Create label matrix from set of ROIs |
| poly2mask | Convert region of interest (ROI) polygon to region mask |

## Texture Analysis

| | |
|---|---|
| entropy | Entropy of grayscale image |
| entropyfilt | Local entropy of grayscale image |
| rangefilt | Local range of image |
| stdfilt | Local standard deviation of image |
| graycomatrix | Create gray-level co-occurrence matrix from image |
| graycoprops | Properties of gray-level co-occurrence matrix |

## Image Quality

**Full Reference Quality Metrics**

| | |
|---|---|
| immse | Mean-squared error |
| psnr | Peak signal-to-noise ratio (PSNR) |
| ssim | Structural similarity (SSIM) index for measuring image quality |
| multissim | Multiscale structural similarity (MS-SSIM) index for image quality |
| multissim3 | Multiscale structural similarity (MS-SSIM) index for volume quality |

**No-Reference Quality Metrics**

| | |
|---|---|
| brisque | Blind/Referenceless Image Spatial Quality Evaluator (BRISQUE) no-reference image quality score |
| fitbrisque | Fit custom model for BRISQUE image quality score |
| brisqueModel | Blind/Referenceless Image Spatial Quality Evaluator (BRISQUE) model |
| niqe | Naturalness Image Quality Evaluator (NIQE) no-reference image quality score |

| fitniqe | Fit custom model for NIQE image quality score |
|---------|-----------------------------------------------|
| niqeModel | Naturalness Image Quality Evaluator (NIQE) model |
| piqe | Perception based Image Quality Evaluator (PIQE) no-reference image quality score |

**Test Chart Based Quality Measurements**

| esfrChart | **Imatest edge spatial frequency response (eSFR) test chart** |
|-----------|---------------------------------------------------------------|
| colorChecker | X-Rite ColorChecker test chart |
| measureSharpness | Measure spatial frequency response using Imatest eSFR chart |
| measureChromaticAberration | Measure chromatic aberration at slanted edges using Imatest eSFR chart |
| measureColor | Measure color reproduction using test chart |
| measureNoise | Measure noise using Imatest eSFR chart |
| measureIlluminant | Measure scene illuminant using test chart |
| displayChart | Display test chart with overlaid regions of interest |
| displayColorPatch | Display measured and reference color as color patches |
| plotSFR | Plot spatial frequency response of edge |
| plotChromaticity | Plot color reproduction on chromaticity diagram |

## Image Transforms

| hough | **Hough transform** |
|-------|---------------------|
| houghlines | Extract line segments based on Hough transform |
| houghpeaks | Identify peaks in Hough transform |
| dct2 | 2-D discrete cosine transform |
| dctmtx | Discrete cosine transform matrix |
| fan2para | Convert fan-beam projections to parallel-beam |
| fanbeam | Fan-beam transform |
| idct2 | 2-D inverse discrete cosine transform |
| ifanbeam | Inverse fan-beam transform |
| iradon | Inverse Radon transform |
| para2fan | Convert parallel-beam projections to fan-beam |
| radon | Radon transform |

# Deep Learning for Image Processing

## Create Datastores for Image Preprocessing

| blockedImageDatastore | **Datastore for use with blocks from `blockedImage` objects** |
|-----------------------|---------------------------------------------------------------|
| denoisingImageDatastore | Denoising image datastore |
| randomPatchExtractionDatastore | Datastore for extracting random 2-D or 3-D random patches from images or pixel label images |

## Augment Images

| jitterColorHSV | **Randomly alter color of pixels** |
|----------------|------------------------------------|
| randomWindow2d | Randomly select rectangular region in image |
| randomCropWindow3d | Create randomized cuboidal cropping window |
| centerCropWindow2d | Create rectangular center cropping window |
| centerCropWindow3d | Create cuboidal center cropping window |
| Rectangle | Spatial extents of 2-D rectangular region |
| Cuboid | Spatial extents of 3-D cuboidal region |
| randomAffine2d | Create randomized 2-D affine transformation |
| randomAffine3d | Create randomized 3-D affine transformation |
| affineOutputView | Create output view for warping images |
| imerase | Remove image pixels within rectangular region of interest |

## Resize and Reshape Deep Learning Input

| resize2dLayer | **2-D resize layer** |
|---------------|----------------------|
| resize3dLayer | 3-D resize layer |
| dlresize | Resize spatial dimensions of `dlarray` object |
| DepthToSpace2DLayer | Depth to space layer |
| SpaceToDepthLayer | Space to depth layer |
| depthToSpace | Rearrange `dlarray` data from depth dimension into spatial blocks |
| spaceToDepth | Rearrange spatial blocks of `dlarray` data along depth dimension |

## Create Deep Learning Networks

| encoderDecoderNetwork | **Create encoder-decoder network** |
|-----------------------|------------------------------------|

| blockedNetwork | Create network with repeating block structure |
|---|---|
| pretrainedEncoderNetwork | Create encoder network from pretrained network |
| cycleGANGenerator | Create CycleGAN generator network for image-to-image translation |
| patchGANDiscriminator | Create PatchGAN discriminator network |
| pix2pixHDGlobalGenerator | Create pix2pixHD global generator network |
| addPix2PixHDLocalEnhancer | Add local enhancer network to pix2pixHD generator network |
| unitGenerator | Create unsupervised image-to-image translation (UNIT) generator network |
| unitPredict | Perform inference using unsupervised image-to-image translation (UNIT) network |

## Denoise Images

| denoiseImage | **Denoise image using deep neural network** |
|---|---|
| denoisingNetwork | Get image denoising network |
| dnCNNLayers | Get denoising convolutional neural network layers |

# 3-D Volumetric Image Processing

## Volume Display

| volshow | **Display volume** |
|---|---|
| labelvolshow | Display labeled volume |
| sliceViewer | Browse image slices |
| orthosliceViewer | Browse orthogonal slices in grayscale or RGB volume |
| obliqueslice | Extract oblique slice from 3-D volumetric data |

## Image Import and Conversion

| adaptthresh | **Adaptive image threshold using local first-order statistics** |
|---|---|
| dicomread | Read DICOM image |
| dicomreadVolume | Create 4-D volume from set of DICOM images |
| dicomContours | Extract ROI data from DICOM-RT structure set |
| imbinarize | Binarize 2-D grayscale image or 3-D volume by thresholding |
| niftiinfo | Read metadata from NIfTI file |
| niftiwrite | Write volume to file using NIfTI format |
| niftiread | Read NIfTI image |
| tiffreadVolume | Read volume from TIFF file |

## Image Arithmetic

| imabsdiff | **Absolute difference of two images** |
|---|---|
| imadd | Add two images or add constant to image |
| imdivide | Divide one image into another or divide image by constant |
| immultiply | Multiply two images or multiply image by constant |
| imsubtract | Subtract one image from another or subtract constant from image |

## Geometric Transformations and Image Registration

| affine3d | **3-D affine geometric transformation** |
|---|---|
| imcrop3 | Crop 3-D image |
| imref3d | Reference 3-D image to world coordinates |
| imregister | Intensity-based image registration |
| imregdemons | Estimate displacement field that aligns two 2-D or 3-D images |
| imresize3 | Resize 3-D volumetric intensity image |
| imrotate3 | Rotate 3-D volumetric grayscale image |
| imwarp | Apply geometric transformation to image |

## Image Filtering and Enhancement

| fibermetric | **Enhance elongated or tubular structures in image** |
|---|---|
| fspecial3 | Create predefined 3-D filter |
| histeq | Enhance contrast using histogram equalization |
| imadjustn | Adjust intensity values in $N$-D volumetric image |
| imboxfilt3 | 3-D box filtering of 3-D images |
| imfilter | N-D filtering of multidimensional images |
| imgaussfilt3 | 3-D Gaussian filtering of 3-D images |
| imhistmatchn | Adjust histogram of N-D image to match histogram of reference image |
| imnoise | Add noise to image |
| integralBoxFilter3 | 3-D box filtering of 3-D integral images |
| integralImage3 | Calculate 3-D integral image |

| medfilt3 | 3-D median filtering |
|---|---|

## Morphology

| bwareaopen | **Remove small objects from binary image** |
|---|---|
| bwconncomp | Find connected components in binary image |
| bwmorph3 | Morphological operations on binary volume |
| bwskel | Reduce all objects to lines in 2-D binary image or 3-D binary volume |
| imbothat | Bottom-hat filtering |
| imclose | Morphologically close image |
| imdilate | Dilate image |
| imerode | Erode image |
| imopen | Morphologically open image |
| imreconstruct | Morphological reconstruction |
| imregionalmax | Regional maxima |
| imregionalmin | Regional minima |
| imtophat | Top-hat filtering |
| offsetstrel | Morphological offset structuring element |
| strel | Morphological structuring element |
| watershed | Watershed transform |

## Image Segmentation

| activecontour | **Segment image into foreground and background using active contours (snakes) region growing technique** |
|---|---|
| bfscore | Contour matching score for image segmentation |
| dice | Sørensen-Dice similarity coefficient for image segmentation |
| gradientweight | Calculate weights for image pixels based on image gradient |
| graydiffweight | Calculate weights for image pixels based on grayscale intensity difference |
| imsegfmm | Binary image segmentation using fast marching method |
| imsegkmeans3 | K-means clustering based volume segmentation |
| jaccard | Jaccard similarity coefficient for image segmentation |
| superpixels3 | 3-D superpixel oversegmentation of 3-D image |

## Image Analysis

| bwselect3 | **Select objects in binary volume** |
|---|---|
| edge3 | Find edges in 3-D intensity volume |
| imgradient3 | Find gradient magnitude and direction of 3-D image |
| imgradientxyz | Find directional gradients of 3-D image |
| imhist | Histogram of image data |
| regionprops3 | Measure properties of 3-D volumetric image regions |

## Image Augmentation for Deep Learning

| randomPatchExtractionDatastore | **Datastore for extracting random 2-D or 3-D random patches from images or pixel label images** |
|---|---|
| centerCropWindow3d | Create cuboidal center cropping window |
| randomCropWindow3d | Create randomized cuboidal cropping window |
| randomAffine3d | Create randomized 3-D affine transformation |
| affineOutputView | Create output view for warping images |

# Hyperspectral Image Processing

## Explore, Analyze, and Visualize

| hypercube | **Read hyperspectral data** |
|---|---|
| enviwrite | Write hyperspectral data to ENVI file format |
| enviinfo | Read metadata from ENVI header file |
| selectBands | Select most informative bands |
| removeBands | Remove spectral bands from data cube |
| assignData | Assign new data to hyperspectral data cube |
| cropData | Crop regions-of-interest |
| colorize | Estimate color image of hyperspectral data |

## Filtering and Enhancement

| denoiseNGMeet | **Denoise hyperspectral images using non-local meets global approach** |
|---|---|
| sharpencnmf | Sharpen hyperspectral data using coupled nonnegative matrix factorization (CNMF) method |

## Data Correction
### Radiometric Calibration

| | |
|---|---|
| dn2radiance | **Convert digital number to radiance** |
| dn2reflectance | Convert digital number to reflectance |
| radiance2Reflectance | Convert radiance to reflectance |

### Atmospheric Correction

| | |
|---|---|
| correctOOB | **Correct out-of-band effect using sensor spectral response** |
| empiricalLine | Empirical line calibration of hyperspectral data |
| fastInscene | Perform fast in-scene atmospheric correction |
| flatField | Apply flat field correction to hyperspectral data cube |
| iarr | Apply internal average relative reflectance (IARR) correction to hyperspectral data cube |
| logResiduals | Apply log residual correction to hyperspectral data cube |
| rrs | Compute remote sensing reflectance |
| subtractDarkPixel | Subtract dark pixel value from hyperspectral data cube |
| sharc | Perform atmospheric correction using satellite hypercube atmospheric rapid correction (SHARC) |

### Spectral Correction

| | |
|---|---|
| reduceSmile | **Reduce spectral smile effect in hyperspectral data cube** |

## Dimensionality Reduction

| | |
|---|---|
| hyperpca | **Principal component analysis of hyperspectral data** |
| hypermnf | Maximum noise fraction transform of hyperspectral data |
| inverseProjection | Reconstruct data cube from principal component bands |

## Spectral Unmixing

| | |
|---|---|
| ppi | **Extract endmember signatures using pixel purity index** |
| fippi | Extract endmember signatures using fast iterative pixel purity index |
| nfindr | Extract endmember signatures using N-FINDR |
| countEndmembersHFC | Find number of endmembers |
| estimateAbundanceLS | Estimate abundance maps |

## Spectral Matching and Target Detection

| | |
|---|---|
| readEcostressSig | **Read data from ECOSTRESS spectral library** |
| sam | Measure spectral similarity using spectral angle mapper |
| sid | Measure spectral similarity using spectral information divergence |
| jmsam | Measure spectral similarity using Jeffries Matusita-Spectral Angle Mapper method |
| sidsam | Measure spectral similarity using spectral information divergence-spectral angle mapper hybrid method |
| ns3 | Measure normalized spectral similarity score |
| spectralMatch | Identify unknown regions or materials using spectral library |
| spectralIndices | Compute hyperspectral indices |
| ndvi | Normalized vegetation index |
| anomalyRX | Detect anomalies using Reed-Xiaoli detector |

# APPENDIX    *B*

## Face Datasets

| Database | Samples | Sensor | Usage | Data type | Dimensions | Expressions | Year | Ref |
|----------|---------|--------|-------|-----------|------------|-------------|------|-----|
| Eurecom Kinect Face | 14 male and 38 females | Kinect V.1 | FER – Face Recognition | 1248 Color + depth images | RGB= 256*256 Depth= 256*256 | 3 expressions of neutral, joy and surprise | 2014 | [89] |
| VAP RGB-D Face | 13 | Kinect V.1 | FER – Face Recognition | 2960 color + depth images | RGB= 351*421 Depth= 480*640 | 4 expressions of neutral, joy, anger, surprise and sadness | 2012 | [58] |
| VAP RGB-D-T Face | 51 | Kinect V.1 And AXIS Q1922 | FER – Face Recognition | 46360 color + depth + thermal images | RGB= 480*640 Depth= 480*640 Thermal= 288*384 | 4 expressions of neutral, joy, anger and surprise | 2014 | [90] |
| Curtin Face | 25 | Kinect V.1 | FER – Face Recognition | 5000 color + depth images | RGB= 480*640 Depth= 480*640 | 7 main expressions | 2013 | [91] |
| FEEDB | 50 | Kinect V.1 | FER - FMER – Face Recognition | 30 color and depth videos | RGB= 480*640 Depth= 480*640 | 33 facial expressions | 2013 | [55] |
| Face Grabber | 33 male and 7female | Kinect V.2 | FER - FMER – Face Recognition | 67159 color + depth images | RGB= 2080*1920 Depth= 424*512 | 7 main expressions | 2016 | [92] |
| SMIC | 16 | Pixel INK PL-B774U | FMER | 164 video files in 100 frame fps | 640*480 | 4 expressions | 2013 | [93] |
| CASME | 19 | BenQ M31 GRAS-03K2C | FMER | 195 video files in 60 fps | 640*480 1280*780 | 7 expressions | 2013 | [51] |
| Polikovsky's | 10 | Grasshopper | FMER | 200 video files in 200 fps | 640*480 | 13 expressions | 2009 | [94] |
| USF-HD | 100 | - | FER - FMER | 56 video files in 30 fps | 1280*780 | 6 expressions | 2011 | [95] |
| YorkDDT | 9 | - | FER | 30 fps | 640*480 | 18 expressions | 2009 | [96] |
| JAFFE | 10 | - | FER – Face Recognition | 212 gray images | 256*256 | 7 main expressions | 1998 | [49] |
| KDEF | 70 | Pentax LX | FER – Face Recognition | 5041 Color Images | 562*762 | 7 main expressions | 1998 | [35] |
| IKFDB | 40 | Kinect V.2 | FER - FMER – Face Recognition | 100000 + Color and Depth Images | RGB= 2080*1920 Depth= 424*512 | 7 main expressions + Pitch, Yaw and Roll | 2021 | [32] |

# Appendix C

## Advanced FER Using Metaheuristics and Neural Gas Networks (NGN)

- ## Metaheuristics

Metaheuristics are problem-solving methodologies that are designed to find good approximate solutions for complex optimization problems [109]. They are general-purpose algorithms that can be applied to a wide range of problems. They are particularly useful when traditional exact methods, such as mathematical or dynamic programming, are not feasible or efficient. Metaheuristics are inspired by natural phenomena, optimization principles, and computational intelligence [97, 98]. They provide a flexible framework for exploring and exploiting the search space of a problem, allowing them to escape local optima and find near-optimal or satisfactory solutions. Unlike exact algorithms, which guarantee optimal solutions but may be computationally expensive or infeasible for large-scale problems, metaheuristics aim to efficiently explore the solution space and provide acceptable solutions within a reasonable amount of time. They trade off solution quality for computational efficiency, making them suitable for real-world problems where finding exact solutions is impractical. Common examples of metaheuristic algorithms include: 1. Genetic Algorithms (GA) [99]: Inspired by the process of natural evolution, GA uses mechanisms such as selection, crossover, and mutation to iteratively improve a population of candidate solutions. 2. Particle Swarm Optimization (PSO) [100, 108]: Inspired by the social behavior of bird flocking or fish schooling, PSO uses a population of particles that move in the search space and exchange information to find promising solutions. 3. Simulated Annealing (SA) [101]: Inspired by the annealing process in metallurgy, SA starts with an initial solution and iteratively explores the solution space by allowing occasional "worse" moves, gradually reducing the search radius over time. 4. Tabu Search (TS) [102]: TS maintains a tabu list of previously visited solutions to avoid cycling and uses neighborhood exploration to find better solutions. 5. Ant Colony Optimization (ACO) [103]: Inspired by the behavior of ants searching for food, ACO uses pheromone trails and stochastic decision rules to guide the search process. Here, three algorithms namely Weevil Damage Optimization Algorithm (WDOA) [104], the Bee-Eater Hunting strategy algorithm (BEH) [105], and Victoria Amazonica Algorithm (VAO) [106] are employed for various image processing [107] tasks such as image contrast enhancement, image segmentation, image quantization, and feature selection out of extracted features. WDOA mimics weevils' natural behavior over agricultural products in order to search for food using weevils' fly power, snout power, and damage elements in the algorithm. BEH simulates the bird bee-eater's hunting strategy over prey in the sky. Also, the VAO algorithm mimics the life cycle of the giant water lily plant for optimization purposes. These are just a few examples, and there are many other metaheuristic algorithms, each with its own unique characteristics and strengths. Metaheuristics have found applications in various domains, including logistics, scheduling, engineering, finance, and more.

Neural Gas Network (NGN) is a type of unsupervised learning algorithm that is used for data clustering and visualization tasks. It is inspired by the self-organizing map (SOM) algorithm [110] and was introduced by Martinetz and Schulten in 1991 [111]. Neural Gas networks are based on a collection of artificial neurons or nodes arranged in a one-dimensional or multi-dimensional lattice. Each neuron represents a prototype or codebook vector that is associated with a specific region in the input space. During the learning process, the NG algorithm adjusts the positions of these neurons to adapt to the distribution of input data. The learning process in Neural Gas involves the following steps: 1. Initialization: The positions of the neurons are randomly initialized in the input space. 2. Input presentation: Input samples are presented to the network one by one. 3. Distance calculation: The distances between the input sample and all neurons are calculated. The distance can be defined using various metrics, such as Euclidean distance or Manhattan distance. 4. Winning neuron selection: The neuron with the smallest distance to the input sample is selected as the winner. 5. Update: The winning neuron and its neighboring neurons in the lattice are updated to better represent the input sample. The update can involve adjusting the positions of the neurons and updating their connection weights. 6. Repeat: Steps 3-5 are repeated for a fixed number of iterations or until convergence. The Neural Gas algorithm gradually adapts the neuron positions to form a representation of the input space. The neurons that are closer to a specific region of the input space become more sensitive to the input samples from that region. Neural Gas networks have several applications, including data clustering, data visualization, feature extraction, and anomaly detection. They can be useful in exploratory data analysis, pattern recognition, and understanding the structure of high-dimensional data. Here NGN is used for image segmentation and quantization of facial samples (depth) [112].

## • **WDOA Contrast Adjustment**

<p style="text-align:center;">***"ClusterCost.m"***</p>

```
function [z, out] = ClusterCost(m, X)
% Calculate Distance Matrix
d = pdist2(X, m);
% Assign Clusters and Find Closest Distances
[dmin, ind] = min(d, [], 2);
% Sum of Within-Cluster Distance
WCD = sum(dmin);
z=WCD;
out.d=d;
out.dmin=dmin;
out.ind=ind;
out.WCD=WCD;
end
```

<p style="text-align:center;">***"RouletteWheelS.m"***</p>

```
function j = RouletteWheelS(P)
r = rand;
C = cumsum(P);
```

```matlab
j = find(r <= C, 1, 'first');
end
```

*"WDOAContrastEnhancment.m"*

```matlab
%% Weevil Damage Optimization Algorithm (WDOA) Image Contrast
Enhancement by Clustering
clear;
clc;
close all;
warning('off');
% Loading
img=imread('tst1.jpg');
img=im2double(img);
gray=rgb2gray(img);
% Reshaping image to vector
X=gray(:);
%% Starting WDOA
k = 6;
CostFunction=@(m) ClusterCost(m, X);         % Cost Function
VarSize=[k size(X,2)];                  % Decision Variables Matrix Size
nVar=prod(VarSize);                     % Number of Decision Variables
VarMin= repmat(min(X),k,1);             % Lower Bound of Variables
VarMax= repmat(max(X),k,1);             % Upper Bound of Variables
%% WDOA Parameters
MaxIt = 100;                            % Maximum Number of Iterations
nPop = 10;                              % Number of weevils
DamageRate = 0.3;                       % Damage Rate
nweevil = round(DamageRate*nPop);       % Number of Remained weevils
nNew = nPop-nweevil;                    % Number of New weevils
mu = linspace(1, 0, nPop);              % Mutation Rates
pMutation = 0.1;                        % Mutation Probability
MUtwo = 1-mu;                           % Second Mutation
SnoutPower = 0.8;                       % Weevil Snout power Rate
FlyPower = 0.003;                       % Weevil Fly Power Rate
%----------------------------------------
%% Basics
% Empty weevil
weevil.Position = [];
weevil.Cost = [];
% Weevils Array
pop = repmat(weevil, nPop, 1);
% First weevils
```

```matlab
for i = 1:nPop
pop(i).Position = unifrnd(VarMin, VarMax, VarSize);
pop(i).Cost = CostFunction(pop(i).Position);
end;
% Sort
[~, SortOrder] = sort([pop.Cost]);pop = pop(SortOrder);
% Best Solution
BestSol = pop(1);
% Best Costs Array
BestCost = zeros(MaxIt, 1);
%-------------------------------

%% WDOA Body
for it = 1:MaxIt
newpop = pop;
for i = 1:nPop
for k = 1:nVar
if rand <= MUtwo(i)
TMP = mu;TMP(i) = 0;TMP = TMP/sum(TMP);
j = RouletteWheelS(TMP);
newpop(i).Position(k) =
pop(i).Position(k)+SnoutPower*(pop(j).Position(k)-
pop(i).Position(k)+FlyPower);
end;
% Mutation
if rand <= pMutation
newpop(i).Position(k) = newpop(i).Position(k);
end;end;
% Apply Lower and Upper Bound Limits
newpop(i).Position = max(newpop(i).Position, VarMin);
newpop(i).Position = min(newpop(i).Position, VarMax);
newpop(i).Cost = CostFunction(newpop(i).Position);end;% Asses power
[~, SortOrder] = sort([newpop.Cost]);newpop = newpop(SortOrder);% Sort
pop = [pop(1:nweevil);newpop(1:nNew);% Select
[~, SortOrder] = sort([pop.Cost]);pop = pop(SortOrder);% Sort
BestSol = pop(1);% Update
BestCost(it) = BestSol.Cost;% Store
% Iteration
disp(['In Iteration No ' num2str(it) ': WDOA Best Cost = '
num2str(BestCost(it))]);
end;
% Plot
figure;
Thresh=sort(BestSol.Position);
Adj = imadjust(img,[Thresh(1) Thresh(2) Thresh(3); Thresh(4) Thresh(5)
Thresh(6)]);
```

```
subplot(2,2,1);
imshow(img); title('Original');
subplot(2,2,2)
imshow(Adj,[]);title('WDOA Enhanced');
subplot(2,2,[3 4])
plot(BestCost,'k', 'LineWidth', 2);
title(['WDOA Best Cost is :  ',num2str(BestCost(it))])
xlabel('Iteration');
ylabel('Cost Value');
ax = gca;
ax.FontSize = 10;
ax.FontWeight='bold';
set(gca,'Color','c')
grid on;
```

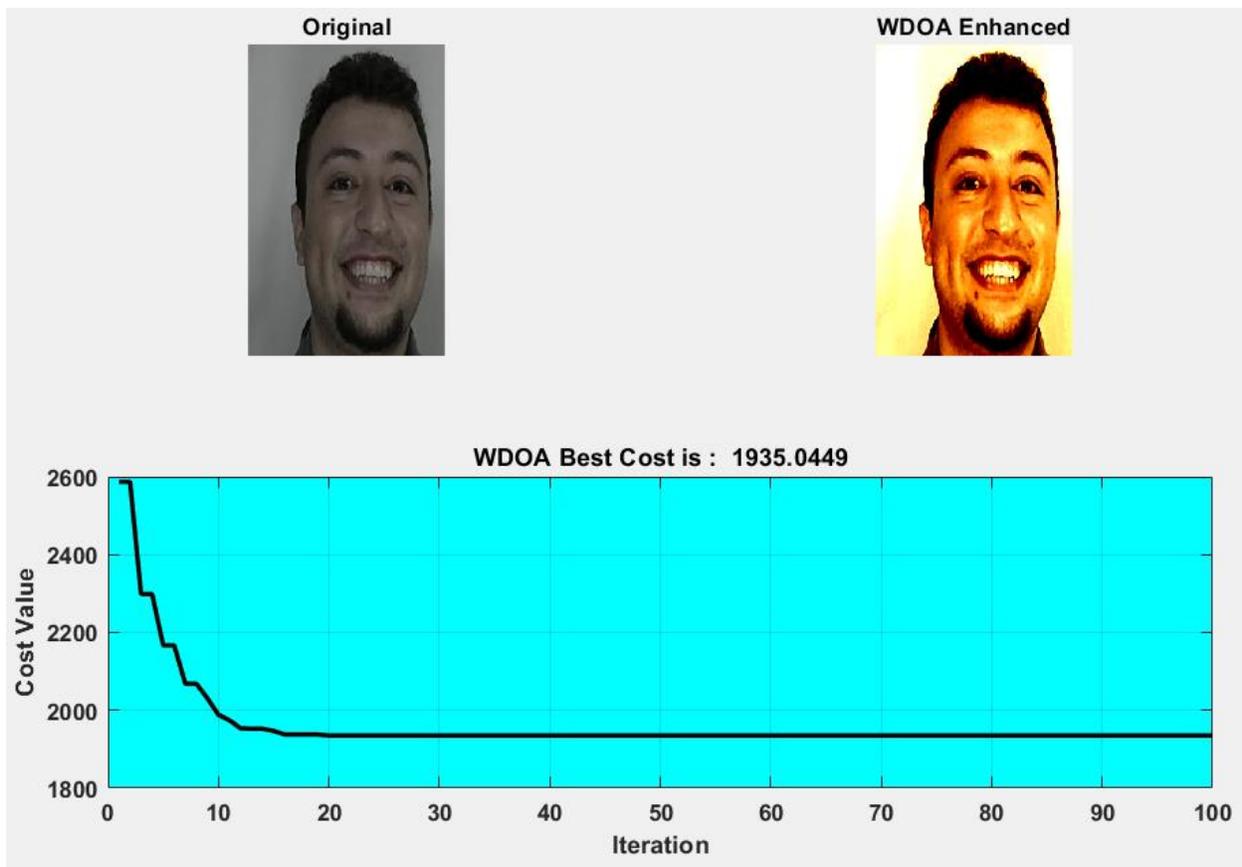

## • **BEH Image Segmentation and Quantization**

*"ClusterCost.m"* and *"RouletteWheelS.m"* are as WDOA contrast enhancement.

*"Res.m"*

```matlab
function m=Res(X, sol)
% Cluster Centers
m = sol.Position;
k = size(m,1);
% Cluster Indices
ind = sol.Out.ind;
Colors = hsv(k);
for j=1:k
Xj = X(ind==j,:);
end
end
```

*"BEHAlgorithmImageSegmentation.m"*

```matlab
%% Bee-Eater Hunting (BEH) Strategy Algorithm for Image Segmentation
and Quantization
clear;
clc;
close all;
warning('off');
img=imread('test2.jpg');
img=im2double(img);
gray=rgb2gray(img);
gray=imadjust(gray);
warning ('off');
% Separating color channels
R=img(:,:,1);
G=img(:,:,2);
B=img(:,:,3);
% Reshaping each channel into a vector and combine all three channels
X=[R(:) G(:) B(:)];

%% Starting BEH
k = 5; % Number of Colors (segments)
%------------------------------------------------------
CostFunction=@(m) ClusterCost(m, X);       % Cost Function
VarSize=[k size(X,2)];                      % Decision Variables Matrix Size
nVar=prod(VarSize);                         % Number of Decision Variables
VarMin= repmat(min(X),k,1);                 % Lower Bound of Variables
VarMax= repmat(max(X),k,1);                 % Upper Bound of Variables
MaxIt = 200;                                % Maximum Number of Iterations
nPop = 20;                                  % Number of bee-eaters
DamageRate = 0.2;                           % Damage Rate
```

```matlab
nbeeeater = round(DamageRate*nPop);     % Number of Remained beeeaters
nNew = nPop-nbeeeater;                   % Number of New beeeaters
mu = linspace(1, 0, nPop);              % Mutation Rates
pMutation = 0.1;                         % Mutation Probability
MUtwo = 1-mu;                            % Fight Mutation
PeakPower = 0.8;                         % BeeEater Peack power Rate
AdjustPower = 0.03*(VarMax-VarMin);      % BeeEater Adjustment Power
Rate
PYR= -0.2;                               % Pitch Yaw Roll Rate
%----------------------------------------
%% Basics
% Empty bee-eater
beeeater.Position = [];
beeeater.Cost = [];
% BeeEaters Array
pop = repmat(beeeater, nPop, 1);
% First bee-eaters
for i = 1:nPop
pop(i).Position = unifrnd(VarMin, VarMax, VarSize);
[pop(i).Cost pop(i).Out] = CostFunction(pop(i).Position);end;
% Sort
[~, SortOrder] = sort([pop.Cost]);pop = pop(SortOrder);
% Best Solution
BestSol = pop(1);
% Best Costs Array
BestCost = zeros(MaxIt, 1);
%-------------------------------
%% BEH Body
for it = 1:MaxIt
newpop = pop;
for i = 1:nPop
for k = 1:nVar
if rand <= MUtwo(i)
TMP = mu;TMP(i) = 0;TMP = TMP/sum(TMP);
j = RouletteWheelS(TMP);
newpop(i).Position(k) =
pop(i).Position(k)*PYR+PeakPower*(pop(j).Position(k)-
pop(i).Position(k));
end;
% Mutation
if rand <= pMutation
newpop(i).Position(k) = newpop(i).Position(k)+PYR;
end;end;
% Apply Lower and Upper Bound Limits
newpop(i).Position = max(newpop(i).Position, VarMin);
newpop(i).Position = min(newpop(i).Position, VarMax);
```

```matlab
[newpop(i).Cost newpop(i).Out] =
CostFunction(newpop(i).Position+AdjustPower);end;% Asses power
[~, SortOrder] = sort([newpop.Cost]);newpop = newpop(SortOrder);% Sort
pop = [pop(1:nbeeeater);newpop(1:nNew)];% Select
[~, SortOrder] = sort([pop.Cost]);pop = pop(SortOrder);% Sort
BestSol = pop(1);% Update
BestCost(it) = BestSol.Cost;% Store
% Iteration
disp(['In Iteration No ' num2str(it) ': BeeEater Optimizer Best Cost =
' num2str(BestCost(it))]);
BEHCenters=Res(X, BestSol);
end;
BEHlbl=BestSol.Out.ind;
%% Converting cluster centers and its indexes into image
Z=BEHCenters(BEHlbl,:);
R2=reshape(Z(:,1),size(R));
G2=reshape(Z(:,2),size(G));
B2=reshape(Z(:,3),size(B));
% Attaching color channels (BEH Quantization Result)
quantized=zeros(size(img));
quantized(:,:,1)=R2;
quantized(:,:,2)=G2;
quantized(:,:,3)=B2;
% BEH Segmentation Result
gray2=reshape(BEHlbl(:,1),size(gray));
segmented = label2rgb(gray2);
%% Plot Results
figure;
subplot (2,2,1)
imshow(img);title('Original');
subplot (2,2,2)
imshow(quantized);title('Quantized Image');
subplot(2,2,3);
imshow(segmented,[]);title('Segmented Image');
subplot(2,2,4);
plot(BestCost,'k','LineWidth',2);
xlabel('Iteration');
ylabel('Best Cost');
title(['BEH Best Cost is :  ',num2str(BestCost(it))])
ax = gca;
ax.FontSize = 10;
ax.FontWeight='bold';
set(gca,'Color','c')
grid on;
```

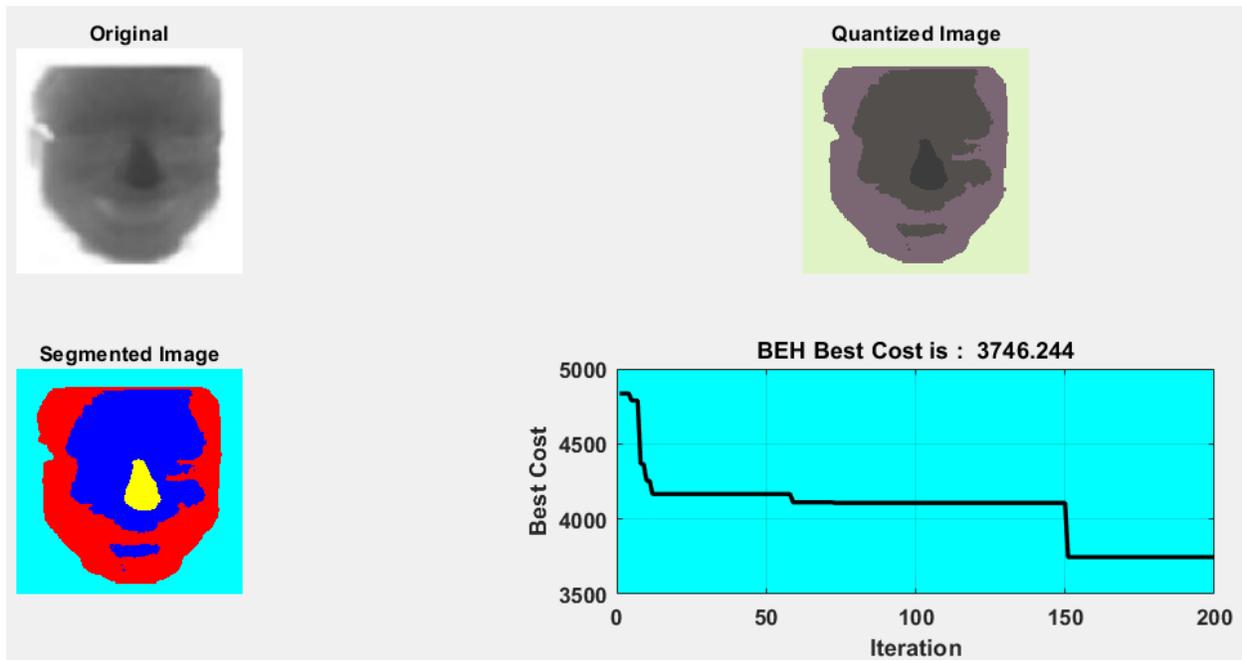

- **Neural Gas Network (NGN) Image Segmentation and Quantization**

*"GasNN.m"*

```
function NGNnetwork = GasNN(X, params)
%% Load
nData = size(X,1);
nDim = size(X,2);
X = X(randperm(nData),:);
Xmin = min(X);
Xmax = max(X);
%% Params
N = params.N;
MaxIt = params.MaxIt;
tmax = params.tmax;
epsilon_initial = params.epsilon_initial;
epsilon_final = params.epsilon_final;
lambda_initial = params.lambda_initial;
lambda_final = params.lambda_final;
T_initial = params.T_initial;
T_final = params.T_final;
%% Initialization
```

```matlab
w = zeros(N, nDim);
for i = 1:N
w(i,:) = unifrnd(Xmin, Xmax);
end
C = zeros(N, N);
t = zeros(N, N);
tt = 0;

%% Body
for it = 1:MaxIt
for l = 1:nData
% Slect Input Vector
x = X(l,:);
% Competion and Ranking
d = pdist2(x,w);
[~, SortOrder] = sort(d);
% Calculate Parameters
epsilon = epsilon_initial*(epsilon_final/epsilon_initial)^(tt/tmax);
lambda = lambda_initial*(lambda_final/lambda_initial)^(tt/tmax);
T = T_initial*(T_final/T_initial)^(tt/tmax);
% Adaptation
for ki = 0:N-1
i=SortOrder(ki+1);
w(i,:) = w(i,:) + epsilon*exp(-ki/lambda)*(x-w(i,:));
end
tt = tt + 1;
% Creating Links
i = SortOrder(1);
j = SortOrder(2);
C(i,j) = 1;
C(j,i) = 1;
t(i,j) = 0;
t(j,i) = 0;
% Aging
t(i,:) = t(i,:) + 1;
t(:,i) = t(:,i) + 1;
% Remove Old Links
L = t(i,:)>T;
C(i, L) = 0;
C(L, i) = 0;
end
disp(['Iteration ' num2str(it)]);
end
%% Results
NGNnetwork.w = w;
NGNnetwork.C = C;
```

```matlab
NGNnetwork.t = t;
end
```

*"NGNImageSegmentation.m"*

```matlab
%% Neural Gas Network (NGN) Image Segmentation and Quantization
clc;
clear;
close all;
%% Load Image
Org=imread('tst5.jpg');
X = rgb2gray(Org);
X=double(X);
img=X;
X=X(:)';
%% Neural Gas Network (NGN) Parameters
ParVal.N = 8; % Number of Segments
ParVal.MaxIt = 100; % Number of runs
ParVal.tmax = 100000;
ParVal.epsilon_initial = 0.3;
ParVal.epsilon_final = 0.02;
ParVal.lambda_initial = 2;
ParVal.lambda_final = 0.1;
ParVal.T_initial = 5;
ParVal.T_final = 10;
%% Training Neural Gas Network
NGNnetwok = GasNN(X, ParVal);
%% Vector to image and plot
Weight=sum(round(rescale(NGNnetwok.w,1,ParVal.N)));
Weight=round(rescale(Weight,1,ParVal.N));
indexed=reshape(Weight(1,:),size(img));
segmented = label2rgb(indexed);
% Plot Res
figure('units','normalized','outerposition',[0 0 1 1])
subplot(1,3,1)
imshow(Org,[]); title('Original');
subplot(1,3,2)
imshow(segmented);
title(['Segmented in [' num2str(ParVal.N) '] Segments']);
subplot(1,3,3)
imshow(indexed,[]);
title(['Quantized in [' num2str(ParVal.N) '] Thresholds']);
```

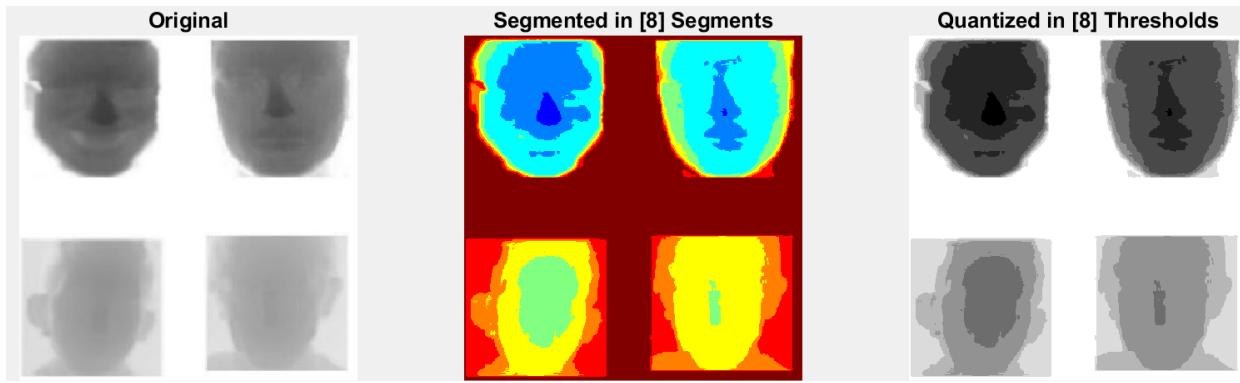

| Original | Segmented in [8] Segments | Quantized in [8] Thresholds |

- **VAO Feature Selection**

*"LoadData.m"*

```
function data=LoadData()
dataset=load('FER2');
dataset=dataset.FER2;
data.x=dataset.Inputs;
data.t=dataset.Targets;
data.nx=size(data.x,1);
data.nt=size(data.t,1);
data.nSample=size(data.x,2);
end
```

*"CreateAndTrainANN.m"*

```
function results=CreateAndTrainANN(x,t)
if ~isempty(x)
% Choose a Training Function
% For a list of all training functions type: help nntrain
% 'trainlm' is usually fastest.
% 'trainbr' takes longer but may be better for challenging problems.
% 'trainscg' uses less memory. NFTOOL falls back to this in low memory
situations.
trainFcn = 'trainlm';  % Levenberg-Marquardt
% Create a Fitting Network
hiddenLayerSize = 10;
net = fitnet(hiddenLayerSize,trainFcn);
% Choose Input and Output Pre/Post-Processing Functions
% For a list of all processing functions type: help nnprocess
```

```matlab
net.input.processFcns = {'removeconstantrows','mapminmax'};
net.output.processFcns = {'removeconstantrows','mapminmax'};
% Setup Division of Data for Training, Validation, Testing
% For a list of all data division functions type: help nndivide
net.divideFcn = 'dividerand';  % Divide data randomly
net.divideMode = 'sample';  % Divide up every sample
net.divideParam.trainRatio = 70/100;
net.divideParam.valRatio = 15/100;
net.divideParam.testRatio = 15/100;
% Choose a Performance Function
% For a list of all performance functions type: help nnperformance
net.performFcn = 'mse';  % Mean squared error
% Choose Plot Functions
% For a list of all plot functions type: help nnplot
net.plotFcns = {};
% net.plotFcns = {'plotperform','plottrainstate','ploterrhist',
'plotregression', 'plotfit'};
net.trainParam.showWindow=false;
% Train the Network
[net,tr] = train(net,x,t);
% Test the Network
y = net(x);
e = gsubtract(t,y);
E = perform(net,t,y);
else
y=inf(size(t));
e=inf(size(t));
E=inf;
tr.trainInd=[];
tr.valInd=[];
tr.testInd=[];
end
% All Data
Data.x=x;
Data.t=t;
Data.y=y;
Data.e=e;
Data.E=E;
% Train Data
TrainData.x=x(:,tr.trainInd);
TrainData.t=t(:,tr.trainInd);
TrainData.y=y(:,tr.trainInd);
TrainData.e=e(:,tr.trainInd);
if ~isempty(x)
TrainData.E=perform(net,TrainData.t,TrainData.y);
else
```

```matlab
TrainData.E=inf;
end
% Validation and Test Data
TestData.x=x(:,[tr.testInd tr.valInd]);
TestData.t=t(:,[tr.testInd tr.valInd]);
TestData.y=y(:,[tr.testInd tr.valInd]);
TestData.e=e(:,[tr.testInd tr.valInd]);
if ~isempty(x)
TestData.E=perform(net,TestData.t,TestData.y);
else
TestData.E=inf;
end
% Export Results
if ~isempty(x)
results.net=net;
else
results.net=[];
end
results.Data=Data;
results.TrainData=TrainData;
% results.ValidationData=ValidationData;
results.TestData=TestData;
end
```

*"FeatureSelectionCost.m"*

```matlab
function [z, out]=FeatureSelectionCost(u,nf,data)
% Read Data Elements
x=data.x;
t=data.t;
% Create Permutation uinsg Random Keys
[~, q]=sort(u);
% Selected Features
S=q(1:nf);
% Ratio of Selected Features
rf=nf/numel(q);
% Selecting Features
xs=x(S,:);
% Weights of Train and Test Errors
wTrain=0.8;
wTest=1-wTrain;
% Number of Runs
nRun=3;
```

```matlab
EE=zeros(1,nRun);
for r=1:nRun
% Create and Train ANN
results=CreateAndTrainANN(xs,t);
% Calculate Overall Error
EE(r) = wTrain*results.TrainData.E + wTest*results.TestData.E;
end
E=mean(EE);
% Calculate Final Cost
z=E;
% Set Outputs
out.S=S;
out.nf=nf;
out.rf=rf;
out.E=E;
out.z=z;
end
```

<p style="text-align:center"><em><strong><span style="color:red">"VAOFS.m"</span></strong></em></p>

```matlab
%% Victoria Amazonica Optimization (VAO) Feature Selection
clc;
clear;
close all;
warning('off');
%% Problem Definition
data=LoadData();
nf=7;    % Desired Number of Selected Features
CostFunction=@(u) FeatureSelectionCost(u,nf,data);        % Cost
Function
nVar=data.nx;
VarSize = [1 nVar];    % Decision Variables Matrix Size
VarMin = -10;          % Decision Variables Lower Bound
VarMax = 10;           % Decision Variables Upper Bound
%% VAO Algorithm Parameters
MaxIt = 50;                  % Maximum Number of Iterations
nPop = 3;                    % Number of Plants (Leaf anf Flower (xi,
..., xn))
omega = 5;    % Drawback coefficient of ω
psi = 4;      % Drawback coefficient of ψ
lambda = 2;    % Intra Competition Rate of λ
mu = 0.2;                        % Mutation Coefficient of μ
mu_damp = 0.98;                  % Mutation Coefficient Damping
Ratio
```

```matlab
delta = 0.05*(VarMax-VarMin);        % Uniform Mutation Range
%----------------------------------------
if isscalar(VarMin) && isscalar(VarMax)
dmax = (VarMax-VarMin)*sqrt(nVar);
else
dmax = norm(VarMax-VarMin);
end
%% Basics
% Empty Plants Structure
% Position = Plant place on the pond surface
% Cost = Plant Expansion Value in Diameter or ⌀
plants.Position = [];
plants.Cost = [];
plants.Sol = [];
% Initialize Population Array
pop = repmat(plants, nPop, 1);
% Initialize Best Solution Ever Found
BestSol.Cost = inf;
% Create Initial Plants
for i = 1:nPop
pop(i).Position = unifrnd(VarMin, VarMax, VarSize);
[pop(i).Cost pop(i).Sol] = CostFunction(pop(i).Position);
if pop(i).Cost <= BestSol.Cost
BestSol = pop(i);
end;end
% Array to Hold Best Cost Values
BestCost = zeros(MaxIt, 1);
%% VAO Algorithm Main Body
for it = 1:MaxIt
newpop = repmat(plants, nPop, 1);
for i = 1:nPop
newpop(i).Cost = inf;
for j = 1:nPop
if pop(j).Cost < pop(i).Cost
rij = norm(pop(i).Position-pop(j).Position)/dmax;
beta = psi*exp(-omega*rij^lambda);
e = delta*unifrnd(-1, +1, VarSize);
%----------------
newsol.Position = pop(i).Position ...
+ beta*rand(VarSize).*(pop(j).Position-pop(i).Position) ...
+ mu*e;
%----------------
newsol.Position = max(newsol.Position, VarMin);
newsol.Position = min(newsol.Position, VarMax);
%----------------
[newsol.Cost newsol.Sol] = CostFunction(newsol.Position);
```

```matlab
%----------------
if newsol.Cost <= newpop(i).Cost
newpop(i) = newsol;
if newpop(i).Cost <= BestSol.Cost
BestSol = newpop(i);
AllSol(i)=newpop(i);
end;end;end;end;end
% Merge
pop = [pop
newpop];
% Sort
[~, SortOrder] = sort([pop.Cost]);
pop = pop(SortOrder);
% Truncate
pop = pop(1:nPop);
% Store Best Cost Ever Found
BestCost(it) = BestSol.Cost;
alpha=BestSol.Cost;
% Show Iteration Information
disp(['In Iteration No ' num2str(it) ': VAO Best Cost Is = '
num2str(BestCost(it))]);
% Damp Mutation Coefficient
mu = mu*mu_damp;
end
%% VAO Feature Matrix
% Extracting Data
RealData=data.x';
% Extracting Labels
RealLbl=data.t';
FinalFeaturesInd=BestSol.Sol.S;
% Sort Features
FFI=sort(FinalFeaturesInd);
% Select Final Features
VAO_Features=RealData(:,FFI);
% Adding Labels
VAO_Features_Lbl=VAO_Features;
VAO_Features_Lbl(:,end+1)=RealLbl;
VAOFinal=VAO_Features_Lbl;
%% Plot
figure;
plot(BestCost,'k','LineWidth',2);
xlabel('Iteration');
ylabel('Best Cost');
ax = gca;
ax.FontSize = 10;
ax.FontWeight='bold';
```

```
grid on;
title(['VAO Best Cost is :  ',num2str(BestCost(it))])
```

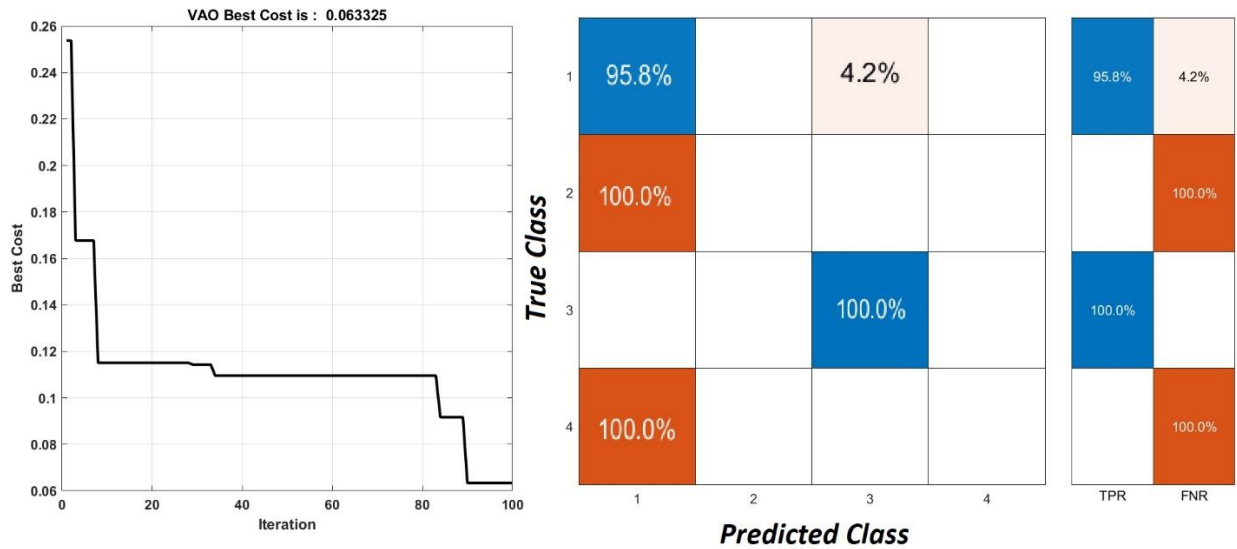

# ● **Bibliography**

---

**My MathWorks:**
https://www.mathworks.com/matlabcentral/profile/authors/9763916
**My GitHub:**
https://github.com/SeyedMuhammadHosseinMousavi
**My LinkedIn:**
https://www.linkedin.com/in/smuhammadhosseinmousavi/
**My ORCID:**
https://orcid.org/0000-0001-6906-2152
**My Google Scholar:**
https://scholar.google.com/citations?user=PtvQvAQAAAAJ&hl=en
**My RG:**
https://www.researchgate.net/profile/Seyed-Mousavi-17

---

*GitHub Repository:*

*https://github.com/SeyedMuhammadHosseinMousavi/Introduction-to-Facial-Micro-Expressions-Analysis-Using-Color-and-Depth-Images-a-Matlab-Coding-Appr*

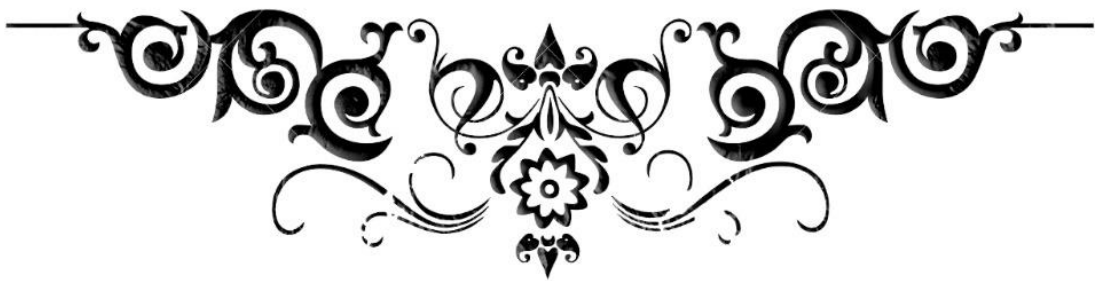